%% file: acl_latex.tex
\documentclass[12pt]{article}

\usepackage[margin=1in]{geometry}
\usepackage{times}
\usepackage{latexsym}
\usepackage{amsmath}
\usepackage{amssymb}

\usepackage{xcolor}
\usepackage{tikz}
\usetikzlibrary{arrows.meta,positioning,calc,fit,backgrounds,decorations.pathreplacing}

\definecolor{ink}{HTML}{1F2428}
\definecolor{mid}{HTML}{6B7075}
\definecolor{light}{HTML}{E9ECEF}
\definecolor{paper}{HTML}{FAFAF7}
\definecolor{accent}{HTML}{2F6F73}
\definecolor{accentlight}{HTML}{DCECEA}
\definecolor{warm}{HTML}{F1EEE7}

\usepackage[T1]{fontenc}
\usepackage[utf8]{inputenc}
\usepackage{microtype}
\usepackage{inconsolata}

\usepackage{setspace}

\usepackage{multirow}
\usepackage{array}
\usepackage{graphicx}
\graphicspath{{./}}
\usepackage{subcaption}
\usepackage{booktabs}
\usepackage{natbib}
\usepackage{xcolor}
\usepackage[hyphens,spaces,obeyspaces]{url}
\usepackage{float}
\usepackage{caption}
\usepackage{tabularx}
\usepackage{authblk}

\makeatletter
\AtBeginDocument{\check@mathfonts}
\makeatother

\usepackage{listings}
\usepackage{titlesec}

\titleformat{\paragraph}
  {\normalfont\normalsize\bfseries}
  {}{0pt}{}

\titlespacing*{\paragraph}
  {0pt}{1.0ex}{0.8ex}

\usepackage[colorlinks=true,linkcolor=blue!60!black,citecolor=blue!60!black,urlcolor=blue!60!black]{hyperref}

\title{Interpreting and Steering LLM Agents \\ for Social Simulations}

\author[1]{Jiayue Gaveal Fan%
  \thanks{Corresponding author: \texttt{gaveal.fan@berkeley.edu}. We are grateful to conference participants at the MIT BIG.AI conference and the NBER AI and Economic Measurement conference for helpful feedback. All errors are our own.}}
\author[1]{Arul Murugan}
\author[1]{Shreyas Krishnan}
\author[1,2]{Abhishek Nagaraj}

\affil[1]{Data Innovation and AI Lab (DIAL), University of California, Berkeley}
\affil[2]{National Bureau of Economic Research (NBER)}

\date{September, 2026}

\begin{document}
\maketitle
\begin{abstract}

\input{sections/abstract.tex}
\end{abstract}

\newpage

\setstretch{1.5}

\section{Introduction}
\input{sections/introduction}

\section{Related Work}
\input{sections/related_work}

\section{Methodology}
\input{sections/methodology}

\section{Results}
\input{sections/results}

\section{Discussion}
\input{sections/discussion}

\bibliographystyle{acl_natbib}
\bibliography{backup,custom}

\appendix

\section{Technical Appendix for Prompting and Sparse Autoencoder Setting}
\input{sections/appendix_sae}

\section{Technical Appendix for Probe-based Steering and Analysis}
\input{sections/appendix_probes}

\clearpage
\section{Appendix Figures and Tables}
\input{sections/figures_tables_appendix}

\end{document}

%% file: sections/abstract.tex
Simulations based on large language models (LLMs) have proven to be powerful for understanding human behavior, making them valuable additions to the social scientific toolkit. However, LLMs are ultimately black boxes based on deep neural networks which limits their value for social science. This is because of a lack of (i) \textit{interpretability}: i.e. the ability to assign clear mechanisms driving observed behavior; and a lack of (ii) \textit{steerability}: i.e. the ability to mute or amplify specific theoretically meaningful mechanisms of action to drive specific model behavior. Here, we demonstrate how the black box could be opened up to further enrich LLM-based simulations. Specifically, we compare three types of methods: (1) prompt-based manipulation, (2) SAE-derived feature steering, and (3) probe-based direction steering and examine their utility for LLM-based social scientific simulations. We do so by interpreting and steering two foundational components of human behaviors, namely preferences (risk attitudes, altruism) and capabilities (divergent creativity, product innovation), operationalized using four classic economic and creative tasks implemented as natural-language interactions.  Overall, our results show that SAE- and probe-based techniques often outperform basic prompt-based methods for steering LLM agents, although this advantage depends on the specific prompting strategy involved. Together, SAEs and probes constitute an effective pipeline for social scientists seeking to interpret and steer agents in social simulations: SAEs decompose agents' internal representations into human-readable features, after which probes can reliably shift agents' behaviors in specified directions. We discuss implications of these methods for future work using LLM agents for social scientific simulations.

%% file: sections/introduction.tex
Large Language Models (LLMs) mimic humans on multiple dimensions: they understand natural languages, can communicate with humans, and exhibit similar behavioral patterns as humans \citep{Anthis2025}. They are helpful for social science as they can respond to different experimental scenarios at a lower cost as compared to an equivalent number of humans \citep{Park2024}. Further, generative agents based on LLMs can also be trained to mimic different personas of real humans, which allows social scientists to study heterogeneity in human behavior \citep{Jackson2025}. With generative agents based simulations, social scientists can develop new theories of human behavior more cost-effectively and more efficiently by conducting experiments with these agents rather than relying solely on human participants \citep{Tranchero2024}. Given these advantages, social scientists have been motivated to use generative agents to conduct social simulations \citep{Horton2023, Chen2023}. 

Having said that, LLM-based simulations might not always faithfully match human behavior in all cases \citep{gao2025take, hullman2026human}. Driving such inconsistency, the black-box nature of generative agents has been proven to be a major roadblock for social scientific simulations \citep{Anthis2025}. Social scientists build new theories either inductively, by uncovering the mechanisms underlying observed behaviors, or deductively, by deriving hypotheses from abstract concepts and testing them through quantitative or experimental methods \citep{Davis2007, Hedstrom2005}. Black-box LLMs pose important roadblocks for both these approaches. Induction requires interpretability, i.e. the ability to observe behaviors and understand underlying mechanisms driving such behavior. Deduction on the other hand requires steerability, i.e the ability to magnify or mute underlying mechanisms to observe predicted behavior. Out-of-the-box, LLMs are unable to be interpreted or steered, which limits their potential as a powerful tool for social science research. Insofar as LLM-based simulations offer a useful avenue for theorizing about human behavior in similar settings, the ability to understand their inner workings and steer them via targeted interventions would greatly add to their utility for advancing social science.

Here, we develop a framework to compare alternate methods for looking into the LLM black-box for their utility in interpreting and steering the generative agents in social scientific simulations. We have two goals. Interpretability, or understanding potential inner mechanisms driving observed LLM behavior and Steering, or the ability to perturb mechanisms of action to study the causal effects of those mechanisms on downstream outcomes. To do so, we leverage emerging methods in mechanistic interpretability, often dubbed as "neuroscience" for generative AI models \citep{nanda2023progress}, a field of computer science that has applied these techniques to promote model safety and improve the AI training pipeline. Here, we adapt these techniques to social scientific simulation and demonstrate their utility in this context as well. 

First, we demonstrate how sparse autoencoders (SAEs) can open up the LLM black-box to interpret model behavior. SAEs are pre-trained neural networks that accompany the focal LLM that learn to decompose model activations into human-interpretable features \citep{cunningham2023sparse}. In doing so, they can inductively reveal the behavioral concepts represented inside the model in an unsupervised fashion at inference. We then compare the effectiveness of standard prompt-based methods of control (where LLM agents are prompted to assume certain personas) with SAE-based feature steering and a simpler alternative, linear probes \citep{Alain2017}. Both methods rely on amplifying or muting identified features representing certain mechanisms of underlying interest that are hypothesized to be driving LLM behavior. The analogy is to the genetically-modified mouse used in biological settings that are tailored to study specific diseases, genes or mutations. 

We demonstrate the validity of the proposed methods and focus on simulating two primitive dimensions of human behavior, specifically preferences and capabilities, across a set of classic experimental tasks. For preferences, we look at the lottery game (risk) \citep{Edwards1954} and the ultimatum game (altruism) \citep{Camerer1995}. For capabilities, we examine simple tasks capturing divergent creativity and product innovation. For each game, we use SAEs to understand what features are activated within the model with a view to uncovering mechanisms potentially driving agentic behavior. We then build probes for the central feature driving model behavior (e.g. risk-taking intention in the case of lotteries) and compare the effectiveness of using probe- and SAE-based method of steering these features (e.g. amplifying the desire for risk-taking) with prompt-based methods of steering. 

Across all four games, we find that SAE and probe-based steering techniques perform more favorably as compared to basic prompt-based methods
 of intervention, although this advantage narrows when compared with more advanced prompting strategies, such as few-shot and chain-of-thought prompting. Specifically, (1) SAE-based methods provide both better understanding and greater control of agent behavior as
 compared to the basic prompt-based methods; (2) Probe-based methods exhibit greater steerability on pre-defined traits of the agents while SAE-based methods are capable of exploring and steering previously unknown traits. For example, in the lottery game, SAEs are able to identify the underlying mechanism of ``attitudes towards risk'' that could be driving agent behavior and we can modify agent's choices by steering this feature. Probes go further by allowing precise control for the targeting of a specific degree of risk aversion. Our approach illustrates how social scientists can gain deeper insight into and exert finer control over generative agent
 simulations by leveraging the representational and mechanistic understanding of agent behavior through interpretability and steering
 techniques for deep neural networks. 

We make three contributions to the emerging literature of AI simulations in social science \citep{Horton2023, Park2024, Kazinnik2023}. First, we provide the first comparative exploration of prompt-based, mechanistic interpretability (SAEs), and representational interpretability (probes) approaches for understanding and steering LLM agent behavior in social science simulations. Second, we demonstrate how SAEs and probes can provide interpretable, fine-grained control over agent behaviors along major human-like dimensions, surpassing the capabilities of prompt-based techniques alone. And finally, we validate our framework across a set of classic experimental tasks that measure both preferences and capabilities, providing early proof-of-concept for leveraging mechanistic and representational interpretability in generative agent simulation. Collectively, our paper points the route towards a more interpretable and steerable tool of social scientific simulations using generative agents.

%% file: sections/related_work.tex
\subsection{Using Generative Agents in Social Science}
Social scientists have long leveraged the power of digital transformation to process massive data and address social scientific questions \citep{King2011}. Recently, the adoption of machine learning and artificial intelligence as a methodological approach to both inductive and deductive research \citep{Davis2007} has aided social scientists on a comprehensive scale. For example, text mining via machine learning tools aids pattern recognition for inductive research \citep{Lazer2009, Salganik2017}. On the other hand for deductive research, advancements in computational capability have enabled researchers to use methods such as agent-based modeling to deduce macro-level patterns from sets of micro-level rules \citep{Epstein1999, Macy2002}. 

Given the recent rapid development of LLMs out of traditional computational methods, we ask whether LLMs can serve as a new methodological instrument for the social sciences. Current research presents LLMs as a tool that leverages artificial intelligence systems designed to understand and generate human language by learning from large-scale human-produced text data \citep{Llama2024}. With language as its interface, LLMs are capable of communicating with others, including humans and other LLMs. This allows LLMs to be capable of simulating human behaviors through their output texts. Their similarity with humans is seen in both individual and group settings. LLMs are able to generate similarities of real-human level of heterogeneity in demographics \citep{Park2024}, economic rationality \citep{Chen2023, Xie2025}, cognitive abilities \citep{Toubia2025}, and other fundamental attributes \citep{Abdulhai2024}. LLMs also carry social abilities that allow them to organize social events in artificial settings \citep{Park2023}, and can generate networks whose structures are similar to human social networks \citep{Chang2025}. The promising performance of LLMs has prompted researchers to replicate experiments traditionally conducted on humans using generative agents powered by LLMs \citep{Horton2023, Tranchero2024}. 

Using generative agents in experiments offers several advantages, including low cost, easy accessibility, and a close resemblance to real-human behaviors. Complementing the deterministic nature of agents in traditional agent-based modeling, generative agents introduce real-world contextual nuance and variability into simulations \citep{Gao2024}. Hence, generative agents are helpful in multiple avenues: piloting untested experiments to observe possible outcomes and improving existing deductive tools, which complement the traditional data-driven approach in using machine learning to social scientific research \citep{Grimmer2021}.

Despite their many benefits, generative agents behave like a black box -- we cannot clearly interpret why they act as they do, nor reliably steer them toward specific mechanisms \citep{Jackson2025, Ludwig2024}. The interpretability problem arises because LLMs are trained on massive, weakly documented corpora, making it impossible to know whose voices and assumptions are represented; this turns them into ``stochastic parrots'' with little accountability for output quality \citep{Bender2021}. Even when post-hoc explanations are available, they rarely reflect the model’s true reasoning and can mislead downstream analyses \citep{Rudin2019}. The steerability problem follows from this opacity: the same observed behavior can emerge from entirely different internal mechanisms, meaning researchers cannot isolate or manipulate the causal processes they wish to study. As a result, generative agents risk obscuring rather than clarifying the mechanisms behind theories \citep{gao2025take}. However, if we can open this black box, we gain the ability to both understand the mechanisms driving the agents' behaviors and intentionally modify them, expanding the credible uses of generative agents for the theory building in social science.

\subsection{Opening the Black Box}
Most LLMs are built on the Transformer architecture, which converts text into numerical representations and processes these representations through multiple layers of attention and feedforward networks to predict the probability distribution over the next token \citep{Vaswani2017}. An input prompt is first split into text tokens, and each token is mapped to a high-dimensional embedding vector. These embeddings are then passed through a sequence of Transformer blocks. At each layer, the model produces internal activations, or hidden states, that encode information about the input in increasingly contextualized ways. These activations are not directly human-readable, but they contain the intermediate representations through which the model tracks linguistic structure, semantic meaning, task context, and behaviorally relevant concepts. After the final layer, the model projects the hidden state into logits over the vocabulary, which are then converted into probabilities for possible next tokens. The next generated token is sampled or selected from this distribution, appended to the text, and the process repeats autoregressively.

Given the way LLMs are constructed, one approach to opening the black box is to prompt them to reveal their intermediate reasoning or ``chain of thought'' in their output. This approach demonstrates the behavioral codes of generative agents via natural language, making text-based analysis of their underlying reasoning processes possible \citep{Xie2025}. It also opens the black box as output generates,
making the reasoning process accessible as the behavioral patterns. 

Another approach to opening the black box is to investigate the interpretability of the models \citep{Alain2017, Elhage2021}. Because
a given model’s decisions are mediated by internal activations before they are converted into output logits, these activations provide a natural site for interpretation and intervention. Interpretability methods such as probes and sparse autoencoders can therefore be applied at selected layers to analyze or modify these internal representations before the model produces its next-token prediction. Probing, which is typically trained on labeled datasets for predefined concepts, is useful for revealing what information can be decoded from a neural network’s hidden representations \citep{Alain2017, Hewitt2019, Belinkov2022}. When applied to the behavioral reasoning of
generative agents, probes can serve as measures of specific behavioral dimensions, such as risk, by assessing the extent to which agents’ internal representations encode that characteristic. For example, if models can be clearly taught the difference between safe and risky choices in a narrow setting, then we might be able to train probes that can be used to steer this behavior. 

SAEs takes a different approach. They open the black box by reverse engineering the computations within Transformer layers to understand which activations contribute to particular behaviors \citep{Elhage2021, Nanda2023}. SAEs are the tools to achieve mechanistic interpretability, as they are small neural networks trained to reconstruct a model’s internal activations using a sparse latent code \citep{Bricken2023}. Because of this sparsity, each latent dimension captures a relatively distinct pattern, allowing various dimensions to function as human-interpretable "features." In the context of generative-agent behavior, SAEs can identify top-activating features associated with corresponding behavioral patterns. This provides a more transparent basis for interpreting and steering on simulations using generative agents.

Both probes and SAEs have been recently used within the computer science literature for safety-oriented analysis and behavioral intervention on frontier LLMs. Linear probes have been used to recover internal directions corresponding to truthfulness \citep{MarksTegmark2023} and for testing refusal behavior \citep{Arditi2024}, with the latter showing that ablating a single direction can reliably bypass an aligned model's safety training. SAEs have been scaled to frontier models for feature discovery and circuit-level analysis, including Anthropic's work on Claude 3 Sonnet \citep{Templeton2024}, DeepMind's Gemma Scope release \citep{Lieberum2024}, and end-to-end sparse feature circuits \citep{Marks2024}. To our knowledge, however, neither family of techniques has been systematically adapted to social-scientific simulation, where the target is not safety auditing but the inductive and deductive study of agentic behavior, which is the gap our paper addresses.

%% file: sections/methodology.tex
It is to the best of our knowledge that none of the existing methods for opening the black box of generative agents in social science simulations have been systematically explored or compared. Previous attempts have utilized SAEs to demonstrate the steerability of generative agents in financial activities or marketing messaging design \citep{Chen2025, angelopoulos2024causal}. We would like to push this methodological advance further by proposing to use prompting, SAE-based, and probe-based interpretability tools to understand generative agents’ behaviors during simulations and compare their usefulness in interpretability and steerability.

In this paper, we demonstrate our ways of opening the black box by focusing on individual agents' traits, as they are the foundations of other levels of analysis in social sciences, such as groups, teams, organizations, and societies. We evaluate two primitives of generative agents that mimic human behaviors: preferences and capabilities. We selected social experiments that have been conducted on humans and which established underlying mechanisms driving in the decision-making processes of experiment subjects. Existing results from these experiments help us interpret the mechanisms demonstrated by generative agents and verify if their behaviors are coherent with real humans if we prompt or steer the agents to have a specific characteristics.

Table~\ref{tab:implemented_experiments} lists the tasks and the settings for our simulations with generative agents. We begin with the lottery \citep{Edwards1954} and
the ultimatum game \citep{Camerer1995} paradigms, which are widely used in economics and psychology to study risk and altruism as core behavioral tendencies in one's preferences. We ask 40 agents to perform the task in 4 scenarios: baseline; prompting, where we use prompts to change the agents' persona to be more risk-taking or altruistic; SAE-steering, where we find the SAE feature activation for risk-taking and altruistic behaviors and manually steer them to a high value; and probe-steering, where we train a probe with data that exemplifies risk-taking or altruistic behaviors and use that to steer the model along the trait direction. Similarly, we find divergent creativity and product innovation as important components of capabilities and use tasks similar to Torrance Tests of Creative Thinking \citep{Torrance1966} to see how prompting and steering can change the generative agents' performances. In addition to baseline, prompting, and steering, we added one scenario to the capability-related tasks: high temperature, indicating the high randomness of model output. We added this scenario because psychologists argue that there is a positive correlation between randomness and creativity \citep{Campbell1960}, and we take that as a potential mechanism for generative agents to have creative outputs. We document the detailed description of how we set up the agents for different tasks in the appendix.

\input{tab-fig/expts.tex}

\subsection{Prompting Setting}

We conducted our experiments using Llama-3.3-70B-Instruct and built our experiment pipeline via Expected Parrot, an open-source app for simulations with generative agents \citep{Horton2024}, where we set up the agents for completing tasks and prompted them to boost specific traits. To establish whether interpretability-based interventions (SAE and probe steering) provide leverage beyond what is achievable through prompt engineering alone, we compare against a graded ladder of prompting baselines applied to the same model via Expected Parrot. Each baseline keeps the original task instruction $P_{\text{task}}$. The resulting condition set is $C \in \{\text{baseline}, \text{persona}, \text{cot}, \text{fewshot}, \text{fewshot\_cot}\}$ with no modifications to model weights or decoding parameters across conditions.

\paragraph{Few-shot exemplar prompting.}
Few-shot prompting prepends a small set of demonstrations to the task instruction so the model can in-context learn the desired output distribution \citep{Brown2020}. We use $N{=}6$ demonstrations for every task. For divergent creativity we draw the demonstrations from the Ocsai Alternative Uses Task corpus \citep{Organisciak2023}, a publicly released dataset of human-authored brick uses paired with human-rated originality scores on a 10--50 scale; we rank the brick subset by human originality, deduplicate, drop entries where Ocsai's object-anonymization left ungrammatical text, and retain the top-$N$. The exemplars are inserted as an inspiration block to the prompt. They are human-generated and the selection criterion (human originality) is independent of the downstream GPT-based judge, which avoids selection-evaluation circularity. For the lottery and ultimatum games, the demonstrations are synthetic question-answer pairs constructed over held-out reward / offer grids that are disjoint from the test sweeps. Formally, the prompt becomes
\[
P_{\text{fewshot}} = E_{1:N} \oplus P_{\text{task}},
\]
where $E_{1:N}$ denotes the bulleted exemplar block and $\oplus$ denotes string concatenation. Because the Ocsai corpus is task-specific to brick AUT, the few-shot condition is omitted from the stapler product-innovation task.

\paragraph{Chain-of-thought prompting.}
Standard chain-of-thought prompting elicits intermediate reasoning prior to a final answer \citep{Wei2022, Kojima2022}. We append a single, task-appropriate trigger $I_{\text{cot}}$ to $P_{\text{task}}$, yielding
\[
P_{\text{cot}} = P_{\text{task}} \oplus I_{\text{cot}}.
\]
The trigger is tailored to the task family but always uses the same one-completion, reasoning-then-answer format. We document the detailed steps in the appendix.

\paragraph{Combined few-shot CoT and dose-response.}
The few-shot and CoT scaffolds are orthogonal and compose by concatenation:
\[
P_{\text{fewshot\_cot}} = E_{1:N} \oplus P_{\text{task}} \oplus I_{\text{cot}}.
\]
For divergent creativity this single combined condition (\texttt{fewshot\_cot}) is our strongest prompting variation. For the lottery and ultimatum games we additionally trace a prompting dose-response: we vary the fraction $\rho \in \{0.0, 0.25, 0.5, 0.75, 1.0\}$ of the $N$ demonstrations that take the target action (risky for lottery, accept for ultimatum), with the remaining $1{-}\rho$ taking the opposite action, and use reasoning-bearing demonstrations throughout. 

\subsection{Sparse Autoencoders Setting}
 On the interpretability side, we used a set of trained SAEs by Goodfire (an AI interpretability lab) on the same model trained on LMSYS Chat data \citep{McGrath2024a, Zheng2024}. These SAEs are trained on layer 50, which is an intermediate position among all the 80 layers of Llama-3.3-70B-Instruct, capturing the agent-level behavioral codes instead of just the syntax and perception of agents' conversations. While our SAEs already carry substantial meaning after training, they are not in human-readable forms. Goodfire then uses Claude to interpret the tokens and contexts from the trained SAEs and excludes 3.5\% of all the features that are socially harmful, leading to 65,536 features in total \citep{McGrath2024b}.

\begin{figure}[H]
    \centering
    \includegraphics[width=\linewidth]{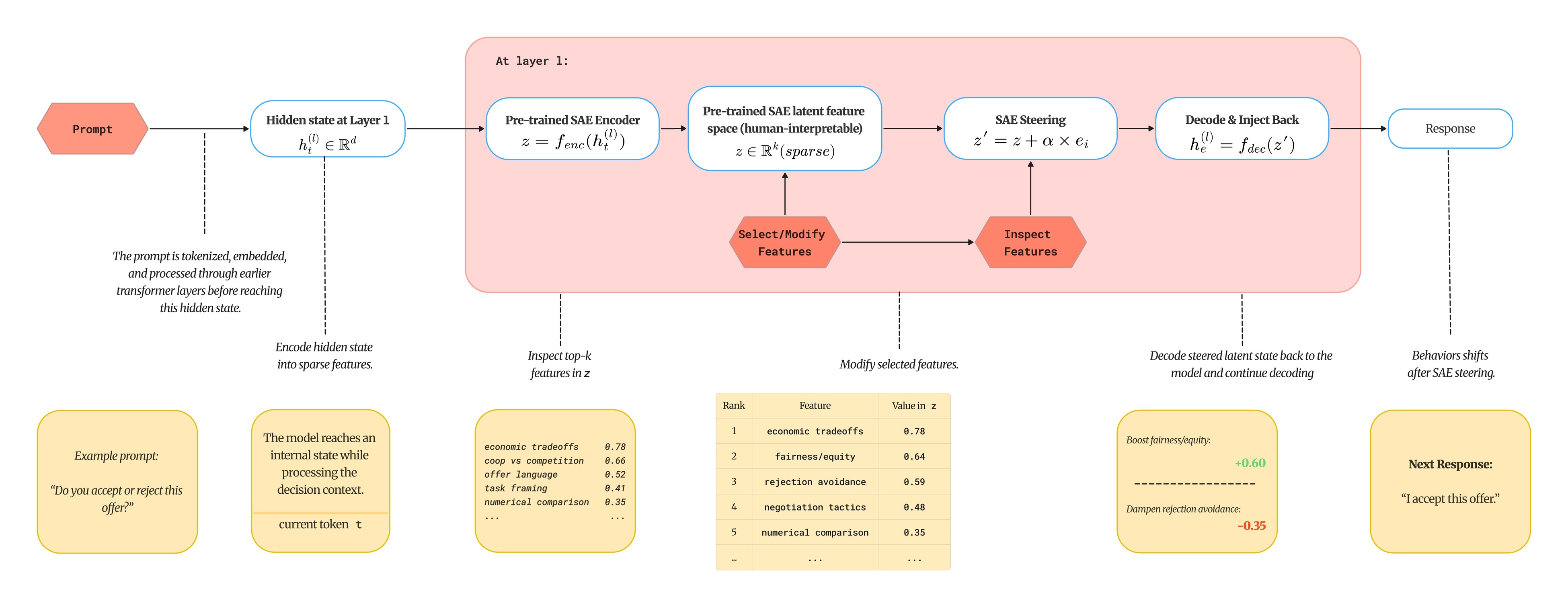}
    \caption{SAE interpretability and steering pipeline.}
    \label{fig:sae_schematic}
\end{figure}

Figure~\ref{fig:sae_schematic} illustrates the SAE pipeline for both interpretability and steering. Given a prompt, the LLM performs a forward pass up to intervention layer $\ell$, yielding a hidden state $\mathbf{h}_t^{(\ell)} \in \mathbb{R}^d$ at token position $t$. The SAE encoder maps this activation to a sparse latent code:
\[
\mathbf{z} = f_{\text{enc}}(\mathbf{h}_t^{(\ell)})
\]
where $f_{\text{enc}}: \mathbb{R}^d \to \mathbb{R}^k$ denotes the learned SAE encoder function and $\mathbf{z} \in \mathbb{R}^k$ is sparse (most entries are zero) and each non-zero dimension corresponds to an interpretable feature \citep{Bricken2023}. This sparse decomposition addresses the problem of superposition, where neural networks pack multiple features into fewer dimensions, making individual features difficult to isolate \citep{Elhage2022}. For interpretability, we inspect the top-$k$ activated features in $\mathbf{z}$ to understand which behavioral concepts drive the agent's response.

For steering, we modify selected feature activations in the latent space:
\[
\mathbf{z}' = \mathbf{z} + \alpha \cdot \mathbf{e}_i
\]
where $\mathbf{e}_i$ is a one-hot vector that selects feature $i$. A one-hot vector is a standard basis vector with exactly one entry equal to 1 at position $i$ and all other entries equal to 0. The scalar $\alpha \in [-1, 1]$ controls the steering strength. Negative $\alpha$ dampens the feature while positive $\alpha$ amplifies it. The modified latent is then decoded back to activation space:
\[
\mathbf{h}_t^{\prime(\ell)} = f_{\text{dec}}(\mathbf{z}')
\]
This steered activation $\mathbf{h}_t^{\prime(\ell)}$ is injected at layer $\ell$, and decoding continues through the remaining layers to produce the final response. This encode-modify-decode pipeline enables fine-grained control over specific interpretable features while preserving the model's overall coherence.

\subsection{Probes Setting}

\begin{figure}[H]
    \centering
    \includegraphics[width=\linewidth]{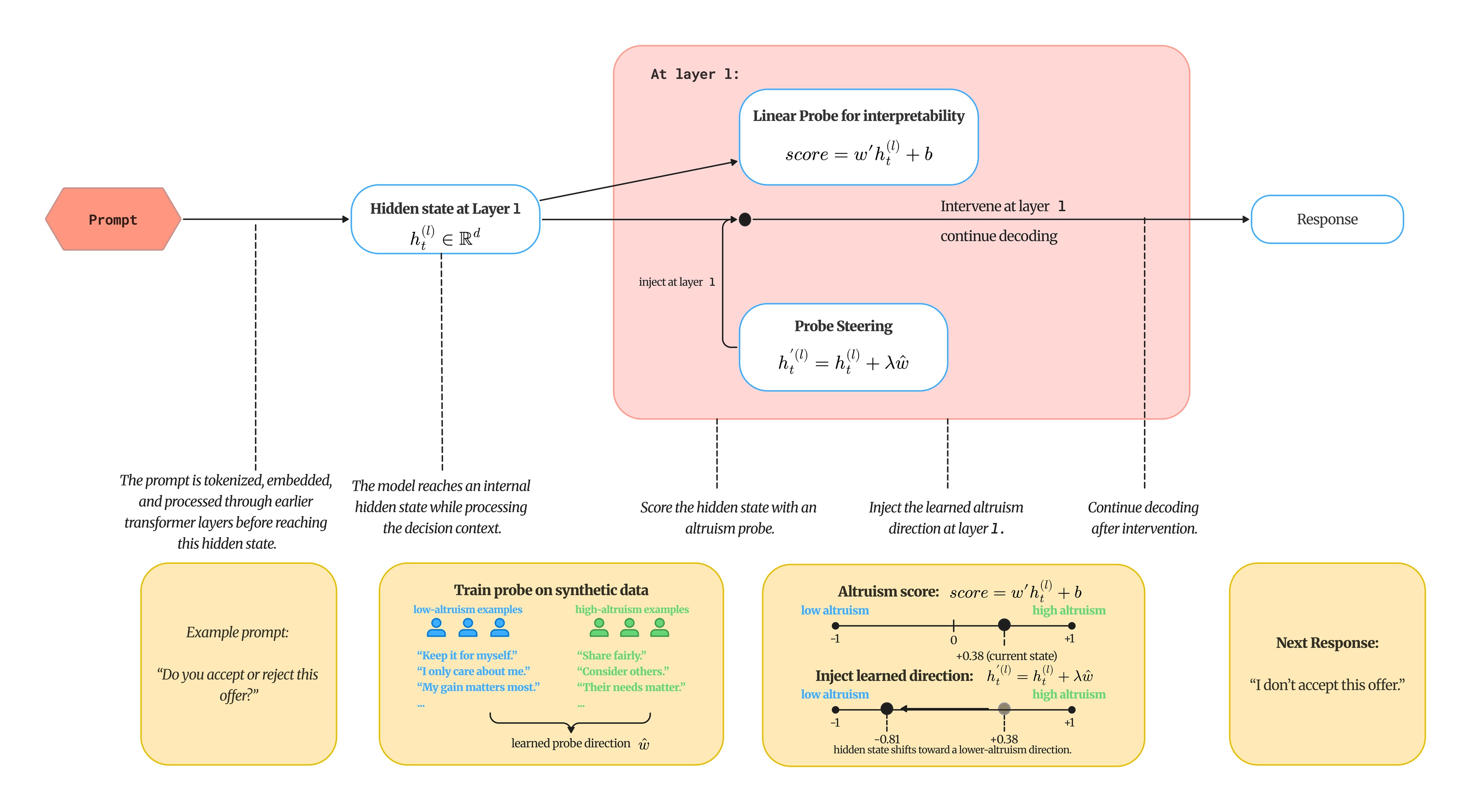}
    \caption{Probe scoring and steering pipeline.}
    \label{fig:probe_schematic}
\end{figure}

Figure~\ref{fig:probe_schematic} illustrates our probe-based pipeline for both scoring and steering. We train a logistic regression probe on intermediate layer activations (layer 48 of Llama-3.3-70B-Instruct) to identify directions associated with agents' traits \citep{Alain2017, Conneau2018}. Layer 48 was selected via a cross-validated sweep over layers and achieved the highest held-out performance.

Given a prompt, the LLM performs a forward pass up to intervention layer $\ell$, yielding a hidden state $\mathbf{h}_t^{(\ell)} \in \mathbb{R}^d$ at token position $t$. For scoring, the linear probe computes:
\[
\text{score} = \mathbf{w}^\top \mathbf{h}_t^{(\ell)} + b
\]
where $\mathbf{w} \in \mathbb{R}^d$ is the learned probe weight vector and $b$ is the bias term. This score quantifies the model's internal representation along the trait dimension (e.g., risk-seeking vs. risk-averse).

For steering, we add a scaled version of the probe direction to shift the model's behavior:
\[
\mathbf{h}_t^{\prime(\ell)} = \mathbf{h}_t^{(\ell)} + \lambda \cdot \hat{\mathbf{w}}
\]
where $\hat{\mathbf{w}} = \mathbf{w}/||\mathbf{w}||_2$ is the
unit-normalized probe direction and $\lambda$ controls the steering
strength \citep{Turner2024}. Positive $\lambda$ pushes the
representation toward the probe's positive class (e.g., more
risk-seeking), while negative $\lambda$ pushes toward the negative
class. The steered activation $\mathbf{h}_t^{\prime(\ell)}$ is then
injected at layer $\ell$, and decoding continues through the remaining
layers to produce the final response.

\paragraph{Synthetic Data Generation}

Building probes requires pre-defined features of agents to be measured. Because probes need labeled training examples that specify what constitutes ``high'' versus ``low'' on a given trait, we generated controlled synthetic trials where the ground-truth labels could be defined programmatically.

For the preference tasks, we generated synthetic data through the following procedure. First, we ran many trials where the agent is the base LLM decoding under a deterministic seed. Second, we varied the task parameter to trace a psychometric curve. In the lottery game, we varied the risky reward value and measured the switching point where $P(\text{Risky}) = 0.5$. In the ultimatum game, we varied the offer size and measured the acceptance threshold where $P(\text{Accept}) = 0.5$. Third, we converted each trial into a binary label from the agent's observed choice rather than from a median split of the target metric: in the lottery game, trials in which the agent chose the risky option were labeled ``high risk-taking'' and those in which it chose the safe option ``low risk-taking'', and in the ultimatum game accepted offers were labeled ``high altruism'' and rejected offers ``low altruism''. Labeling by the choice keeps the polarity faithful to the trait, because a risk-seeking agent takes the gamble even when the risky reward is small (equivalently, it switches to the risky option early, at a \emph{low} switching point), so its risky choices define the high-risk-taking class; the switching point and acceptance threshold are retained as the target metrics for calibrating steering strength, not as training labels. Fourth, we extracted activations at the selected intervention layer using a forward hook that captures the hidden state tensor at each layer.

For the creativity tasks, we generated synthetic data using an independent evaluation architecture. First, we generated creative outputs from the model across multiple prompts. Second, we used a separate LLM judge (GPT-5; \citealp{OpenAI2025}) to score each output on a 1 to 10 scale based on fluency, flexibility, originality, and elaboration.\footnote{Definitions of the 4 criteria: 1. Fluency: The ability to produce a significant number of relevant ideas in response to a given question. 2. Flexibility: The variety of categories from which one can generate ideas. 3. Originality: The uniqueness of the ideas generated. 4. Elaboration: The ability to expand upon, refine, and embellish an idea.} Third, we assigned binary labels based on median split, where outputs scoring above the median were labeled ``high creativity'' and those below were labeled ``low creativity.'' This independent evaluation ensures that the steering model is not grading its own homework, thereby addressing potential circularity concerns in capability assessment.

\paragraph{Probe Training Procedure}

We trained an $\ell_2$-regularized logistic regression probe to predict the binary labels from intermediate layer activations. The training procedure consisted of the following steps. First, we captured the hidden state tensor at the selected layer using a forward hook during a forward pass. We used the final token representation as a compact summary of the prompt context, as this position aggregates information from the entire input sequence. Second, we standardized the activation features prior to fitting by applying z-score normalization across the training set. Third, we selected the regularization strength via cross-validation to prevent overfitting while retaining discriminative power. Fourth, we trained probes across multiple layers and chose the layer with the best held-out discriminative performance. For our Llama-3.3-70B-Instruct runs, layer 48 achieved the highest cross-validation score. 

The trained probe yields both a scoring mechanism and a steering direction. The learned weight vector $\mathbf{w}$ and bias $b$ define a score $\mathbf{w}^\top \mathbf{h}_t^{(\ell)} + b$ that quantifies the model's internal representation along the trait dimension. The probe's weight direction, after unit normalization, becomes the steering direction for activation addition-style steering. The probe achieved 82\% classification accuracy on held-out preference trials, suggesting that preference-related information is linearly accessible in the model's representations.

\subsection{Evaluation Protocol for Creativity Tasks}
\label{subsec:eval_protocol}

The capability tasks rely on LLM judges to score open-ended responses on the four Torrance dimensions. Because the creativity probe is trained on machine-generated labels and the base model (Llama-3.3-70B-Instruct) shares much of its pretraining distribution with frontier judge models, single-judge evaluation conflates genuine creativity with one judge's idiosyncratic preferences. We therefore evaluate every creativity response with five independent LLM judges spanning four developers: GPT-5 \citep{OpenAI2025}, Claude Sonnet 4.6, Gemini 2.5 Pro, Kimi K2.6, and DeepSeek V4 Pro. Each judge applies the same four-dimension rubric, and we report the mean across judges as the primary score. Two protocol features prevent known biases. First, scoring is blind: responses are pooled across conditions and presented in randomized order so that no judge can identify which condition produced a response. Second, the rubric is length-controlled: the rubric instructs judges to score idea quality rather than response length, with no generosity prompt. Inter-rater agreement is reported in Appendix~\ref{subsec:multi_judge}; the mean pairwise Spearman correlation across judges is 0.77 (N=680 unique responses), and the relative ordering of conditions is preserved across all five judges.

%% file: tab-fig/expts.tex
\begin{table}[H]
  \centering
  \footnotesize
  \renewcommand{\arraystretch}{1.15}
  \setlength{\tabcolsep}{6pt}

  \begin{tabular}{
    p{3.1cm}
    p{7.4cm}
    p{3.0cm}
  }
    \toprule
    & \centering\arraybackslash \textbf{Task Description}
    & \centering\arraybackslash \textbf{Settings} \\
    \midrule

    \textit{Preference} & & \\
    \cmidrule(lr){1-2}

    Risk
    & Lottery: Focal agents are asked to choose between two options: 1. (Safe) Guaranteed 50 tokens; 2. (Risky) 50\% chance of $n$ tokens, 50\% of 0 tokens.
    & 40 agents, 3 scenarios, $n = [10,180]$ \\[4pt]

    Altruism
    & Ultimatum Game: Focal agents are asked to accept or reject the offer: the proposer was given 100 tokens and has decided to offer you $n$ tokens out of their 100 tokens. If accepted, you keep the proposed amount; if rejected, both of you get 0.
    & 40 agents, 3 scenarios, $n = [10,100]$ \\

    \midrule

    \textit{Capability} & & \\
    \cmidrule(lr){1-2}

    Divergent Creativity
    & List very detailed ways you can use a brick.
    & 4 scenarios \\[4pt]

    Product Innovation
    & List as many specific enhancements as you can that would make a stapler better.
    & 4 scenarios \\

    \bottomrule
  \end{tabular}

  \caption{Implemented Experiments}
  \label{tab:implemented_experiments}
\end{table}

%% file: sections/results.tex
We evaluate how different tools perform on preference and capability tasks in terms of their interpretability and controllability over agents’ behaviors. First, we showcase the top-ranked SAE features decomposed by the agents' responses in different settings to demonstrate the interpretability of their behaviors. Table~\ref{tab:interventions} details the interventions we have introduced to the agents for preference and capability tasks. In baseline scenarios, agents respond according to the task descriptions without interventions. In prompting scenarios, agents are prompted with different personas, specified by the known traits about the specific tasks. In steering scenarios, we manually adjust the activation strength of some specified related SAE features to alter the agents' behaviors. In high-temperature scenarios, which are designed specifically for capability tasks, agents are configured to produce more random responses to assess whether increased randomness leads to higher creativity. With the results from all the scenarios, inspecting the SAE features of generative agents allows us to peek into the behavioral codes of generative agents.

\input{tab-fig/interventions.tex}

Second, we showcase how prompting, SAE steering, and probe steering differ in their performances on the controllability of agents' behaviors. We provide different degrees of control in each approach as separate scenarios and show the agility in the agents' responses. For each scenario, we repeat the same steering interventions on 40 agents. In preference tasks, we record the percentage of agents changing their behaviors as reward values increase in the lottery and the ultimatum games; and in capability tasks, we record the average creativity scores with standard deviation in divergent creativity and product innovation tasks.

\subsection{Interpretability}

\subsubsection{Preference Tasks}
\begin{figure}[H]
    \centering
    \includegraphics[width=\linewidth]{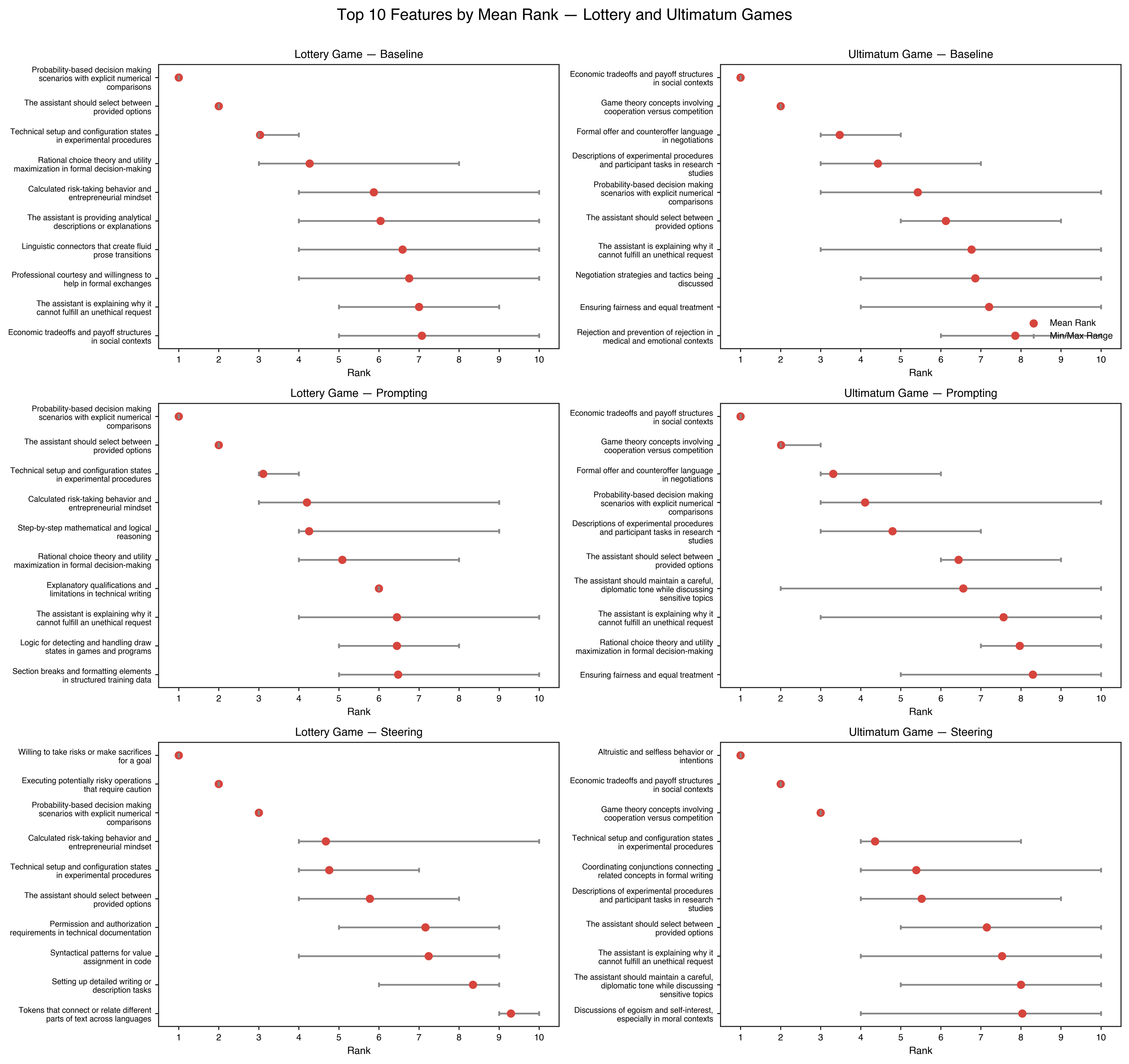}
    \caption{Feature Activations of Generative Agents on Preference Tasks}
    \label{fig:og-pref-activations}
\end{figure}

Figure~\ref{fig:og-pref-activations} details the top 10 features activated for generative agents in preference tasks.\footnote{The range of the top-activated features' rankings is generated from 40 agents.} We observe that most of the top 3 features show no variation in their minimum and maximum ranks. For the lottery game, the top 3 features are consistent across the baseline and prompting scenarios. "Probability-based decision making scenarios with explicit numerical comparisons" and "The assistant should select between provided options" occupy top 1 and 2 features for all agents in both scenarios; and "Technical setup and configuration states in experimental procedures" takes the 3rd place for most agents with few in the 4th place. High rankings of these features reinstate the underlying theory for the lottery game, for which agents use probability-based decision making to support rational choices. This result replicates the existing social theories and behavioral principles. However, the rest of the features are less common. For example, "The assistant is explaining why it cannot fulfill an unethical request" is a feature that provides less intuition in making decisions in the lottery games. Inspecting features like this can potentially inspire social scientists to add another layer of analysis beyond rational choices and probabilistic decision-making, which is to consider ethics in the lottery games. It could also indicate SAEs tendency to surface features that would otherwise be less relevant for humans, highlighting differences between the two data generating processes.

For the ultimatum game, the top 3 features show a similar lack of variation in ranks across baseline and prompting scenarios and disclose the underlying mechanisms of agents' behaviors. "Economic tradeoffs and payoff structures in social contexts", "Game theory concepts involving cooperation versus competition", and "Formal offer and counteroffer language in negotiations" confirm that agents grasp the essence of the ultimatum game in leveraging altruism and selfish decision-making principles in economic games. Other lower-ranked features, such as "Descriptions of experimental procedures and participant tasks in research studies," do not have a direct link with the existing theories about the ultimatum game but rather with the methodology we implement on the agents. This shows that SAE features not only capture the mechanisms underlying specific tasks, but also the agents behavioral codes in general.

Identifying SAE features also helps investigate the effects of controllability. In the steering scenario for both games, the top-ranked feature becomes the feature we manually steered, despite not being in the top-ranked feature list for the baseline scenario. 
The rest of the features are still similar to the ones in the baseline and prompting scenarios. This confirms the effects of feature steering as it mechanistically changes the underlying behavioral codes of generative agents. We can also observe that the top-ranked features for the prompting scenario are more similar to the features for the baseline scenario rather than steering scenario.

\subsubsection{Capability Tasks}

\begin{figure}[H]
    \centering
    \includegraphics[width=\linewidth]{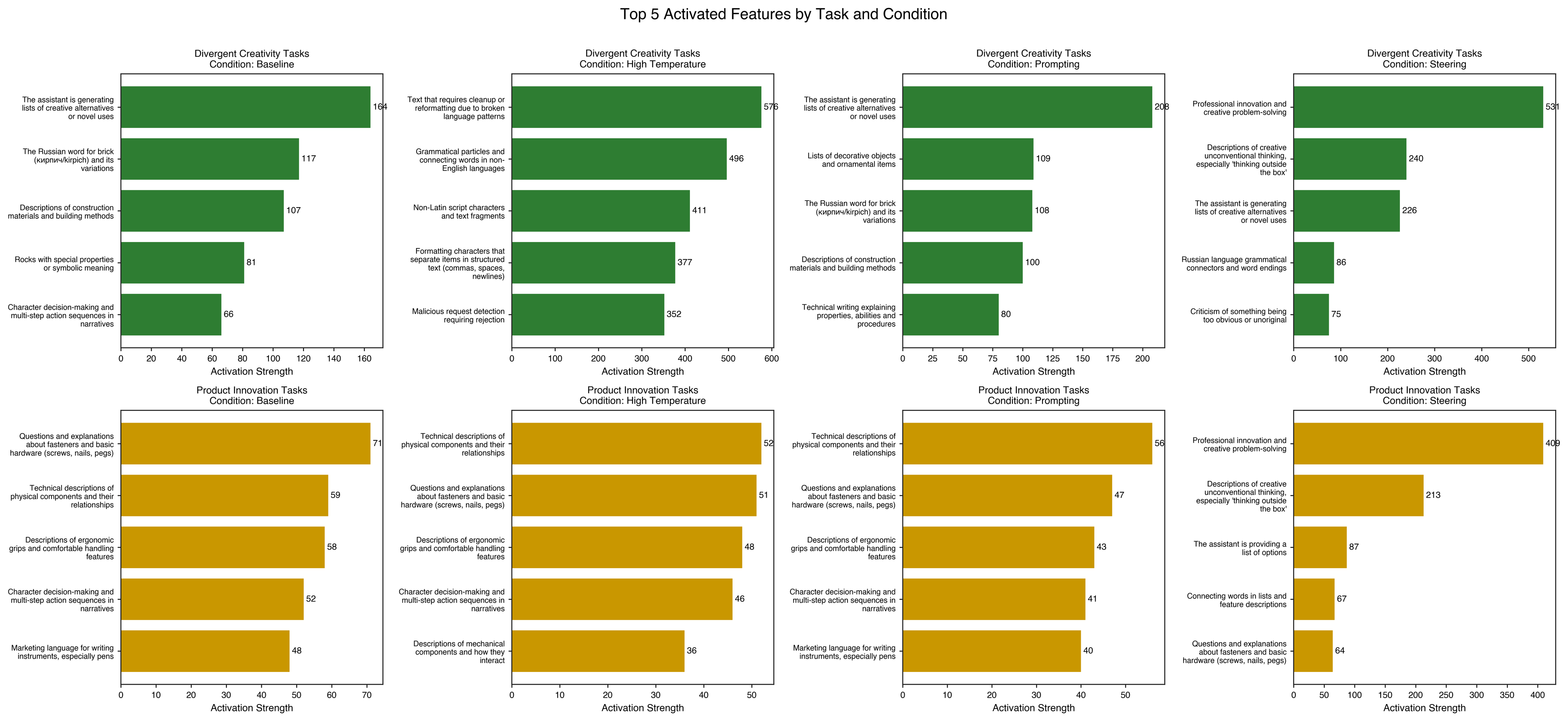}
    \caption{Feature Activations of Generative Agents on Capability Tasks}
    \label{fig:activations_capability}
\end{figure}

Figure~\ref{fig:activations_capability} compares the top five activated features within generative agents across two task types, divergent creativity and product innovation, under four conditions: baseline, high temperature, prompting, and steering.\footnote{Activation strength of top-ranked features is generated with single-agent feature decomposition, different from the range of 40 agents' activated feature rankings in Figure~\ref{fig:og-pref-activations}} In the baseline setting, the model activates reasonable but relatively generic features, such as brainstorming lists or describing technical components. Increasing temperature introduces noise, with features tied to messy formatting, multilingual fragments, and irregular text, indicating exploration without control. Prompting makes behavior more focused, aligning activations more closely with relevant concepts like creative alternatives and technical relationships, though still indirectly. By contrast, steering produces the strongest and most targeted activations, concentrating on concepts such as professional innovation, unconventional thinking, and structured feature descriptions. Together, these patterns show that feature activations decomposed by SAEs are useful for understanding the mechanisms supporting agentic behaviors. 

Two practical implications of this behavioral interpretability include theoretical validation and generation. For example, if one hypothesizes the creativity related to innovation can be extended from considering the physical restraints of the current products, the top 1 or 2 SAE features activated for agents in the product innovation tasks confirm this hypothesis. However, while the SAE features offer some extent of validation, we need to carefully evaluate their activation strength and underlying logic. One counterexample of this is the second SAE feature activated for divergent creativity tasks, "The Russian word for brick", which does not offer intuitive mechanisms of creativity but rather can be seen more as a linguistic hint within language models. Theoretical generation is also possible: we can use the top-ranked activated features to brainstorm mechanisms that may underlie agentic behaviors, even if those mechanisms are not yet theoretically grounded in established social-scientific behavioral patterns. For example, "marketing language for writing instruments" is not an usual behavioral code for a stapler innovation task. This suggests that creativity in product innovation may stem from connecting the focal product to others that appear in the same context. With this example, we demonstrate that investigating SAE features is beneficial for inductive theorizing.

\subsection{Steerability}

After characterizing the interpretability of reasoning that underlies agents’ behaviors, we next examine whether, and to what extent, these behaviors can be steered. To this end, we leverage all three approaches: prompting, probe-based methods, and sparse autoencoders.

\subsubsection{Preference Tasks}

\begin{figure}[H]
    \centering
    \includegraphics[width=\linewidth]{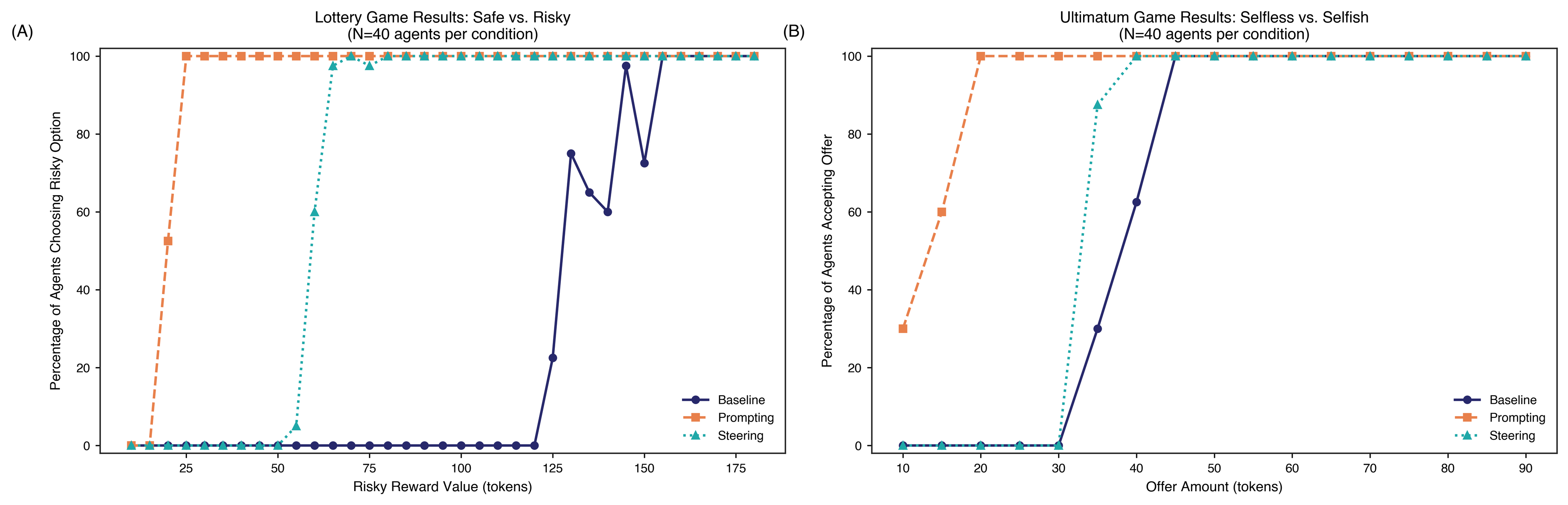}
    \caption{Performance of Generative Agents on Preference Tasks}
    \label{fig:og-preference}
\end{figure}

We first use the lottery game to demonstrate risk-taking behaviors of humans. According to our setting, the risk neutral option to be equal to 100, as that is the expected value of risky option is the same as the safe option. In the baseline scenario demonstrated in the left panel of Figure~\ref{fig:og-preference}, generative agents exhibit human-like risk aversion, shifting their preference only once the risky reward exceeds 100 tokens, around 125. In the prompting scenario where we explicitly prompt the agents to be risk-taking in their traits, they start pivoting to risky option when the reward is only 20 tokens. We also observe changes in the steering scenario, where we identify the SAE features responsible for risky behaviors and manually boost them for outputs. However, agents only start to pivot to risky option when risky rewards become more than 50. This indicates that prompted agents are more risk-taking than steering agents.

\begin{figure}[H]
    \centering
    \includegraphics[width=0.8\linewidth]{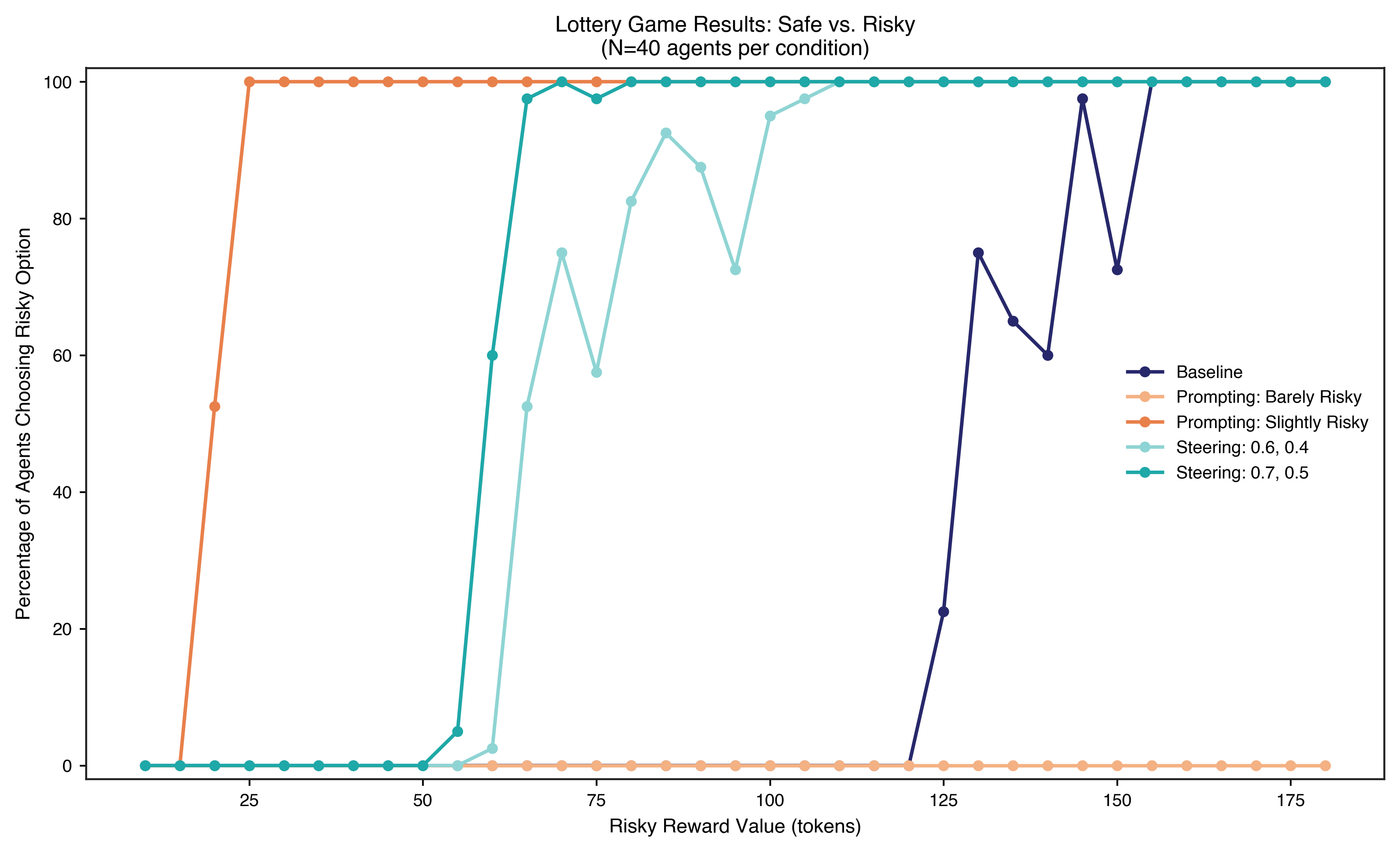}
    \caption{Lottery game results across five experimental conditions.}
    \label{fig:five_conditions}
\end{figure}

We extrapolate further about what is behind this phenomenon and we find that steering via SAEs controls the agents' behaviors in a higher granularity than prompting. As shown in Figure~\ref{fig:five_conditions}, slight changes in prompting language could lead to big yet unpredictable results in agents' behaviors. In contrast, steering with different activation strengths lead to smoother changes in when the agents start to take risky options. The baseline shows risk aversion with agents switching to risky option around 125 tokens. Prompting variants demonstrate binary behavior: ``barely risky'' shows no effect, while ``slightly risky''  produces extreme risk-seeking even at very low rewards (20-25 tokens). In contrast, SAE steering conditions show dose-dependent, controllable effects: moderate steering (activations 0.6/0.4, green dotted) produces gradual transition starting around 75 tokens, while stronger steering (activations 0.7/0.5, red dashed) shows sharper but still controlled transition around 65-70 tokens. This demonstrates that SAE steering provides fine-grained control over agent behavior, whereas basic prompting produces unpredictable all-or-nothing effects.

\begin{figure}[H]
    \centering
    \begin{subfigure}[b]{0.48\linewidth}
        \centering
        \includegraphics[width=\linewidth]{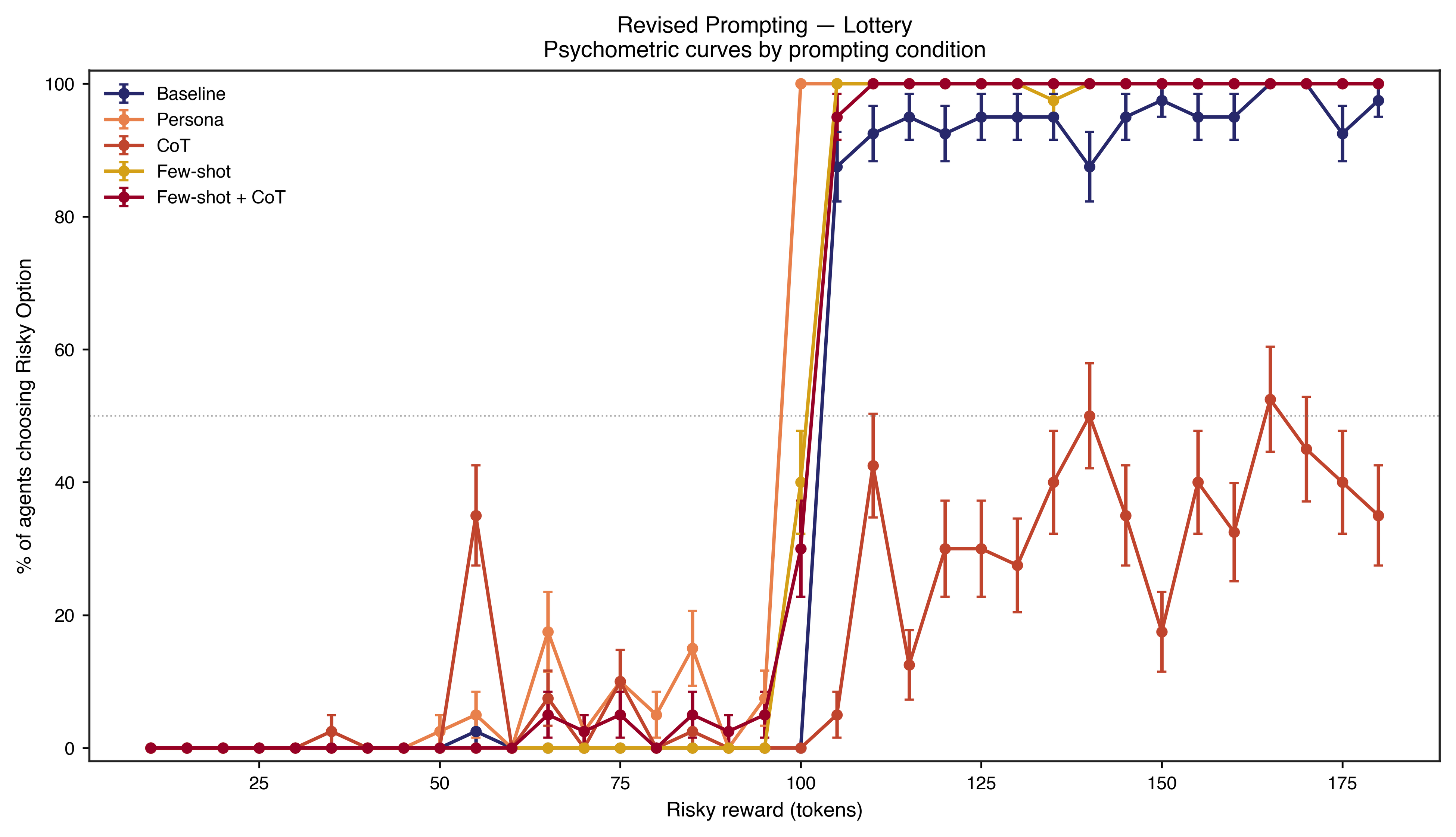}
        \caption{Lottery.}
        \label{fig:revised_prompting_lottery}
    \end{subfigure}
    \hfill
    \begin{subfigure}[b]{0.48\linewidth}
        \centering
        \includegraphics[width=\linewidth]{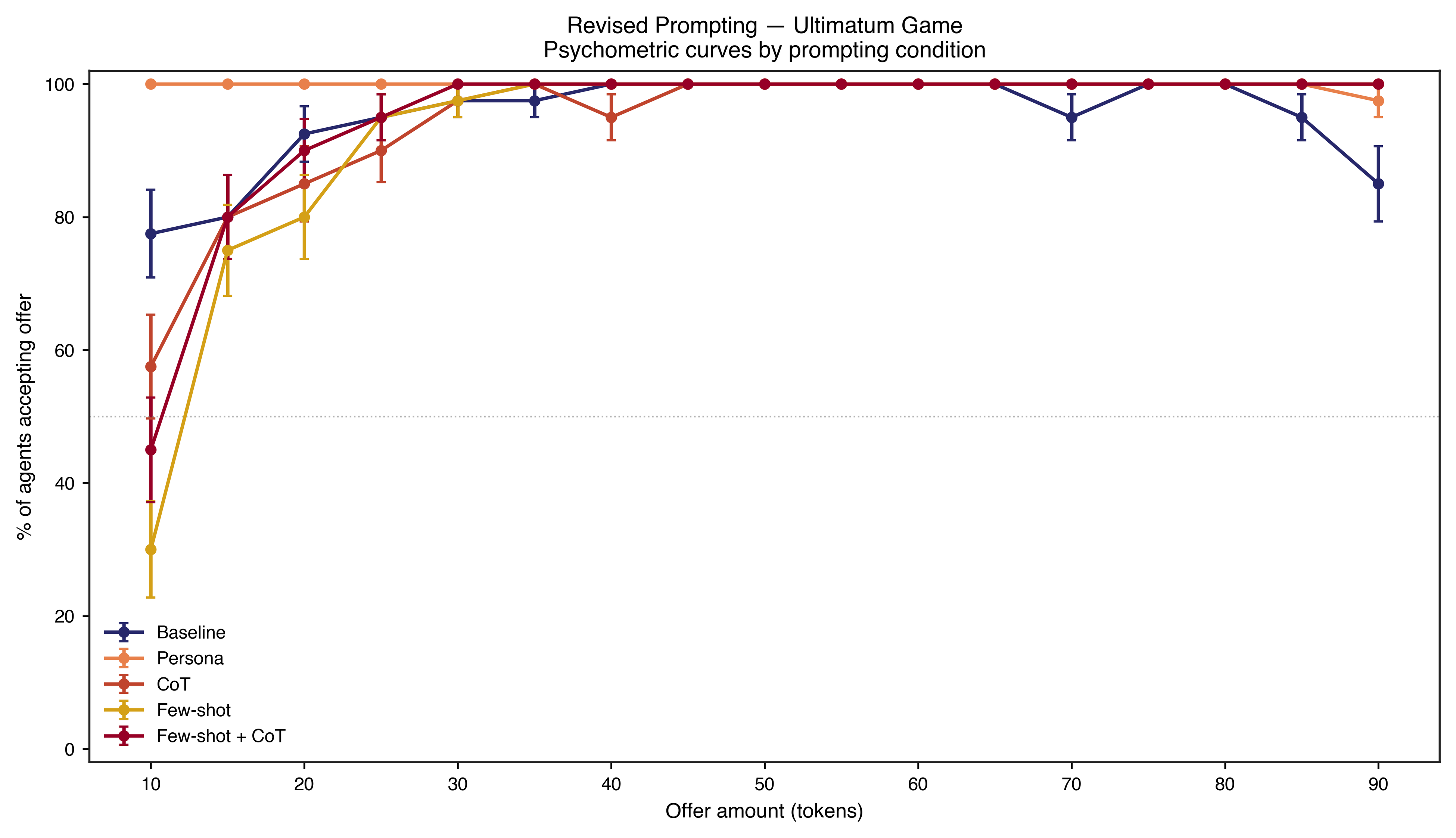}
        \caption{Ultimatum.}
        \label{fig:revised_prompting_ultimatum}
    \end{subfigure}
    \caption{Psychometric curves under the prompting ladder.}
    \label{fig:revised_prompting_preference}
\end{figure}

We enhance our understanding of prompting the lottery and ultimatum games under a graded ladder applied to the same Llama-3.3-70B-Instruct backbone: baseline, persona, zero-shot CoT \citep{Kojima2022}, fewshot \citep{Brown2020, Wei2022}, and a 5-level few-shot CoT dose-response sweep (fewshot\_cot with the fraction of target-aligned demonstrations $\rho \in \{0, 0.25, 0.5, 0.75, 1.0\}$). Figure~\ref{fig:revised_prompting_lottery} and Figure~\ref{fig:revised_prompting_ultimatum} report the psychometric curves; per-condition $\Delta$ vs.\ baseline (Appendix Figures~\ref{fig:rp_lottery_delta} and~\ref{fig:rp_ultimatum_delta}) and the dose-response sweep (Appendix Figures~\ref{fig:rp_lottery_dose} and~\ref{fig:rp_ultimatum_dose}) appear in the appendix. The two preference tasks diverge sharply. In the lottery, no prompting variant induces graded risk-seeking: the switching point stays within 97--101 tokens for all dose levels and CoT pushes it the wrong way to $\sim$140 tokens (more cautious). In the ultimatum game, by contrast, few-shot CoT does deliver a dose-dependent shift in the acceptance threshold from roughly 10 to 40 tokens. The behavioral advantage of SAE/probe steering over prompting therefore depends on the construct: it is sharpest on risk, where prompting offers little leverage, and narrows on altruism, where strong prompting is itself a competitive baseline.

\begin{figure}[H]
    \centering
    \includegraphics[width=\linewidth]{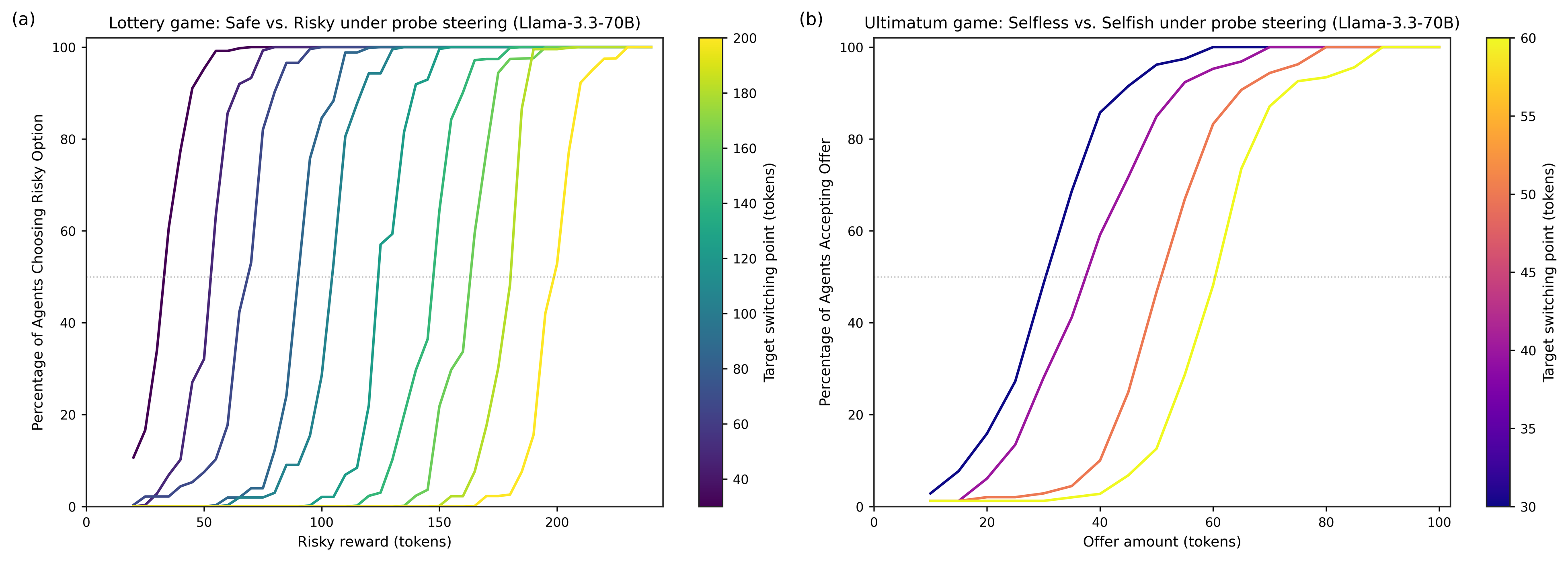}
    \caption{Performance of using probes to steer generative agents in preference tasks}
    \label{fig:preference_dose_response}
\end{figure}

Probe-based steering also demonstrates successful behavioral control in the lottery task, as demonstrated in Figure~\ref{fig:preference_dose_response}. We achieve precise control over risk preferences by calibrating $\lambda$ values to target specific switching points, defined as the risky reward value at which the probability of choosing the risky option equals 0.5. Through a systematic search over $\lambda$ values, we place the switching point across a wide range from approximately 30 to 200 tokens, with achieved values closely matching their targets (mean absolute error of about 2 tokens). Higher positive $\lambda$ values increase risk-seeking behavior (lower switching points), while negative values enhance risk aversion.

We demonstrate that ultimatum game has similar results to lottery game in how prompting and steering could effectively change the behaviors of generative agents. More specifically, our setting of the ultimatum game makes an even share of the tokens at 50. As shown in the right panel of Figure~\ref{fig:og-preference}, the baseline scenario demonstrates that the generative agents behave slightly more altruistically than perfectly rational humans, where they start to take the offer when it exceeds 30. We prompt the agents to have altruism or selfless behaviors in the prompting scenario and the effects are outstanding, as they start to take the offer even when the offer is only 10 tokens. Panel (b) of Figure~\ref{fig:preference_dose_response} for using probes to alter agents' behaviors in ultimatum games also demonstrates similar patterns. It shows that the acceptance threshold shifts systematically under stronger steering, with target switching points ranging from 30 to 60 tokens. Curves are approximately monotonic with diminishing returns at extreme $\lambda$ values, indicating saturation of the intervention.

\subsubsection{Capability Tasks}

\begin{figure}[H]
    \centering
    \begin{subfigure}[b]{0.48\linewidth}
        \centering
        \includegraphics[width=\linewidth]{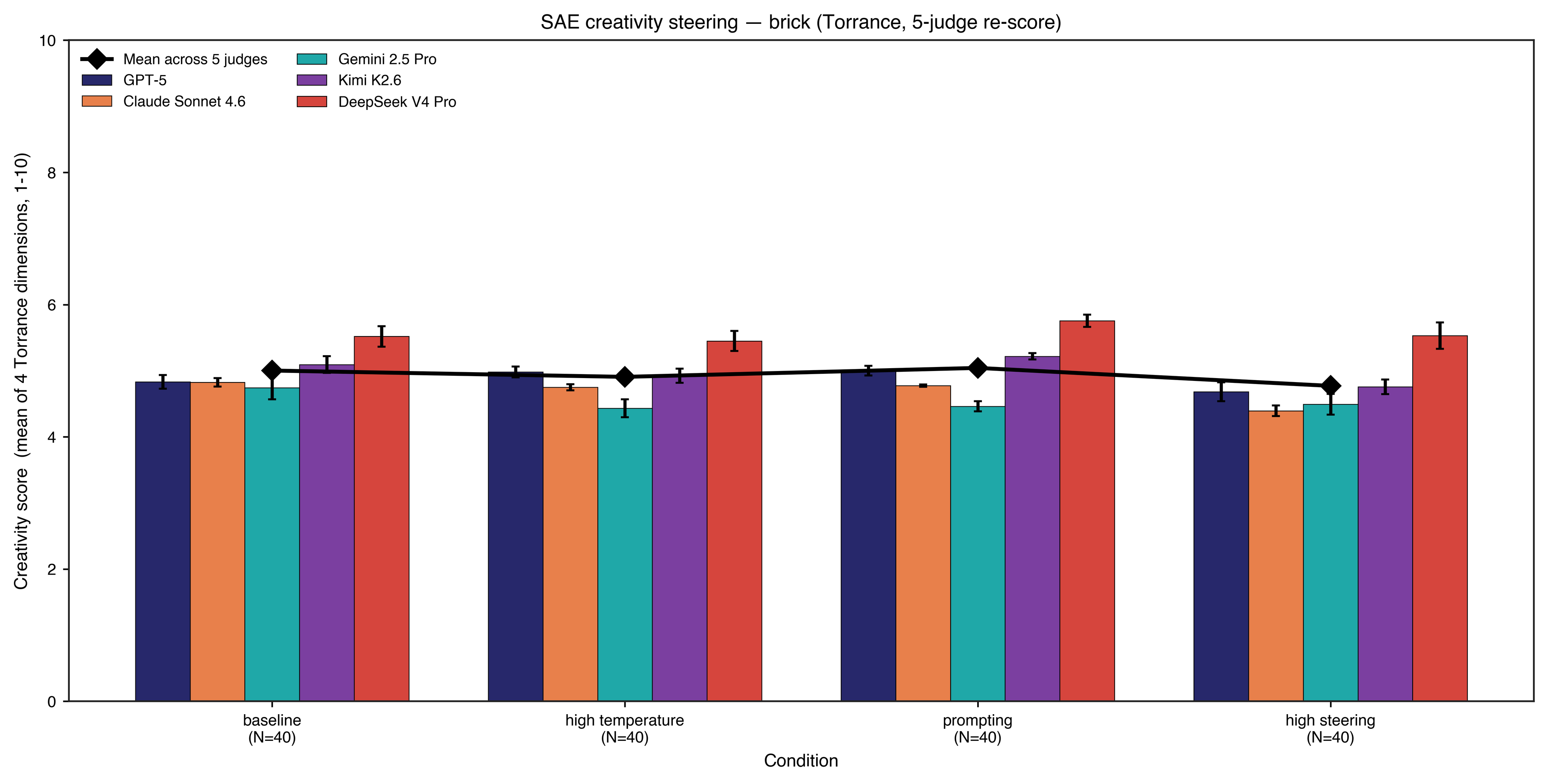}
        \caption{Brick.}
        \label{fig:og-capability-brick}
    \end{subfigure}
    \hfill
    \begin{subfigure}[b]{0.48\linewidth}
        \centering
        \includegraphics[width=\linewidth]{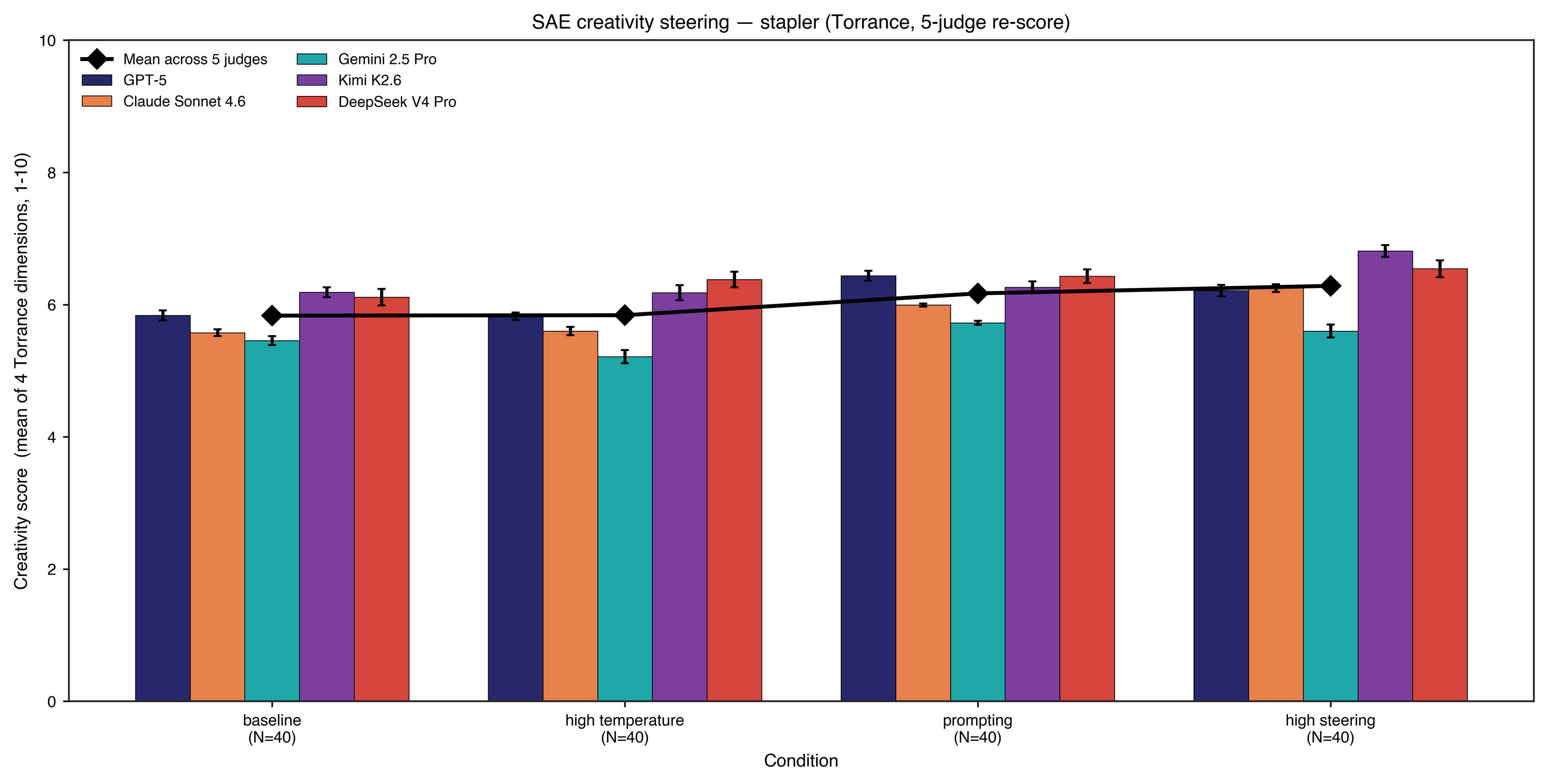}
        \caption{Stapler.}
        \label{fig:og-capability-stapler}
    \end{subfigure}
    \caption{Capability tasks scored by five independent LLM judges.}
    \label{fig:og-capability}
\end{figure}

We implement the divergent creativity and product innovation tasks on generative agents and evaluate their outputs with five independent LLM judges (GPT-5 \citep{OpenAI2025}, Claude Sonnet 4.6, Gemini 2.5 Pro, Kimi K2.6, DeepSeek V4 Pro) under the blind, length-controlled rubric described in \S\ref{subsec:eval_protocol}, scoring along fluency, flexibility, originality, and elaboration \citep{Zhao2025}. Figure~\ref{fig:og-capability} reports the mean creativity score across the five judges per condition. On the brick (divergent creativity) task, SAE steering does not improve over baseline ($\Delta=-0.23$); persona prompting ($\Delta=+0.04$) and high-temperature decoding ($\Delta=-0.10$) leave the brick mean essentially flat. On the stapler (product innovation) task, SAE steering produces a small positive shift ($\Delta=+0.45$); persona prompting gives a smaller gain ($\Delta=+0.34$) and high-temperature decoding is flat. The five judges agree closely on relative ordering (mean pairwise Spearman across judges $=0.77$; see Appendix~\ref{subsec:multi_judge} for per-judge breakdowns), so the qualitative pattern does not depend on any single judge's preferences.

\begin{figure}[H]
    \centering
    \begin{subfigure}[b]{0.48\linewidth}
        \centering
        \includegraphics[width=\linewidth]{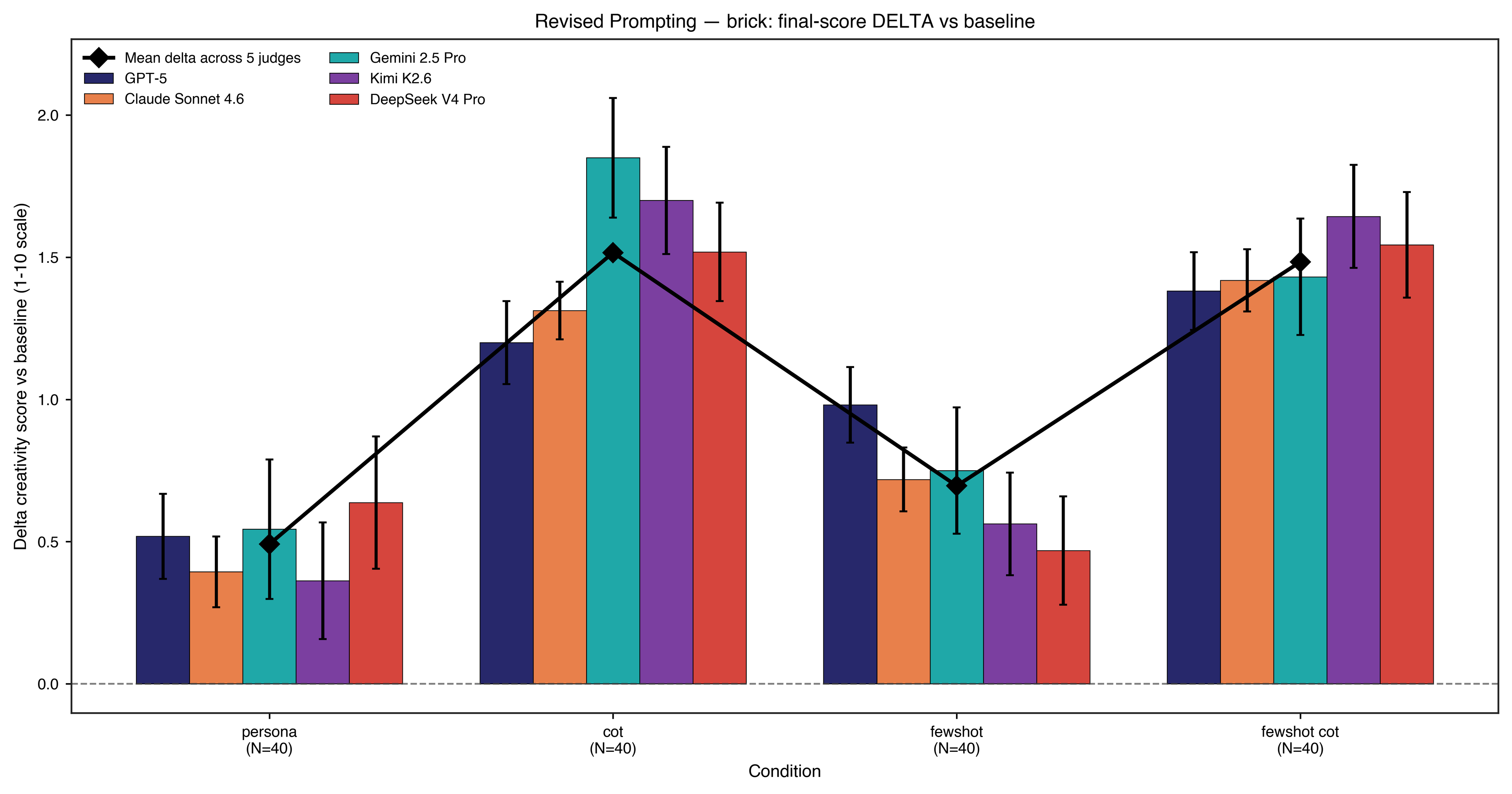}
        \caption{Brick.}
        \label{fig:rp_brick_mj_delta}
    \end{subfigure}
    \hfill
    \begin{subfigure}[b]{0.48\linewidth}
        \centering
        \includegraphics[width=\linewidth]{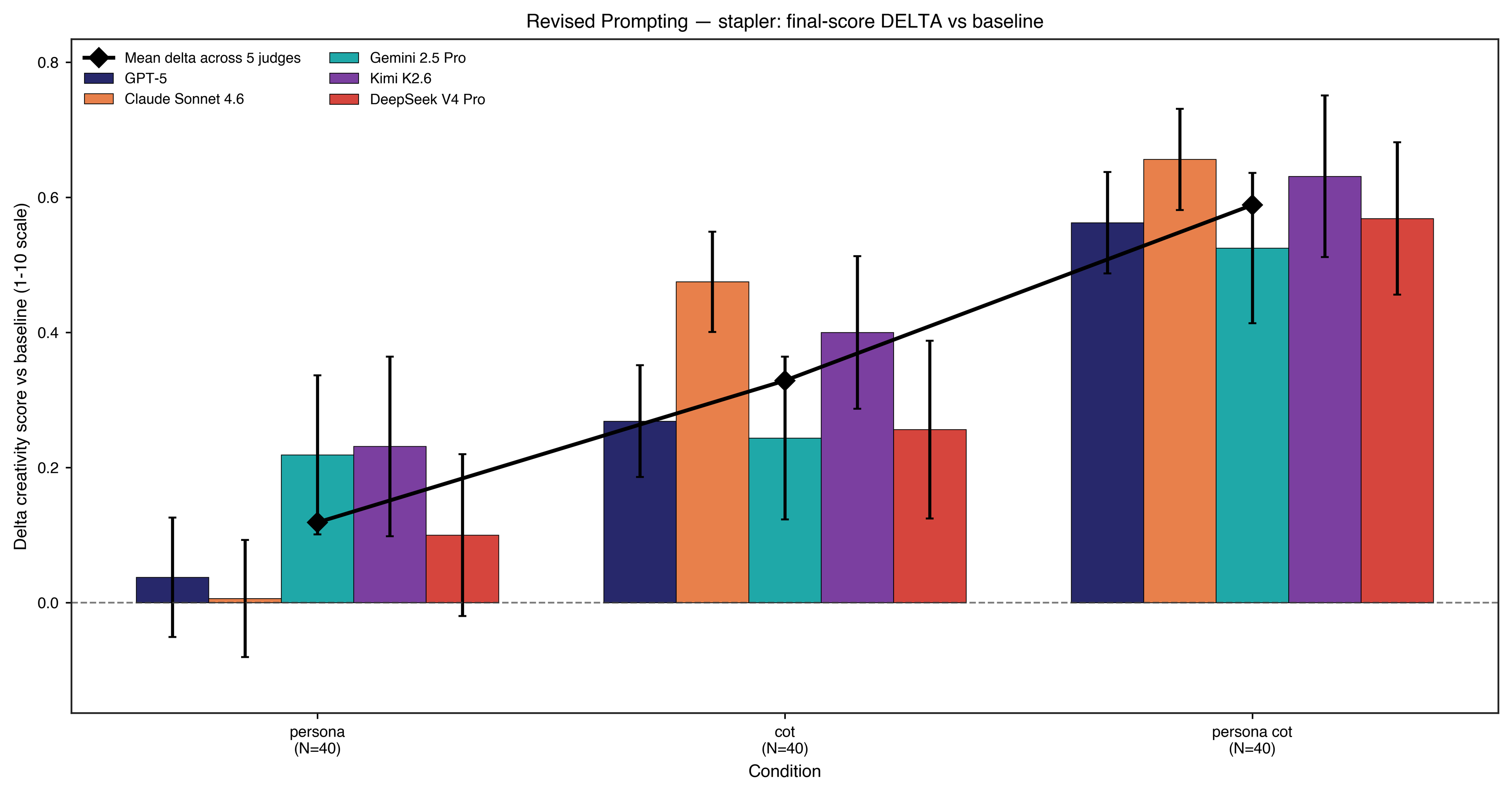}
        \caption{Stapler.}
        \label{fig:rp_stapler_mj_delta}
    \end{subfigure}
    \caption{Per-condition $\Delta$ creativity score vs.\ baseline under the prompting ladder.}
    \label{fig:rp_creativity_mj_delta}
\end{figure}

For comparison against stronger prompting, we score the prompting-ladder responses on the same rubric and judge panel (Figure~\ref{fig:rp_creativity_mj_delta}). On brick, where Ocsai \citep{Organisciak2023} supplies human-rated exemplars, the prompting ladder produces substantially larger deltas than SAE steering: zero-shot CoT moves the mean by $\Delta=+1.52$ and few-shot+CoT by $\Delta=+1.48$, with persona alone giving $\Delta=+0.49$. On stapler, where no public human-rated product-improvement dataset exists and the ladder is therefore limited to persona, CoT, and persona+CoT, persona+CoT moves the mean by $\Delta=+0.59$. Strong prompting therefore outperforms SAE steering on brick and slightly exceeds it on stapler. The contribution of SAE/probe methods on the capability tasks is therefore not behavioral dominance over prompting but the interpretability and calibrated linear control reported in \S\ref{subsec:multi_judge} and below.

\begin{figure}[H]
    \centering
    \includegraphics[width=1\linewidth]{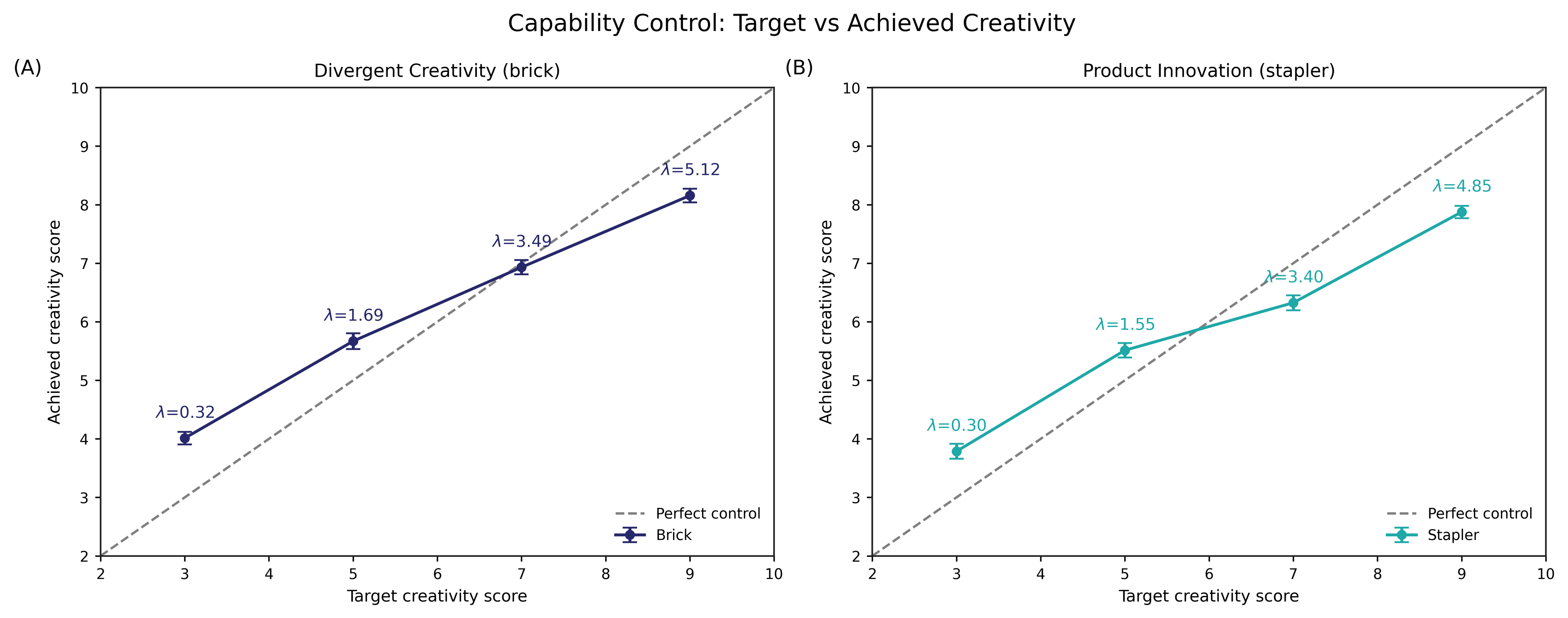}
    \caption{In-distribution capability control for two probes on divergent creativity (left) and product innovation(right). }
    \label{fig:capability_control}
\end{figure}

We then turn to probes to see whether they achieve more reliable performance. The left panel of Figure~\ref{fig:capability_control} presents controllability results for divergent creativity using the brick prompt via our probe-based methods. The figure shows the achieved creativity score, scored by the GPT-5 judge, as a function of the target score. Error bars represent the standard error across agents. The diagonal line indicates perfect controllability. The probe achieves meaningful control, shifting the mean achieved score from approximately 4.0 at a target of 3 to approximately 8.2 at a target of 9, though achieved scores consistently fall below the perfect-control diagonal. This gap between target and achieved scores widens at higher targets, suggesting that while probe-based steering captures a substantial portion of the creativity dimension, it cannot fully saturate the trait. The right panel of Figure~\ref{fig:capability_control}, showing how probes steer the creativity of product innovation tasks, demonstrates similar results to those for divergent creativity. We also examine whether probes trained on one object category can steer creativity for out-of-sample objects. Specifically, the divergent-creativity probe is trained on brick-only labeled data and then applied as a fixed steering direction across four object prompts: brick (in-distribution) and stapler, paperclip, and bowl (out-of-sample). The results, reported in Appendix Figure~\ref{fig:capability_control_full}, show that this transfer-based implementation is both feasible and effective.

Compared to the preference tasks, capability control through probes shows a more compressed range relative to the target scale, despite relatively reliable compared to SAE methods. The achieved-versus-target slope is less than unity, indicating that each unit increase in the target yields a diminishing return in achieved creativity. This partial controllability is consistent with the view that creativity resists compression into a single linear direction, as it involves coordinated changes across fluency, flexibility, originality, and elaboration.

%% file: tab-fig/interventions.tex
\begin{table}[H]
  \centering
  \footnotesize
  \setlength{\tabcolsep}{5pt}
  \renewcommand{\arraystretch}{1.2}
  \begin{tabular}{p{3cm} >{\centering\arraybackslash}p{1.2cm} p{2.5cm} p{5cm} >{\centering\arraybackslash}p{1.8cm}}
    \toprule
    \multirow{2}{*}{Task Type} & \multicolumn{4}{c}{Intervention / Setting} \\
    \cmidrule(lr){2-5}
    & Baseline & Prompting & Steering & Temperature\\
    \midrule
    Lottery Game & --- & Risky persona & 2 risk-related SAE features boosted & 0.5\\[3pt]
    Ultimatum Game & --- & Altruistic persona & 1 altruism SAE feature boosted & 0.5\\[3pt]
    Divergent Creativity & --- & Creative persona & 3 creativity SAE features boosted & 1.0 \\[3pt]
    Product Innovation & --- & Creative persona & 3 creativity SAE features boosted & 1.0 \\
    \bottomrule
  \end{tabular}
  \caption{Interventions used for preference vs.\ capability tasks.}
  \label{tab:interventions}
\end{table}

%% file: sections/discussion.tex
Generative agents powered by LLMs give computational social scientists a scalable, low-cost way to simulate behavior, but their usefulness is limited by the black-box nature of LLM reasoning. We show that interpretability techniques, i.e. sparse autoencoders (SAEs) and linear probes can open this black box by identifying and steering behaviorally meaningful features. Across two primitives of human behavior (preferences, captured by lottery and ultimatum games; capabilities, captured by divergent creativity and product innovation), the interpretability-driven methods produced more interpretable and more controllable agent behavior than prompting alone. SAEs revealed the human-readable features driving each response, while probes provided a calibrated linear knob for shifting agents along a pre-specified trait. Together, SAEs and probes support both inductive theorizing and deductive hypothesis testing in social simulations.

To translate these results into actionable methodological choices, we first compare these approaches on their strengths and trade-offs, and then turn to the boundary conditions and limitations that shape how they should be applied in practice.

First, prompt-based methods have drawbacks of not fully capturing the underlying behavioral mechanisms; even when stronger prompting strategies match SAE/probe steering on the observed behavior, they remain less effective at revealing or modulating the mechanisms behind that behavior. This is because generative agents are auto-regressive in nature, which make them justify what they did than showcase their actual rationale behind behaviors. This indicates that the behaviors we observe from prompts reflect narrative explanations, not the underlying processes that produced those actions, limiting prompting’s inferential validity. In addition, human instructions may not align with the way agents interpret and represent those instructions internally. This misalignment is two-fold: one is on the misunderstanding of natural language and another is on the structural deficiency of agents to encode instructions from humans.

Second, SAE-based methods achieve better controllability than probe-based methods in multi-feature represented behaviors. For example in our creativity tasks, SAE-based methods can capture the fluency, flexibility, originality, and elaboration aspects and coin them together to compose creativity. While probe-based methods achieve meaningful control over creativity (shifting the GPT-5 achieved score from approximately 4.0 to 8.2), the achieved-versus-target slope remains below unity, indicating that the full range of the trait is not accessible through a single linear direction.

Third, the probe approach offers computational advantages: training requires only hundreds of examples rather than the massive datasets needed for SAEs, and inference adds minimal overhead compared to SAE encoding/decoding. Probes provide reliable behavioral control when the target trait aligns with a dominant linear direction in activation space. For complex capabilities such as creativity, probes still achieve partial controllability but with diminishing returns at extreme target values, consistent with the view that such traits involve coordinated changes across multiple semantic dimensions.

These findings suggest some practical guidance for method selection: social scientists should choose the methods depending on the goals of their research. If the goal of research is to inductively investigate hypotheses behind a complex real-world phenomenon, then SAE-based mechanistic interpretability tools will uncover the internal links behind behavioral codes. If deductive experiments on pre-defined measurable traits are the targets, probe-based methods provide more fine-grained controllability efficiently. Together, SAEs and probes serve for a full-cycle inductive and deductive pipeline for using LLM agents in social simulations: researchers can use SAE features to inspect the human-interpretable features activated when an interesting phenomenon is observed in LLM agents, after which probes can be used to experimentally test how a specified trait affects that phenomenon.

Beyond these empirical findings, we also contribute a comparative framework for generative agent steering by systematically contrasting prompting, probes, and SAEs in terms of their strengths, limitations, and best-use cases. This framework, as demonstrated in Table~\ref{tab:method_comparison}, offers practical guidance for researchers in selecting the intervention method that best fits their theoretical aim, whether rapid prototyping, deductive hypothesis testing, or inductive discovery of behavioral mechanisms.

\begin{table}[H]
\centering
\footnotesize
\renewcommand{\arraystretch}{1.25}
\setlength{\tabcolsep}{5pt}

\begin{tabularx}{\textwidth}{
  >{\raggedright\arraybackslash}p{2.1cm}
  >{\raggedright\arraybackslash}X
  >{\raggedright\arraybackslash}X
  >{\raggedright\arraybackslash}p{3.2cm}
}
\toprule
\textbf{Method} & \textbf{Strengths} & \textbf{Limitations} & \textbf{Best Suited For} \\
\midrule

Prompting &
No training required; easy to deploy at scale &
Does not capture underlying mechanisms; susceptible to post-hoc narrative rationalization; may misinterpret ambiguous natural language &
Fast prototyping; scalable batch experiments \\[6pt]

Probes &
Precise calibration of target dimensions; computationally efficient; trainable from small labeled datasets &
Steering effect can saturate at extremes; requires alignment with a pre-specified direction in activation space &
Deductive hypothesis testing on predefined, measurable traits \\[6pt]

Sparse Autoencoders (SAEs) &
High interpretability of learned features; handles representational complexity; can reveal behaviorally meaningful latent features &
Computationally expensive to train; requires large-scale activation datasets; risk of entangled feature representations &
Inductive feature discovery; steering complex, multidimensional behavioral traits \\

\bottomrule
\end{tabularx}

\caption{Comparative Overview of Interpretability Methods for Generative Agent Steering}
\label{tab:method_comparison}
\end{table}

While this paper provides one of the first attempts to use interpretability tools in social experiment simulations, the comparisons above also clarify several limitations that constrain what these methods can currently guarantee.

First, misuse of SAE-based methods can introduce unintended and potentially detrimental side effects. The extent to which we should steer an SAE feature to create a desired LLM persona for experimentation is more art than standardized practice. Steering too much will lead to hallucination, which produces nonsensical output. Furthermore, as SAE features are not perfectly disentangled, boosting one feature too much may activate other off-target representations, causing behavioral shift that are hard to analyze. These risks highlight the need for careful monitoring and calibration while using steering in simulating experiments.

Second, we limit our focus to individuals in this paper. While individuals are the most fundamental component of society, many experiments do not happen on the individual level. This leads to a natural thread of future work in multi-agent settings, where generative agents will interact with others in structured social environments, allowing emerging group dynamics to be studied. Such extensions would explore organizational structure, social influence, and network effects, which are only possible to be accessed in multi-agent settings. This is crucial for advancing generative agent simulations towards a richer and more ecologically valid theorizing process.

Third, our experiments focus on a small set of traits developed from preference and capability tasks. They map directly to behavioral constructs. However, many social scientific mechanisms are more abstract, context-dependent, or embedded within complex structures. Future research in using SAE features to capture higher-order mechanisms need to examine the mechanistic support carefully in order to produce robust theorizing results.

Future work might also explore hybrid approaches, such as multi-probe steering or low-rank adaptations, that balance computational efficiency with representational capacity. We hope this paper can serve both as a primer on interpretable agent steering and as a roadmap for using LLM-based personas to accelerate discovery in the social sciences through computational experimentation.

%% file: sections/appendix_sae.tex
\subsection{Lottery Game Implementation}

\subsubsection{Stack Overview}

\textbf{EDSL orchestration.} We use Expected Parrot's EDSL to define each game as a typed interaction with strict I/O schemas. In the safe-vs-risky game we use:
\begin{itemize}
    \item \texttt{QuestionMultipleChoice} with options \texttt{["Safe Option", "Risky Option"]}
    \item An attached \texttt{*\_comment} field to capture the model's rationale
    \item Built-in validators so malformed outputs are reprompted consistently
\end{itemize}

In code (from \texttt{SafeRiskyChoice.setup()}):

\begin{verbatim}
choice_text = (
    "SAFE VS. RISKY CHOICE GAME\n\n"
    "You must choose between two options:\n"
    f"1. Safe Option: Guaranteed 50 tokens.\n"
    f"2. Risky Option: 50% chance of {self.risky_reward} tokens,
       50% chance of 0 tokens.\n\n"
    "Which option do you choose?"
)
self.questions.append(
    QuestionMultipleChoice(
        question_name="safe_risky_choice",
        question_text=choice_text,
        question_options=["Safe Option", "Risky Option"],
    )
)
\end{verbatim}

\textbf{Goodfire interpretability.} We use Goodfire's OpenAI-compatible API to:
\begin{enumerate}
    \item call \texttt{meta-llama/Llama-3.3-70B-Instruct} as the base model
    \item read SAE feature activations on demand
    \item apply feature-level ``nudge'' steering during decoding
\end{enumerate}

We instantiate both the EDSL model and a Goodfire Variant:

\begin{verbatim}
import goodfire
client = goodfire.Client(os.getenv("GOODFIRE_API_KEY"))
base_variant = goodfire.Variant("meta-llama/Llama-3.3-70B-Instruct")

# EDSL model routed through Goodfire
model = Model("meta-llama/Llama-3.3-70B-Instruct",
              service_name="goodfire")
\end{verbatim}

\textbf{EDSL $\leftrightarrow$ Goodfire bridge.} The bridge maps EDSL's \texttt{.parameters["controller"]} to Goodfire's controller schema. When empty, we run the base model. When populated, Goodfire applies feature-level steering. We reset the Variant when turning steering off to avoid bleed-through.

\subsubsection{Agents, Personas, and Cohorts}

We run three cohorts that differ only in how behavior is induced:

\textbf{Baseline agents:} neutral personas

\begin{verbatim}
dummy_agents = AgentList([
    Agent(name=f"Agent_{i}",
          traits={"subject_id": f"A{i}"},
          instruction="")
    for i, age in enumerate(range(20, 60, 1), start=1)
])
\end{verbatim}

\textbf{Prompting agents:} natural-language trait aligned to ``slight risk preference''

\begin{verbatim}
slightly_risk_agents = AgentList([
    Agent(name=f"SlightlyRisk_Agent_{i}",
          traits={"subject_id": f"SR{i}",
                  "personality": "slightly likes to take risks"},
          instruction="")
    for i, age in enumerate(range(20, 60, 1), start=1)
])
\end{verbatim}

\textbf{SAE-based steering agents:} same as baseline (neutral persona), but we attach a Goodfire controller at decode time (details in \S\ref{subsec:steering}).

All cohorts share identical decoding and validation settings. Caching is disabled for runs that collect activations (\texttt{cache=False}) to ensure fresh generations.

\subsubsection{Scenarios and How We Implement Them}

\paragraph{* Baseline}

\textbf{Meaning.} Measure unmodified model behavior under our EDSL scaffold.

\textbf{Implementation.}
\begin{itemize}
    \item Persona: neutral (baseline agents)
    \item Controller: unset
\end{itemize}

\begin{verbatim}
self.model.parameters["controller"] = {}
self.variant = goodfire.Variant(
    "meta-llama/Llama-3.3-70B-Instruct")
\end{verbatim}

\begin{itemize}
    \item Run: \texttt{Survey(self.questions)}\\
    \texttt{.by(self.agents).by(self.model)}\\
    \texttt{.run(cache=False)}
\end{itemize}

\paragraph{* Prompting}

\textbf{Meaning.} Induce similar behavior using only natural-language persona traits.

\textbf{Implementation.}
\begin{itemize}
    \item Persona: ``slightly likes to take risks'' (see \texttt{slightly\_risk\_agents})
    \item Controller: unset (no steering)
    \item Everything else identical to baseline
\end{itemize}

\paragraph{* SAE-based Steering}
\label{subsec:steering}

\textbf{Meaning.} Induce behavior by nudging human-readable SAE features during decoding, while keeping the persona neutral.

\textbf{Feature selection.} We shortlist risk-related features from Goodfire's 65k SAE library based on label semantics and example firings. In the notebook we use features such as:

Lottery Game:
\begin{itemize}
    \item ``Willing to take risks or make sacrifices for a goal'' (\texttt{index\_in\_sae=184})
    \item ``Executing potentially risky operations that require caution'' (\texttt{index\_in\_sae=4237})
\end{itemize}

\textbf{Controller config.} We define two presets:
\begin{itemize}
    \item Standard (\texttt{get\_intervention}): stronger nudges (e.g., value 0.5 to 0.7)
    \item Lite (\texttt{get\_lite\_intervention}): smaller magnitudes (e.g., value 0.4 to 0.6)
\end{itemize}

Schema shape used by the bridge:

\begin{verbatim}
controller = {
  "interventions": [
    {
      "mode": "nudge",
      "features": {
        "features": [
          {"uuid": "...",
           "label": "...",
           "index_in_sae": 184,
           "max_activation_strength": 1},
          {"uuid": "...",
           "label": "...",
           "index_in_sae": 4237,
           "max_activation_strength": 1},
        ]
      },
      "value": 0.5  # signed magnitude
    },
    # stack multiple nudges if needed
  ],
  "scopes": [],
  "name": "controller__...",
  "nonzero_strength_threshold": None
}
self.model.parameters["controller"] = \
    controller
\end{verbatim}

\subsection{Ultimatum Game Implementation}

The ultimatum game follows the same three-scenario structure as the lottery game (Baseline, Prompting, Steering) to measure altruism and fairness preferences.

\subsubsection{Task Setup}

In code (from \texttt{UltimatumGame.setup()}):

\begin{verbatim}
responder_text = (
    "ULTIMATUM GAME - RESPONDER\n\n"
    "You are participating in an economic decision-making "
    "experiment called the Ultimatum Game.\n\n"
    "Situation:\n"
    "1. Another participant (the proposer) was given 100 tokens "
    "(which have real monetary value).\n"
    f"2. The proposer has decided to offer you {self.offer_amount} "
    "tokens out of their 100 tokens.\n"
    f"3. If you accept this offer, you will receive "
    f"{self.offer_amount} tokens, and the proposer will keep "
    f"{100 - self.offer_amount} tokens.\n"
    "4. If you reject this offer, both of you will receive "
    "0 tokens.\n\n"
    "Do you accept or reject this offer?"
)
self.questions.append(
    QuestionMultipleChoice(
        question_name="ultimatum_response",
        question_text=responder_text,
        question_options=["Accept", "Reject"],
    )
)
\end{verbatim}

\subsubsection{Agent Cohorts}

\textbf{Baseline agents:} neutral personas (same as lottery game)

\begin{verbatim}
dummy_agents = AgentList([
    Agent(name=f"Agent_{i}",
          traits={"subject_id": f"A{i}"},
          instruction="")
    for i, age in enumerate(range(20, 60, 1), start=1)
])
\end{verbatim}

\textbf{Prompting agents:} altruistic/fairness-focused persona

\begin{verbatim}
fairness_agents = AgentList([
    Agent(name=f"Fairness_Agent_{i}",
          traits={"subject_id": f"F{i}",
                  "personality": "Altruistic and selfless "
                                 "behavior or intentions"},
          instruction="")
    for i, age in enumerate(range(20, 60, 1), start=1)
])
\end{verbatim}

\subsubsection{Steering Configuration}

\textbf{Feature selection.} We use an altruism-related feature from Goodfire's SAE library:

\begin{itemize}
    \item ``Altruistic and selfless behavior or intentions'' (\texttt{index\_in\_sae=31935})
\end{itemize}

\textbf{Controller config.} Schema shape for ultimatum game steering:

\begin{verbatim}
controller = {
  "interventions": [
    {
      "mode": "nudge",
      "features": {
        "features": [
          {"uuid": "a239616d4dcf470497b32ad3500f0145",
           "label": "Altruistic and selfless behavior "
                    "or intentions",
           "index_in_sae": 31935,
           "max_activation_strength": 1}
        ]
      },
      "value": 0.5
    }
  ],
  "scopes": [],
  "name": "controller__48491998",
  "nonzero_strength_threshold": None
}
self.model.parameters["controller"] = controller
\end{verbatim}

\subsubsection{Experimental Sweep}

We sweep offer amounts from 10 to 90 tokens:

\begin{verbatim}
run_ultimatum_experiments(
    baseline_agents=dummy_agents,
    fairness_agents=fairness_agents,
    model=model,
    start=10, end=90, step=5
)
\end{verbatim}

For each offer amount we run three experiments:
\begin{itemize}
    \item BASELINE (neutral persona, no controller)
    \item PROMPTING (altruistic persona, no controller)
    \item STEERING (neutral persona, altruism feature boosted)
\end{itemize}

Each run writes one CSV per condition. A summary table tracks:
\begin{itemize}
    \item \texttt{accept\_percent} = fraction of agents choosing ``Accept''
    \item We then plot the three curves across offer amounts
\end{itemize}

\subsection{Capability Task Implementation and Evaluation}
\label{subsec:capability}

For capability tasks (divergent creativity and product innovation), we use the same three-scenario structure (Baseline, Prompting, Steering) but add a fourth scenario with high temperature sampling to test whether randomness enhances creative outputs.

\subsubsection{Task Implementation}

We implement two capability tasks inspired by the Torrance Tests of Creative Thinking:

\begin{itemize}
    \item \textbf{Divergent Creativity}: ``List very detailed ways you can use a brick. Each answer should be a paragraph.''
    \item \textbf{Product Innovation}: ``Your goal is to improve the stapler. List as many specific enhancements as you can that would make it better. You may change features, materials, mechanisms, interfaces, or add/remove parts. Do not list new uses; stay focused on improvements to the object itself. For each idea, add enough detail so someone could build or test it.''
\end{itemize}

Unlike preference tasks which use \texttt{Question\allowbreak MultipleChoice}, capability tasks use \texttt{Question\allowbreak List} with open-ended text generation to allow agents to produce diverse, creative responses (up to 10 items per response).

\subsubsection{Prompting Configuration for Creativity}

For the prompting scenario, we configure agents with creativity-focused persona traits. Rather than using a single natural-language instruction, we assign three trait dimensions that encourage creative thinking:

\begin{verbatim}
prompting_agents = AgentList([
    Agent(name=f"Agent_{i}",
          traits={
              "subject_id": f"P{i}",
              "trait1": "Enabling or empowering creative
                         expression and exploration",
              "trait2": "Descriptions of creative unconventional
                         thinking, especially 'thinking
                         outside the box'",
              "trait3": "Professional innovation and
                         creative problem-solving"
          },
          instruction="")
    for i in range(1, 41)
])
\end{verbatim}

These trait labels were selected to align with the SAE features we identified for steering, enabling a fair comparison between prompting and mechanistic steering approaches.

\subsubsection{SAE Steering Configuration for Creativity}

For the steering scenario, we identify three SAE features from Goodfire's 65k feature library that correspond to creativity-related concepts and boost them during decoding:

\textbf{Feature 1: Creative Expression}
\begin{itemize}
    \item Label: ``Enabling or empowering creative expression and exploration''
    \item SAE Index: 13142
    \item UUID: \texttt{2c83bf952a3a4213b45f098aa8c015e2}
\end{itemize}

\textbf{Feature 2: Unconventional Thinking}
\begin{itemize}
    \item Label: ``Descriptions of creative unconventional thinking, especially `thinking outside the box'\,''
    \item SAE Index: 20117
    \item UUID: \texttt{96f91d8ed8934998b52bf66a5951dafd}
\end{itemize}

\textbf{Feature 3: Professional Innovation}
\begin{itemize}
    \item Label: ``Professional innovation and creative problem-solving''
    \item SAE Index: 4992
    \item UUID: \texttt{c148476f6cdc4cd9a2110c24045bfcbe}
\end{itemize}

The controller configuration for creativity steering:

\begin{verbatim}
controller = {
  "interventions": [
    {
      "mode": "nudge",
      "features": {
        "features": [
          {"uuid": "2c83bf952a3a4213b45f098aa8c015e2",
           "label": "Enabling or empowering creative
                     expression and exploration",
           "index_in_sae": 13142,
           "max_activation_strength": 1}
        ]
      },
      "value": 0.3
    },
    {
      "mode": "nudge",
      "features": {
        "features": [
          {"uuid": "96f91d8ed8934998b52bf66a5951dafd",
           "label": "Descriptions of creative unconventional
                     thinking, especially 'thinking
                     outside the box'",
           "index_in_sae": 20117,
           "max_activation_strength": 1}
        ]
      },
      "value": 0.3
    },
    {
      "mode": "nudge",
      "features": {
        "features": [
          {"uuid": "c148476f6cdc4cd9a2110c24045bfcbe",
           "label": "Professional innovation and
                     creative problem-solving",
           "index_in_sae": 4992,
           "max_activation_strength": 1}
        ]
      },
      "value": 0.3
    }
  ],
  "scopes": [],
  "name": "controller__creativity",
  "nonzero_strength_threshold": None
}
self.model.parameters["controller"] = controller
\end{verbatim}

We use a steering strength of 0.3 for each feature, which was empirically selected to produce meaningful behavioral changes without causing output degradation or hallucination. All three features are boosted simultaneously during decoding to capture multiple dimensions of creative thinking.

\subsubsection{Creativity Evaluation}

To quantitatively assess the creativity of agent responses, we employ a secondary LLM-based evaluation using \texttt{QuestionLinearScale}. The evaluation prompts another instance of the model to rate responses on a 1 to 10 scale based on four established dimensions of creative thinking:

\begin{verbatim}
eval_question = QuestionLinearScale(
  question_name="creativity_score",
  question_text=f"""Rate the creativity of the
following task and its response on a scale of
1-10, where:

The task is: {task_descriptions[task]}

Factors to consider:

1. Fluency: The ability to produce a significant
   number of relevant ideas in response to a
   given question.
2. Flexibility: The variety of categories from
   which one can generate ideas. It's the ability
   to think of alternatives, shift from one class
   or perspective to another, and to approach a
   given problem or task from different angles.
3. Originality: The uniqueness of the ideas
   generated. Original ideas are those that are
   rare or unconventional, differing from the
   norm.
4. Elaboration: The ability to expand upon,
   refine, and embellish an idea. It involves
   adding details, developing nuances, and
   building upon a basic concept to make it more
   intricate or complex.

Response to evaluate:
{response_text}

Rate the creativity (you can be generous):""",
  question_options=[1, 2, 3, 4, 5, 6, 7, 8, 9, 10],
  option_labels={
    1: "Not creative (generic, obvious, "
       "conventional ideas)",
    10: "Very creative (highly original, novel, "
        "unconventional, diverse ideas)"
  }
)
\end{verbatim}

This multi-dimensional evaluation framework aligns with established creativity research \citep{Torrance1966, Guilford1967} and allows us to systematically compare how prompting, steering, and temperature manipulation affect creative output quality.

\subsubsection{Multi-Judge Evaluation Details}
\label{subsec:multi_judge}

Creativity scoring uses a panel of five independent LLM judges spanning four developers: GPT-5 \citep{OpenAI2025}, Claude Sonnet 4.6, Gemini 2.5 Pro, Kimi K2.6, and DeepSeek V4 Pro. Each judge applies the same four-dimension Torrance rubric (fluency, flexibility, originality, elaboration). A panel of multiple judges, rather than a single judge such as GPT-5, is necessary because the base model (Llama-3.3-70B-Instruct) shares much of its pretraining distribution with most frontier judge models, and the creativity probe is trained on machine-generated labels: a single-judge protocol would conflate genuine creativity with one judge's idiosyncratic preferences. Two protocol features prevent known biases. First, scoring is blind: responses are pooled across conditions and presented in randomized order, so that each judge sees only the task and the response and never the condition or the method that produced it. This tests whether a judge scores steered outputs more favorably when it cannot identify them as steered. Second, the rubric is length-controlled: it instructs judges to score idea quality rather than response length, with no generosity prompt to bias scores upward.

\paragraph{Inter-rater agreement.} Across all creativity datasets used in the paper (SAE-steered brick and stapler, prompting-ladder brick, prompting-ladder stapler), $N=680$ unique responses are scored by all five judges. The mean pairwise Spearman correlation across the ten judge pairs is $0.769$. Pairwise values range from $0.675$ (Claude--DeepSeek) to $0.849$ (Gemini--GPT-5); see Appendix Figure~\ref{fig:multijudge_agreement_sae}. Conditional on dataset, agreement is highest on the prompting-ladder brick subset (mean pairwise Spearman $0.68$) and lowest on the prompting-ladder stapler subset ($0.42$). The lower stapler agreement is itself informative: judges agree less when the underlying construct (``what counts as a stapler improvement'') is more open to interpretation than divergent uses of a familiar object.

\paragraph{Length as a confounder.} We regress each judge's score on $\log$(response length in characters) and report the per-judge $R^2$. Length explains a non-trivial share of the variance for some judges: Gemini-2.5-Pro $R^2=0.43$, DeepSeek-V4-Pro $R^2=0.36$, GPT-5 $R^2=0.29$, while Kimi-K2.6 and Claude-Sonnet-4.6 sit at $R^2\approx0.13$. The length-controlled rubric reduces but does not eliminate this dependence; rank ordering of conditions is preserved even if Gemini and DeepSeek are dropped from the mean.

\paragraph{Per-judge breakdowns.} Per-dimension multi-judge panels for the SAE creativity datasets are shown in Appendix Figure~\ref{fig:sae_capability_dim_mj}; per-condition mean-across-judges scores are reported in the main text (Figure~\ref{fig:og-capability}). Per-judge scores, the pairwise Spearman matrix, and the length-regression coefficients are recorded alongside the released artifacts.

\subsubsection{High Temperature Scenario}

For capability tasks only, we add a fourth scenario testing the hypothesis that increased sampling randomness correlates with creative output. We set temperature to a higher value (e.g., 0.9 or 1.0) while keeping all other parameters constant. This scenario helps disentangle whether creativity improvements from SAE steering are simply due to increased output variance or reflect genuine mechanistic changes in idea generation.

\subsection{Feature Activation Analysis}
\label{subsec:activations}

We inspect activated features for any prompt/response pair and write a compact text report:
\begin{itemize}
    \item Directory: \path{safe_risky_choice/results_{YYYYMMDD_HHMMSS}/}
    \item File: \texttt{feature\_activations.txt}
    \item Contents: per response, the user prompt, model response, and top-k features with activation values
\end{itemize}

Implementation sketch (as in \texttt{get\_feature\_activations}):

\begin{verbatim}
inspect_variant = base_variant \
    if use_base_variant else self.variant
inspector = client.feature.inspect(
    messages=[
      {"role":"system","content":"..."},
      {"role":"user",
       "content": user_prompt_text},
      {"role":"assistant",
       "content": model_response_text},
    ],
    model=inspect_variant
)
top = inspector.top(k=10)  # (feature, activation)
\end{verbatim}

We call this for Baseline, Prompting, and Steering. For the baseline, we set \texttt{use\_base\_\allowbreak variant=\allowbreak True} to avoid any stale controller state.

\subsection{Experimental Sweep and Outputs}

We sweep risky-outcome magnitudes:

\begin{verbatim}
run_reward_experiments(
    baseline_agents=dummy_agents,
    slightly_risk_agents=slightly_risk_agents,
    model=model,
    start=10, end=180, step=5,
    analyze_activations=True  # toggled depending on runtime budget
)
\end{verbatim}

For each reward we instantiate and run three experiments:
\begin{itemize}
    \item BASELINE (neutral persona, no controller)
    \item SLIGHTLY PROMPTING (risk-leaning persona, no controller)
    \item STEERING (neutral persona, controller attached)
\end{itemize}

Each run writes one CSV per condition. A summary table is aggregated in-memory to compute:
\begin{itemize}
    \item \texttt{percentage\_risky} = fraction of agents choosing ``Risky Option''
    \item We then plot the three curves across reward magnitudes
\end{itemize}

The helper \texttt{plot\_safe\_risky\_choices()} automatically picks the most recent \texttt{results\_*} folder and renders a comparison plot of the three conditions.

\subsection{Invariants and Controls}

\begin{itemize}
    \item \textbf{Identical decoding and scaffolding across conditions.} Only the persona text (Prompting) or controller (Steering) differ. Temperatures/top-p are held constant within a sweep. Caching is disabled where we collect activations.
    \item \textbf{Schema enforcement.} EDSL validators ensure answers use valid MCQ labels, and rationales are present. One standardized reprompt on violation.
    \item \textbf{No leakage.} The Prompting trait never mentions payoff tables or numeric thresholds. Steering retains the exact prompts as Baseline.
    \item \textbf{Stability checks.} We prefer ``lite'' nudges when strong values saturate. We also check that small weight perturbations do not flip decisions arbitrarily in obviously unstable regimes.
\end{itemize}

\subsection{Reproducibility and Artifacts}

Every run logs:
\begin{itemize}
    \item Timestamped results directory
    \item CSVs per condition and reward
    \item Controller JSON (if steering) with feature UUIDs and weights
    \item \texttt{feature\_activations.txt} with top-k features per response
    \item Prompt columns we rely on for inspection:
    \begin{itemize}
        \item \path{prompt.safe_risky_choice_user_prompt}
        \item \path{generated_tokens.safe_risky_choice_generated_tokens}
        \item \path{answer.safe_risky_choice}
        \item \path{comment.safe_risky_choice_comment}
    \end{itemize}
\end{itemize}

\subsection{Strong Prompting Baselines}
\label{subsec:revised_prompting}

Beyond persona prompting, we evaluate SAE- and probe-based steering against a graded prompting ladder drawn from the standard literature, applied to the same Llama-3.3-70B-Instruct backbone on the same four tasks.

\paragraph{Inference stack.} The prompting-ladder runs use a self-hosted vLLM deployment of \texttt{meta-llama/\allowbreak Llama-3.3-\allowbreak 70B-Instruct} in true bf16 on 2$\times$H100 (80GB) GPUs, rather than the FP8 ``Turbo'' build served by hosted APIs. Decoding parameters are held constant across conditions: temperature 0.7, top-$p$ 1.0, 40 agents per condition. Maximum generation lengths are 512 tokens for preference games and 1{,}536 tokens for creativity tasks.

\paragraph{Condition ladder.} For each task we evaluate:
\begin{itemize}
    \item \texttt{baseline} -- neutral free-text elicitation, no persona.
    \item \texttt{persona} -- one-line persona instruction (the same persona condition used elsewhere in the paper, run on the same bf16 stack for parity).
    \item \texttt{cot} -- zero-shot chain-of-thought trigger \citep{Kojima2022}, single completion with reasoning followed by a \texttt{Final answer:} line that is parsed.
    \item \texttt{fewshot} -- six in-context demonstrations of the task without reasoning \citep{Brown2020, Wei2022}.
    \item \texttt{fewshot\_cot\_dose\{000,025,050,075,100\}} -- six demonstrations with chain-of-thought reasoning, where the fraction $\rho \in \{0, 0.25, 0.5, 0.75, 1.0\}$ of demonstrations exhibits the target disposition (risk-seeking for lottery, accepting for ultimatum, creative for capability tasks). The remaining $(1-\rho)$ exhibit the opposite. This traces a prompting dose-response that can be overlaid on the probe $\lambda$ and SAE $\alpha$ curves.
\end{itemize}

\paragraph{Exemplar provenance.} For divergent creativity (brick alternative uses), the six few-shot exemplars are drawn from the Ocsai Alternative Uses Task corpus \citep{Organisciak2023}, a publicly released MIT-licensed dataset of human-authored responses paired with human-rated originality scores. We rank the brick subset by human originality, deduplicate, drop ungrammatical entries left by Ocsai's object-anonymization, and retain the top $N=6$. Crucially, exemplar selection uses human originality ratings, not LLM-judge scores, which keeps the exemplar pool independent of the downstream Torrance judges and avoids selection--evaluation circularity. For product innovation (stapler enhancements), no comparable public dataset of human-rated product improvements exists, so the few-shot column is intentionally omitted; the strongest stapler prompting condition is therefore persona + plan-then-list CoT. We disclose this asymmetry explicitly because it is one reason SAE/probe steering retains a clearer behavioral edge on stapler than on brick.

%% file: sections/appendix_probes.tex
\subsection{Stack Overview}

\textbf{Software stack.} We implement probe-based steering in PyTorch using HuggingFace Transformers for model inference, and scikit-learn for training $\ell_2$-regularized logistic regression probes.

\textbf{Base model and representation.} Probe-based steering uses a frozen base language model and a linear probe trained on intermediate activations. In our main experiments we use Llama-3.3-70B-Instruct and select a single layer for intervention based on held-out probe performance (layer 48 in our Llama runs). We also replicate the full probe pipeline on Qwen-2-7B-Instruct (layer 17); those results appear in the Qwen replication subsection below.

\textbf{Activation extraction.} To compute probe scores, we run a forward pass and capture the hidden state tensor at the selected layer using a forward hook. We use the final token representation as a compact summary of the prompt context.

\begin{verbatim}
# Sketch of probe score extraction
layer = model.model.layers[layer_idx]
acts = []

def capture_hook(module, input, output):
    hidden = output[0] if isinstance(output, tuple) else output
    acts.append(hidden.detach())
    return output

handle = layer.register_forward_hook(capture_hook)
_ = model(**tokenizer(prompt, return_tensors="pt"))
handle.remove()

last_token = acts[0][0, -1, :].float()
score = torch.dot(last_token, probe_direction_unit)
\end{verbatim}

\textbf{Probe training.} We fit an $\ell_2$-regularized logistic regression probe to predict binary behavioral labels from activations. We standardize activations prior to training and select the regularization strength via cross-validation. We train probes across layers and choose the layer with best held-out discriminative performance. This yields both the steering direction (the probe weight vector) and the intervention site (the best layer).

\subsection{Agents, Cohorts, and Trial Structure}

We implement the appendix cohorts from the main paper in a probe-specific form by treating each agent as an independent decoding instance indexed by a deterministic random seed.

\textbf{Baseline trials.} We sample $n$ agents per task parameter with $\lambda=0$, producing unsteered psychometric curves and probe scores.

\textbf{Steering trials.} We repeat the same prompts and decoding settings but enable a single-layer probe steering hook with a specified $(\lambda,\,\text{direction})$.

\textbf{Deterministic seeding.} For preference tasks, we set the seed as a function of the task parameter index and agent ID so that trial sets are reproducible.

\begin{verbatim}
# Sketch of trial loop and deterministic seeding
for param_idx, param in enumerate(task_parameters):
  for agent_id in range(n_agents):
    seed = seed_base + param_idx * 1000 + agent_id
    torch.manual_seed(seed)
    if torch.cuda.is_available():
      torch.cuda.manual_seed_all(seed)
    prompt = make_prompt(param)
    response = generate(prompt, lambda_value, direction)
\end{verbatim}

\subsection{Tasks, Labels, and Targets}

\textbf{Lottery.} The prompt describes a choice between a safe option (guaranteed 50 tokens, fixed across trials) and a risky option (50\% chance of a variable reward, 50\% chance of 0 tokens). The risky reward is varied to produce a psychometric curve. We define the target behavior as the switching point, the risky reward at which $P(\text{Risky})=0.5$, computed by linear interpolation across reward levels.

\textbf{Ultimatum.} The prompt describes an offer to split 100 tokens, where the model must answer Accept or Reject. The offer amount is varied to produce an acceptance curve. We define the target behavior as the acceptance threshold, the offer size at which $P(\text{Accept})=0.5$.

\textbf{Binary labels for probe training.} For preference tasks, we label each trial by the agent's observed choice (risky vs.\ safe in the lottery game, accept vs.\ reject in the ultimatum game), so that the positive class is the higher-risk (respectively, more-altruistic) behavior. Because a risk-seeking agent takes the gamble even at a low risky reward, i.e.\ it switches early at a low switching point, labeling by the choice keeps these early-switching trials in the high class, rather than inverting them as an above-median split on the switching point would. We discard trials whose choice cannot be parsed and train a logistic-regression probe on the remaining labeled activations.

\subsection{Scenarios and How We Implement Them}

Probe experiments use a fixed prompt template and vary only the steering strength and direction.

\subsubsection{Baseline}

\textbf{Meaning.} Measure unsteered behavior and probe scores under the same decoding settings used for steering.

\textbf{Implementation.}
\begin{itemize}
    \item Steering strength: $\lambda=0$
    \item Direction: unused
    \item Decoding parameters (including temperature) held fixed
\end{itemize}

\textbf{What we log.}
\begin{itemize}
    \item Task parameter (risky reward or offer amount)
    \item Parsed choice (Safe/Risky or Accept/Reject)
    \item Probe score computed on the prompt context
\end{itemize}

\subsubsection{Steering}
\label{subsec:probe_steering}

\textbf{Meaning.} Modify the model's internal state at inference time by adding a scaled probe direction vector to hidden states at a single layer.

\textbf{Intervention hook.} During decoding we register a forward hook on the selected layer and add an intervention of the form $\Delta h = s\,\lambda\,v$, where $v$ is the probe direction scaled to match typical activation magnitudes, $\lambda$ is the steering strength, and $s\in\{+1,-1\}$ encodes the increase versus decrease direction. We apply the intervention only to the final portion of the sequence (the last 20\% of tokens) and scale the intervention linearly from 0.5 to 1.0 over these target positions.

\begin{verbatim}
# Sketch of probe steering at one layer
sign = 1.0 if direction == "increase" else -1.0
start_idx = seq_len - max(1, seq_len // 5)
position_scale = torch.linspace(0.5, 1.0, seq_len - start_idx)
hidden[:, start_idx:, :] += (sign * lambda_value * v_scaled) * \
    position_scale.view(1, -1, 1)
\end{verbatim}

\textbf{Probe scores under steering.} For analysis, we report probe scores both before and after steering. The post-steering score is computed by re-running the probe score extraction on the concatenated prompt and generated response.

\subsection{Lambda Calibration and Dose-Response}

To achieve a desired target behavior (for example, a specified switching point), we solve for the steering strength $\lambda$ using a two-stage search. Figure~\ref{fig:dose_response_lambda} shows the dose-response relationship between target switching points and the required $\lambda$ values for both lottery and ultimatum games.

We first evaluate a coarse grid of $\lambda$ values to identify a promising region, then run a bracketed binary search until the achieved metric is within a fixed tolerance. We cache trial logs for each $(\lambda, \text{direction})$ evaluation to make repeated searches efficient.

\subsection{Internal Probe Dynamics}
Figure~\ref{fig:probe_behavior} analyzes how internal probe scores track target behavior during steering. The probe activation score (y-axis) is a scalar readout computed from the model's internal hidden-state vector at the selected layer using our trained logistic-regression probe. Concretely, for each trial we capture the hidden state $h \in \mathbb{R}^d$ at the intervention layer and compute its signed projection onto the unit-normalized probe direction $\hat{w} = w/||w||_2$: $s = \hat{w}^\top h$. Because the probe is a logistic-regression classifier, this dot product corresponds to the probe's linear decision score up to an additive intercept term, so larger $s$ indicates stronger alignment with the probe's positive-class direction.

For the lottery game (left panel), mean probe scores decrease as the target switching point rises from 40 to 240 tokens, consistent with higher switching points corresponding to more risk-averse behavior. Risky choices consistently receive higher probe scores than safe choices, and this separation is maintained across all target levels. For the ultimatum game (right panel), mean probe scores increase with the target switching point. Accept choices receive higher probe scores than reject choices, and this separation is maintained across target values. In both tasks, probe scores track the intended steering direction while preserving clear separation between choice types, validating that our intervention modulates the probe's decision boundary rather than simply overwhelming it. This confirms that the probe direction captures genuine preference-related variation.

\subsection{Cross-Object Generalization}

To assess whether learned probe directions generalize beyond their training distribution, we implement cross-object generalization experiments. We first train a logistic regression probe on hidden activations from Llama-3.3-70B-Instruct for divergent creativity. The probe takes a single representation layer as input and predicts a binary label (high versus low creativity). For in-distribution evaluation, the probe is trained and tested on examples that all use the same object (for instance, train and test on brick). For cross-object evaluation, the probe is trained on a set of objects and tested on a held-out object that never appears in the training data (for instance, train on brick, stapler, and paperclip and test on bowl). Performance is reported as accuracy averaged over multiple random seeds.

Figure~\ref{fig:capability_control_full} presents the full controllability results for divergent creativity across all four object prompts (brick, stapler, paperclip, bowl). Each curve shows the achieved creativity score as a function of the target score, with error bars representing standard error across 40 agents per condition. The diagonal line indicates perfect controllability. All four objects show a consistent pattern: achieved scores (GPT-5 judge) increase monotonically with target scores, ranging from approximately 3.7 to 8.3, but fall below the diagonal at higher targets. The gap between target and achieved scores widens as the target increases, indicating partial but meaningful control across all objects.

Figure~\ref{fig:crossgen_bar} compares in-distribution and cross-object performance for each test object. The results show that probes maintain reasonable accuracy when tested on held-out objects, though performance consistently drops relative to in-distribution evaluation. Figure~\ref{fig:crossgen_profile} illustrates the magnitude of this performance drop for each object. The degradation ranges from approximately 5 to 6 percentage points depending on the object, suggesting that while probe directions capture substantial generalizable structure, they also encode object-specific patterns that do not transfer perfectly across contexts.

\subsection{Invariants and Controls}

\begin{itemize}
    \item \textbf{Identical prompts and decoding across conditions.} Within a sweep, Baseline and Steering use the same prompt templates and decoding parameters. Only $\lambda$ and direction differ.
    \item \textbf{No prompt leakage of targets.} Prompts do not mention switching points, acceptance thresholds, or desired behaviors. Target values are only used in offline metric computation and $\lambda$ selection.
    \item \textbf{Layer and direction selection.} We select the intervention layer by held-out probe performance and keep it fixed across all steering runs. Steering uses the probe weight vector as the single intervention direction.
    \item \textbf{Seeded sampling.} We set deterministic seeds as a function of parameter index and agent ID so that trial sets are reproducible across reruns.
\end{itemize}

\subsection{Reproducibility and Artifacts}

Every probe run produces:
\begin{itemize}
    \item A probe artifact (pickle) containing the probe weights, bias, layer scores, and selected best layer
    \item JSONL trial logs with fields including task type, task parameter, $\lambda$, direction, decoded response, parsed choice, and probe scores
    \item Cached trial evaluations used by the $\lambda$ solver (one cache file per evaluated $(\lambda, \text{direction})$)
    \item Figure scripts that load cached JSONL logs to produce psychometric curves and dose-response plots
\end{itemize}

\subsection{Robustness Check with Qwen-2-7B-Instruct}

To verify that our probe-based findings are not specific to a single model architecture, we replicate the full probe pipeline on Qwen-2-7B-Instruct. This check is intended to assess the portability of the probe-based intervention, not to provide a full cross-model comparison across all methods. We follow the identical procedure described above: synthetic data generation, layer sweep, probe training, $\lambda$ calibration, and behavioral evaluation. The best-performing layer for Qwen was layer 17, selected via cross-validated sweep with a CV score of 0.99. Held-out classification accuracy on divergent creativity averages approximately 85\% in-distribution across object prompts (see Figure~\ref{fig:qwen_crossgen_bar}), compared to approximately 92\% for Llama on the same construct (Figure~\ref{fig:crossgen_bar}). Both probes substantially exceed chance, confirming that the probe pipeline transfers to a smaller model architecture.

Figure~\ref{fig:qwen_preference_dose} shows the psychometric curves for the lottery and ultimatum games under probe-based steering on Qwen-2-7B-Instruct. Figure~\ref{fig:qwen_dose_response} and Figure\ref{fig:qwen_probe_behavior} shows the internal dynamics of probes for both lottery and ultimatum games. The qualitative patterns are consistent with the Llama results reported in the main text: steering successfully shifts switching points across a wide range, and the dose-response relationship is monotonic. Figure~\ref{fig:qwen_capability_brick} and Figure~\ref{fig:qwen_capability_full} present the capability control results for the brick prompt and all four object prompts, respectively. Figure~\ref{fig:qwen_crossgen_bar} and Figure~\ref{fig:qwen_crossgen_profile} show the overall accuracy of probes achieving the targeted score is high, and out-of-sample steering is also possible. However, as with Llama, a ceiling effect is observed across all objects.

%% file: sections/figures_tables_appendix.tex
\begin{figure}[htbp]
    \centering
    \includegraphics[width=\linewidth]{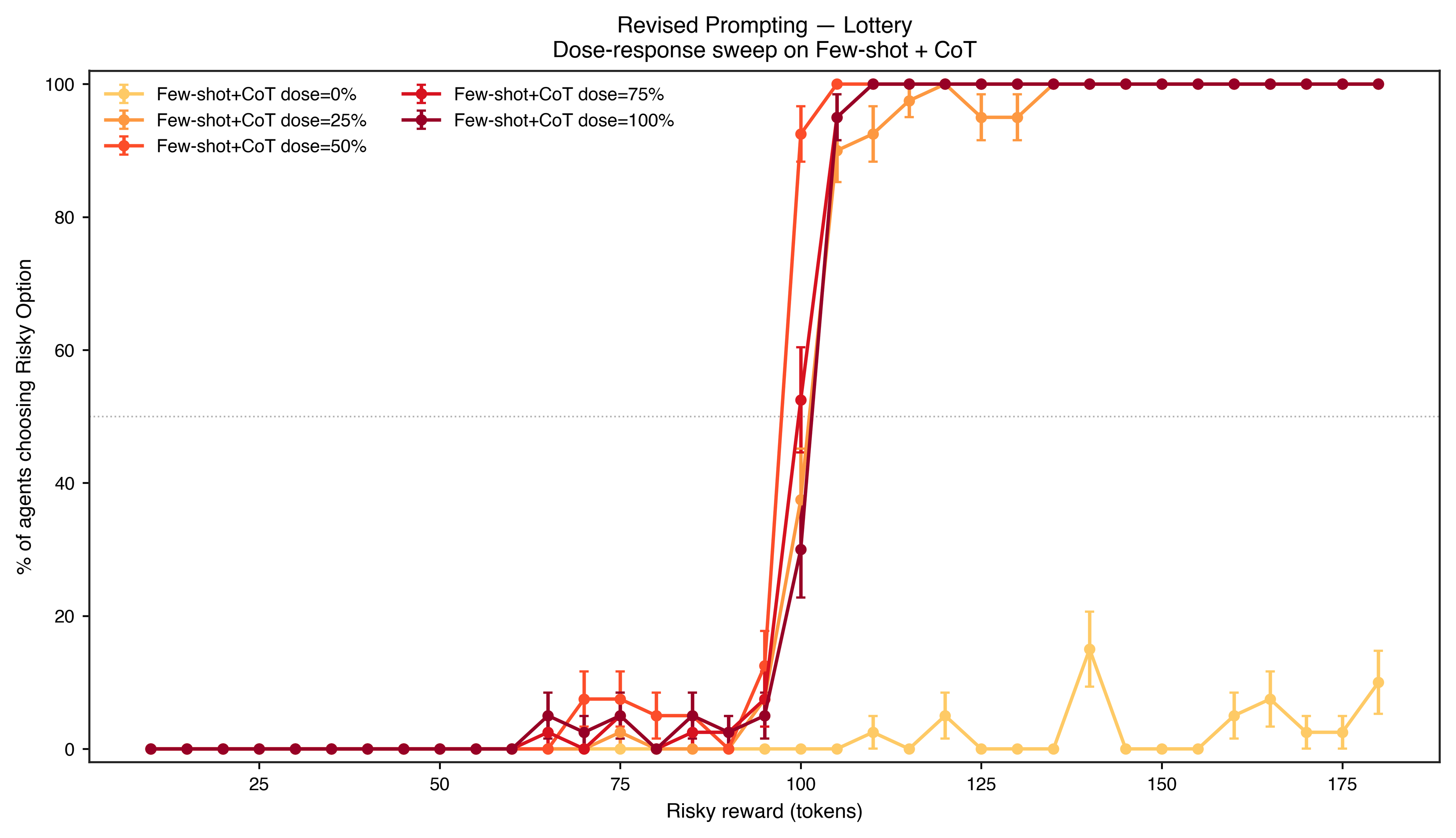}
    \caption{Lottery game: dose-response sweep on \texttt{fewshot\_cot}. Curves correspond to the fraction $\rho \in \{0, 0.25, 0.5, 0.75, 1.0\}$ of demonstrations that take the risky action. Even at $\rho=1.0$, the switch point remains within 97--101 tokens, indistinguishable from baseline; prompting does not provide graded control over risk-seeking. Llama-3.3-70B-Instruct (bf16), 40 agents/condition.}
    \label{fig:rp_lottery_dose}
\end{figure}

\begin{figure}[htbp]
    \centering
    \includegraphics[width=\linewidth]{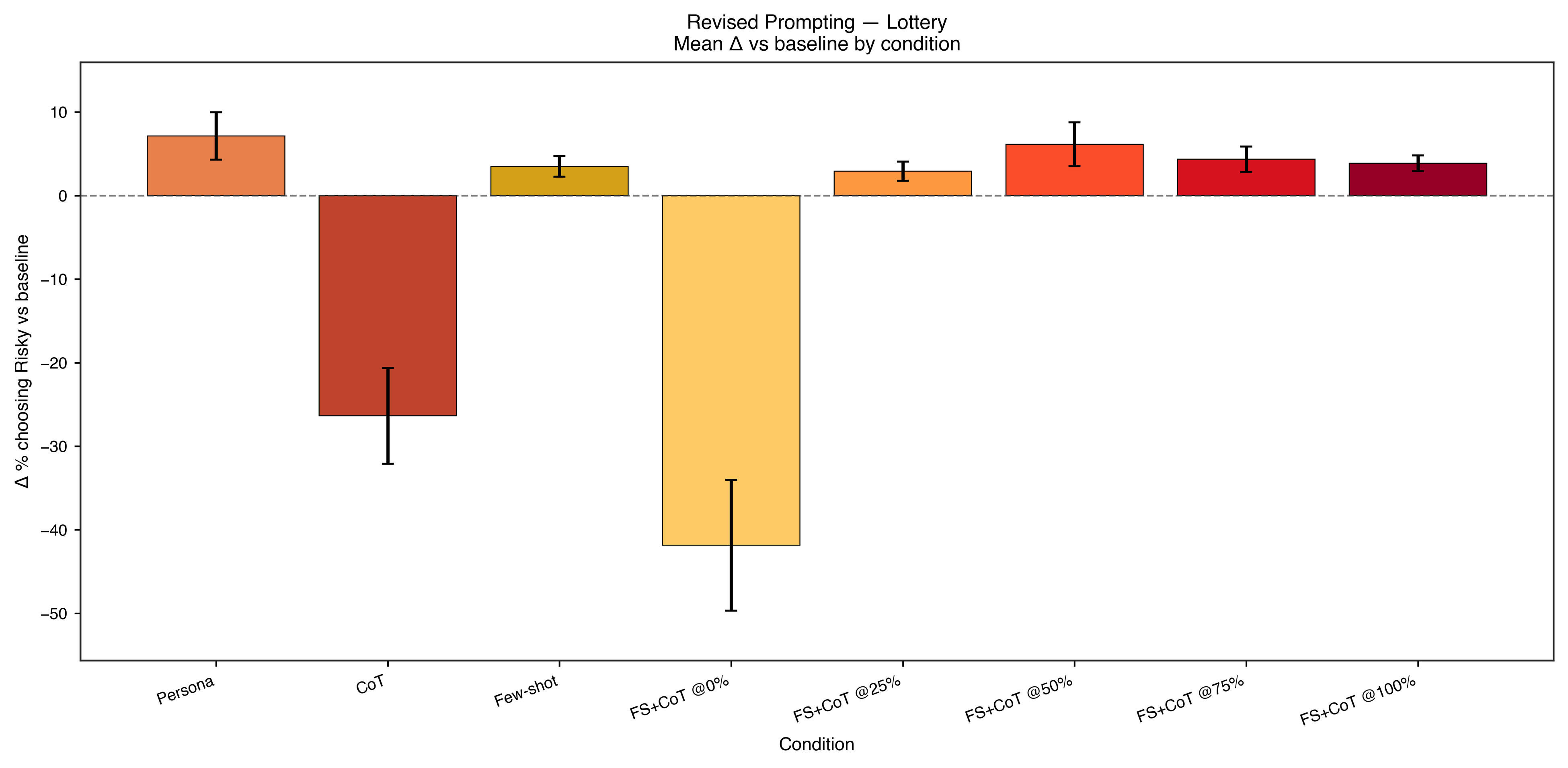}
    \caption{Lottery game: mean $\Delta$ \% choosing Risky vs.\ baseline, averaged across reward levels, per condition. Risk-aversion is inducible (negative $\Delta$ for averse-leaning conditions) but risk-seeking is not (positive doses sit near zero).}
    \label{fig:rp_lottery_delta}
\end{figure}

\begin{figure}[htbp]
    \centering
    \includegraphics[width=\linewidth]{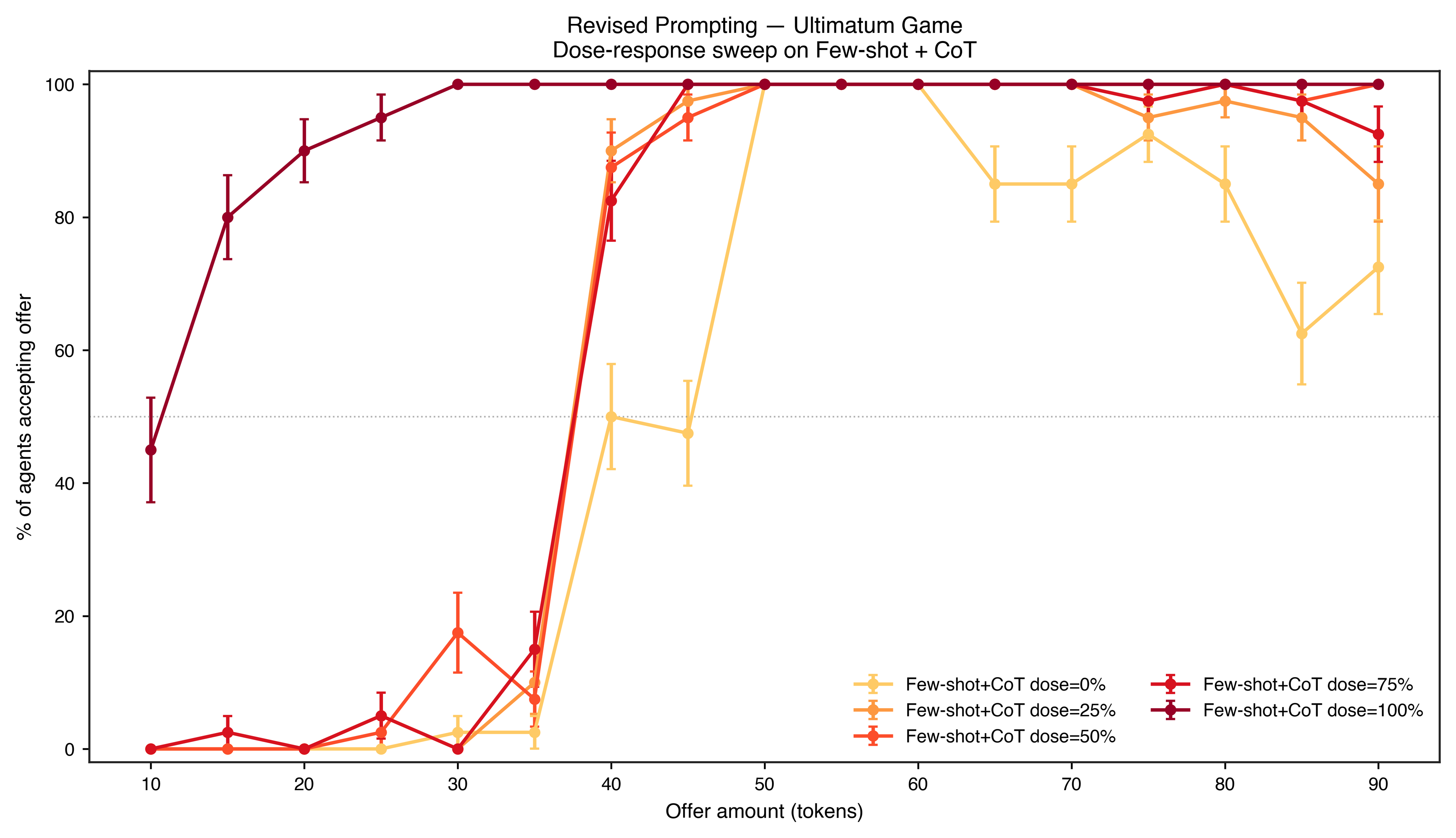}
    \caption{Ultimatum game: dose-response sweep on \texttt{fewshot\_cot}. Higher $\rho$ shifts the acceptance threshold lower, dose-dependently, from $\sim$40 tokens at $\rho=0$ to $\sim$10 tokens at $\rho=1.0$. Llama-3.3-70B-Instruct (bf16), 40 agents/condition.}
    \label{fig:rp_ultimatum_dose}
\end{figure}

\begin{figure}[htbp]
    \centering
    \includegraphics[width=\linewidth]{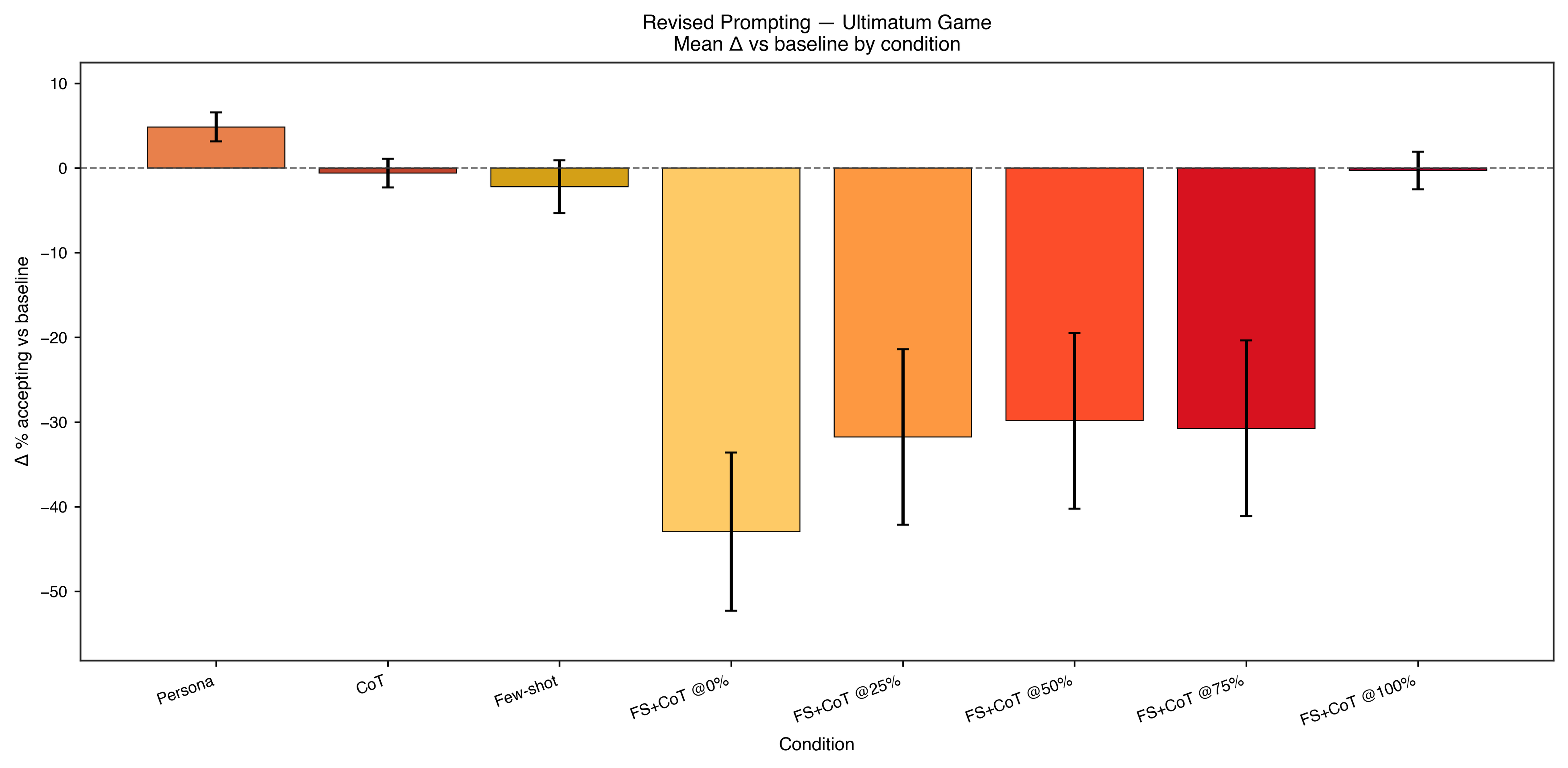}
    \caption{Ultimatum game: mean $\Delta$ \% accepting vs.\ baseline, averaged across offer levels, per condition. Few-shot CoT shows a monotonic dose-response, in contrast to the lottery.}
    \label{fig:rp_ultimatum_delta}
\end{figure}

\begin{figure}[htbp]
    \centering
    \begin{subfigure}[b]{0.48\linewidth}
        \centering
        \includegraphics[width=\linewidth]{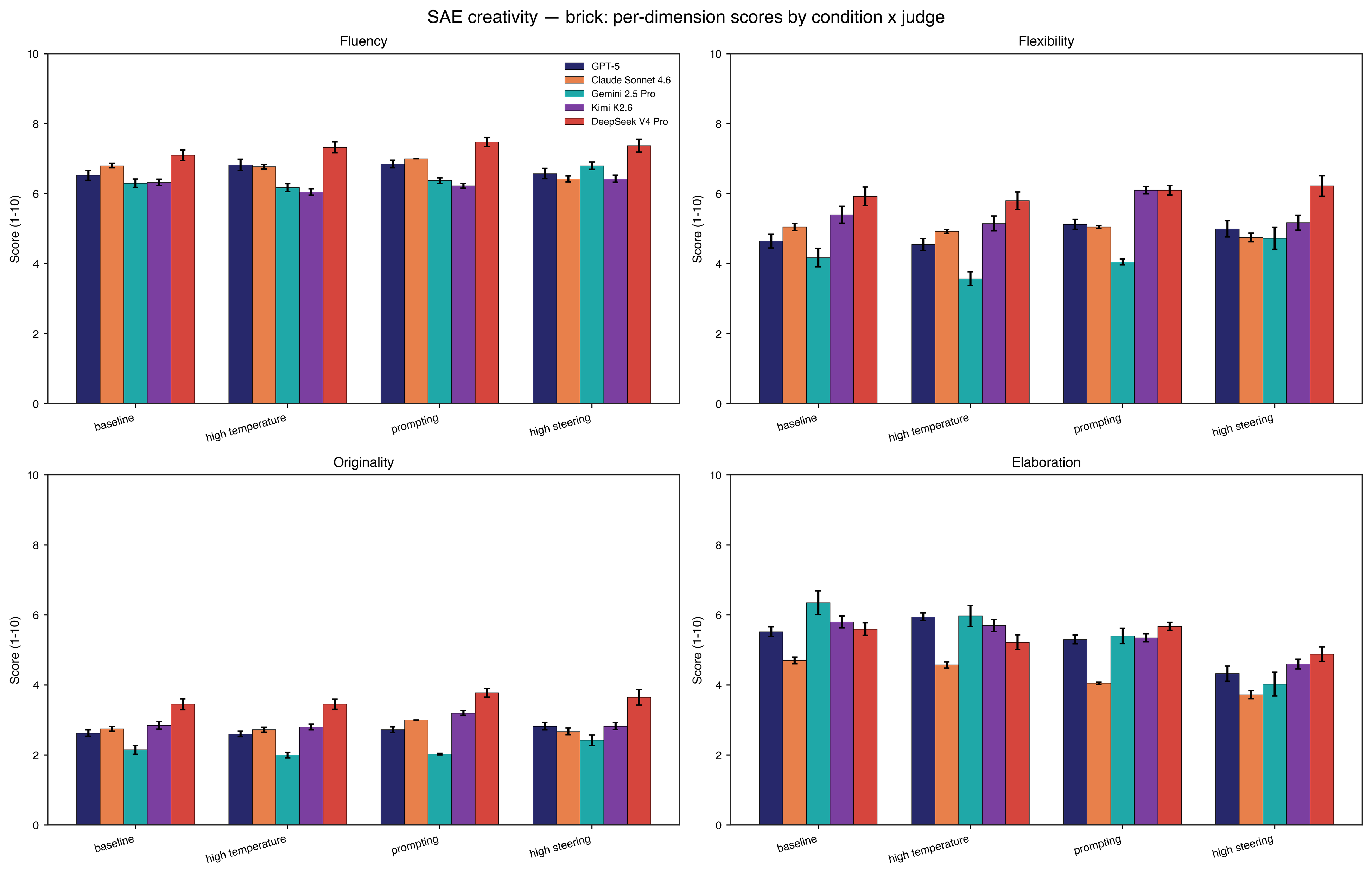}
        \caption{Brick.}
        \label{fig:sae_brick_dim_mj}
    \end{subfigure}
    \hfill
    \begin{subfigure}[b]{0.48\linewidth}
        \centering
        \includegraphics[width=\linewidth]{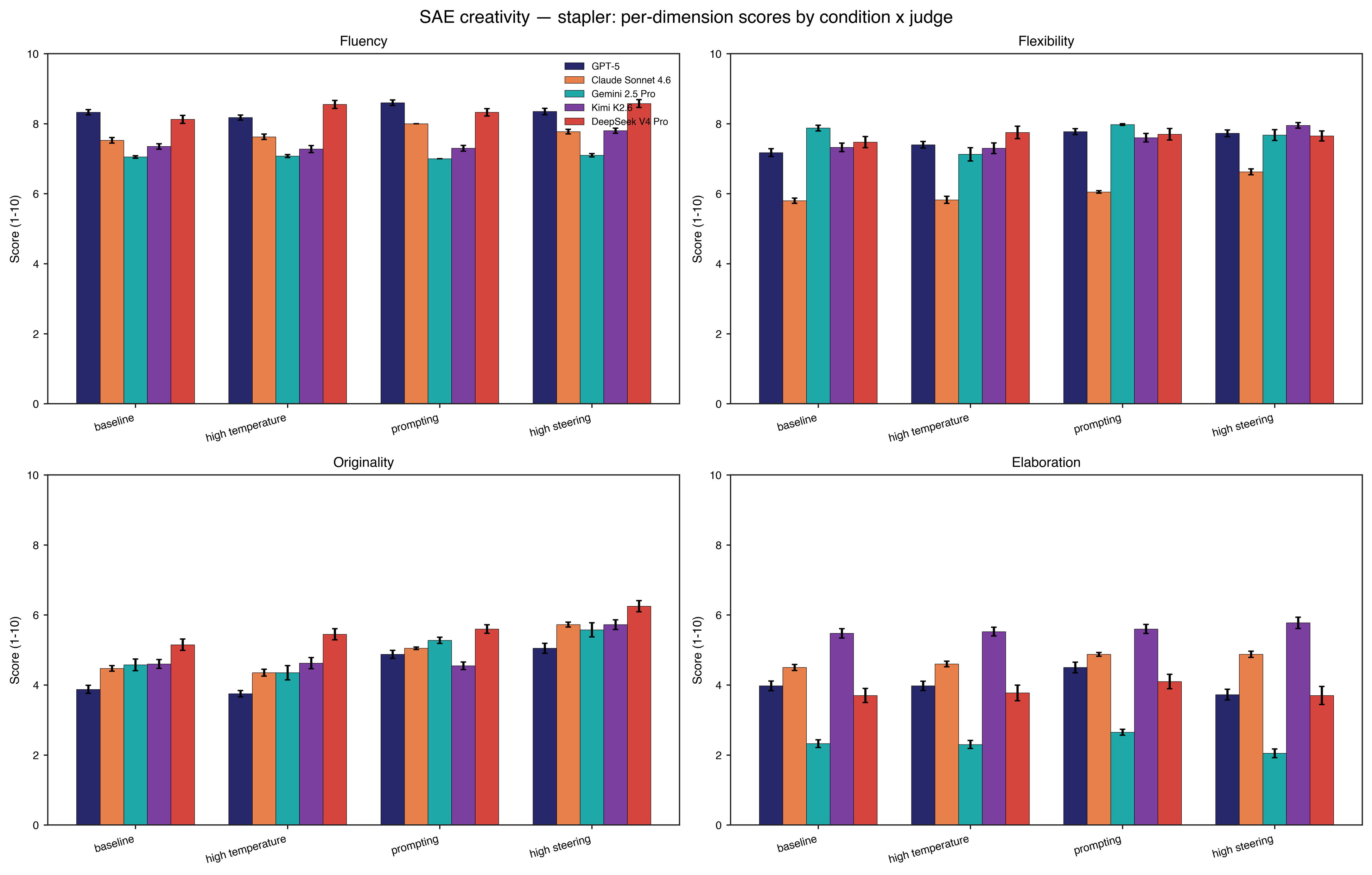}
        \caption{Stapler.}
        \label{fig:sae_stapler_dim_mj}
    \end{subfigure}
    \caption{SAE creativity steering, per-dimension Torrance scores (fluency, flexibility, originality, elaboration) by condition and judge.}
    \label{fig:sae_capability_dim_mj}
\end{figure}

\begin{figure}[htbp]
    \centering
    \includegraphics[width=0.7\linewidth]{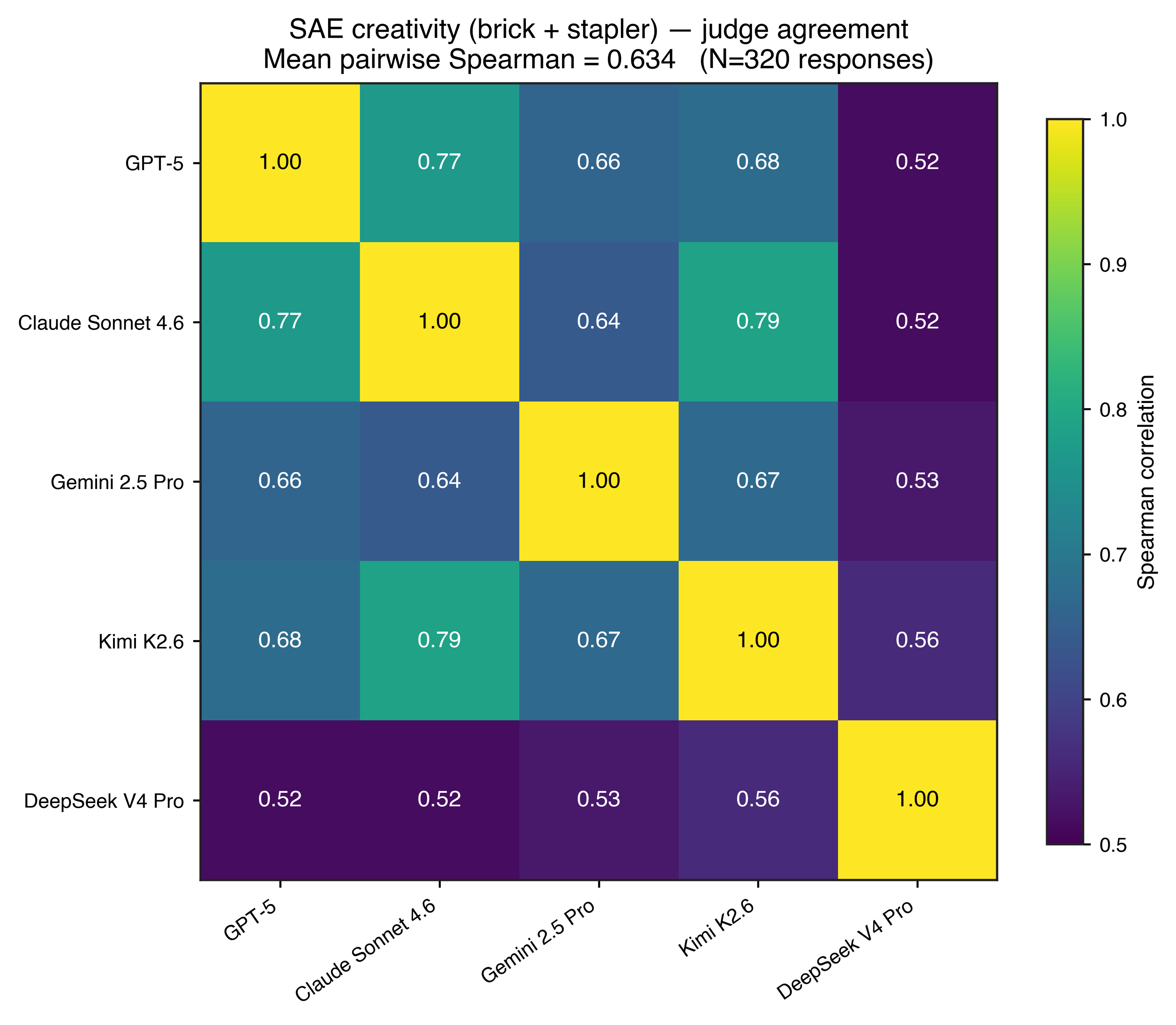}
    \caption{Pairwise Spearman correlation between the five judges on the SAE creativity dataset (brick $+$ stapler). The mean off-diagonal correlation is $0.769$; the lowest pair is Claude--DeepSeek at $0.675$ and the highest is Gemini--GPT-5 at $0.849$. Judges agree closely on the relative ordering of creative responses.}
    \label{fig:multijudge_agreement_sae}
\end{figure}

\begin{figure}[htbp]
    \centering
    \includegraphics[width=\linewidth]{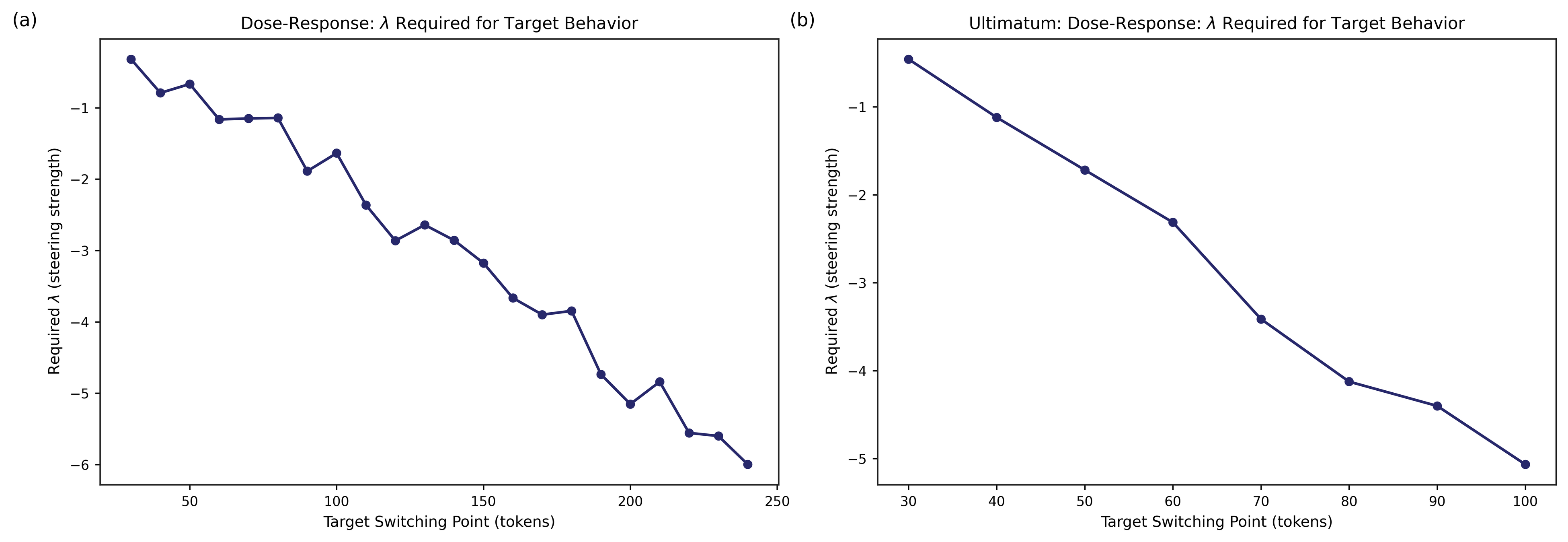}
    \caption{Dose-response showing $\lambda$ required to reach target switching points for lottery (left) and ultimatum (right) games.}
    \label{fig:dose_response_lambda}
\end{figure}

\begin{figure}[htbp]
    \centering
    \includegraphics[width=0.9\linewidth]{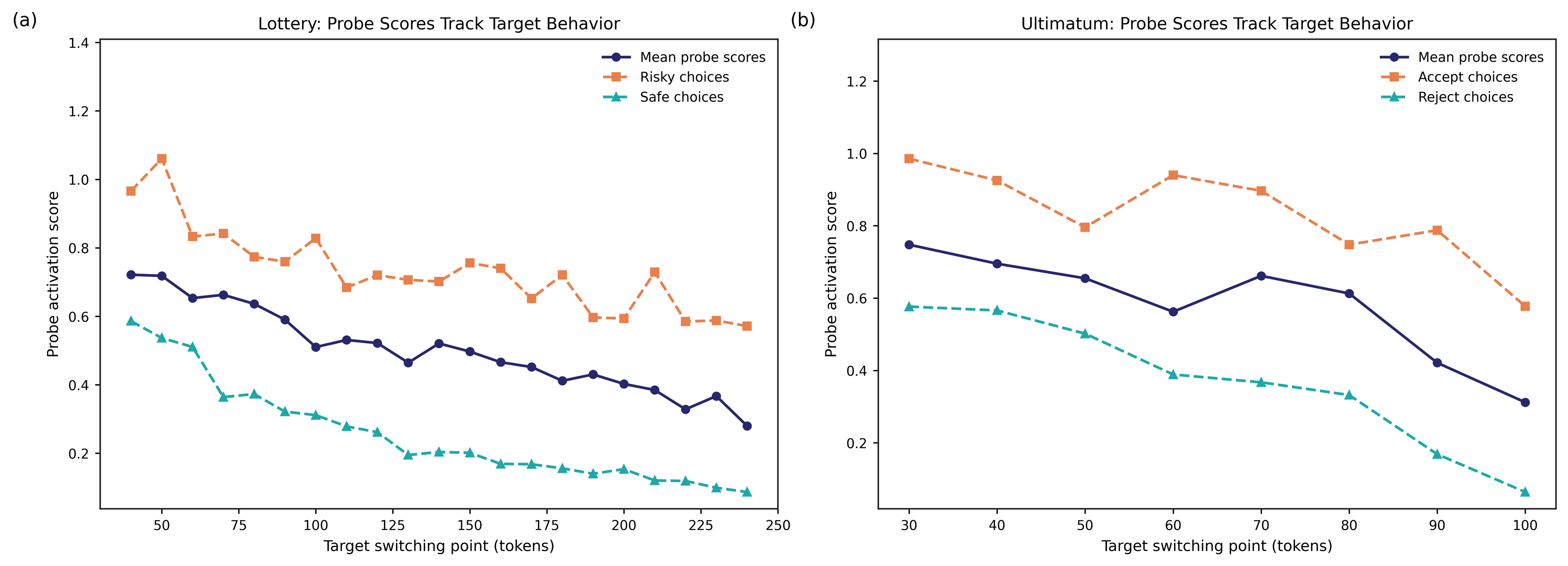}
    \caption{Probe scores tracking target behavior for lottery (left) and ultimatum (right) games on Llama-3.3-70B-Instruct (layer 48).}
    \label{fig:probe_behavior}
\end{figure}

\begin{figure}[htbp]
    \centering
    \includegraphics[width=0.6\linewidth]{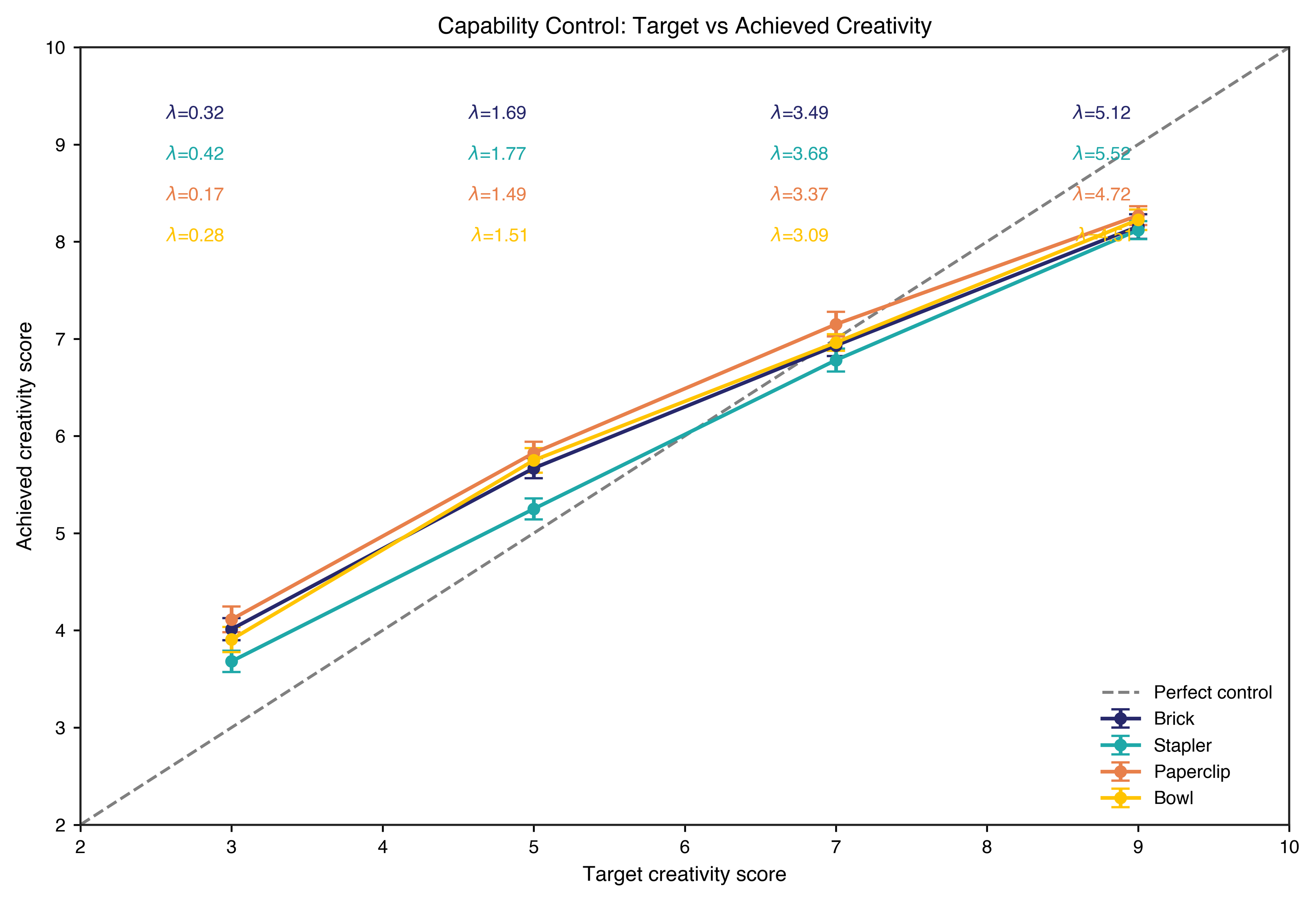}
    \caption{Achieved creativity scores as a function of target scores when the divergent-creativity probe, trained on brick-only labeled data, is applied as a fixed steering direction across four object prompts (brick, stapler, paperclip, bowl) on Llama-3.3-70B-Instruct (layer 48). Brick is the in-distribution reference; stapler, paperclip, and bowl are out-of-sample. The out-of-sample objects achieve a comparable controllability range to the in-distribution brick curve, indicating that out-of-sample control is also effective. Error bars represent standard error across 40 agents per condition; the diagonal line indicates perfect controllability.}
    \label{fig:capability_control_full}
\end{figure}

\begin{figure}[htbp]
    \centering
    \begin{subfigure}[b]{0.48\textwidth}
        \centering
        \includegraphics[width=\textwidth]{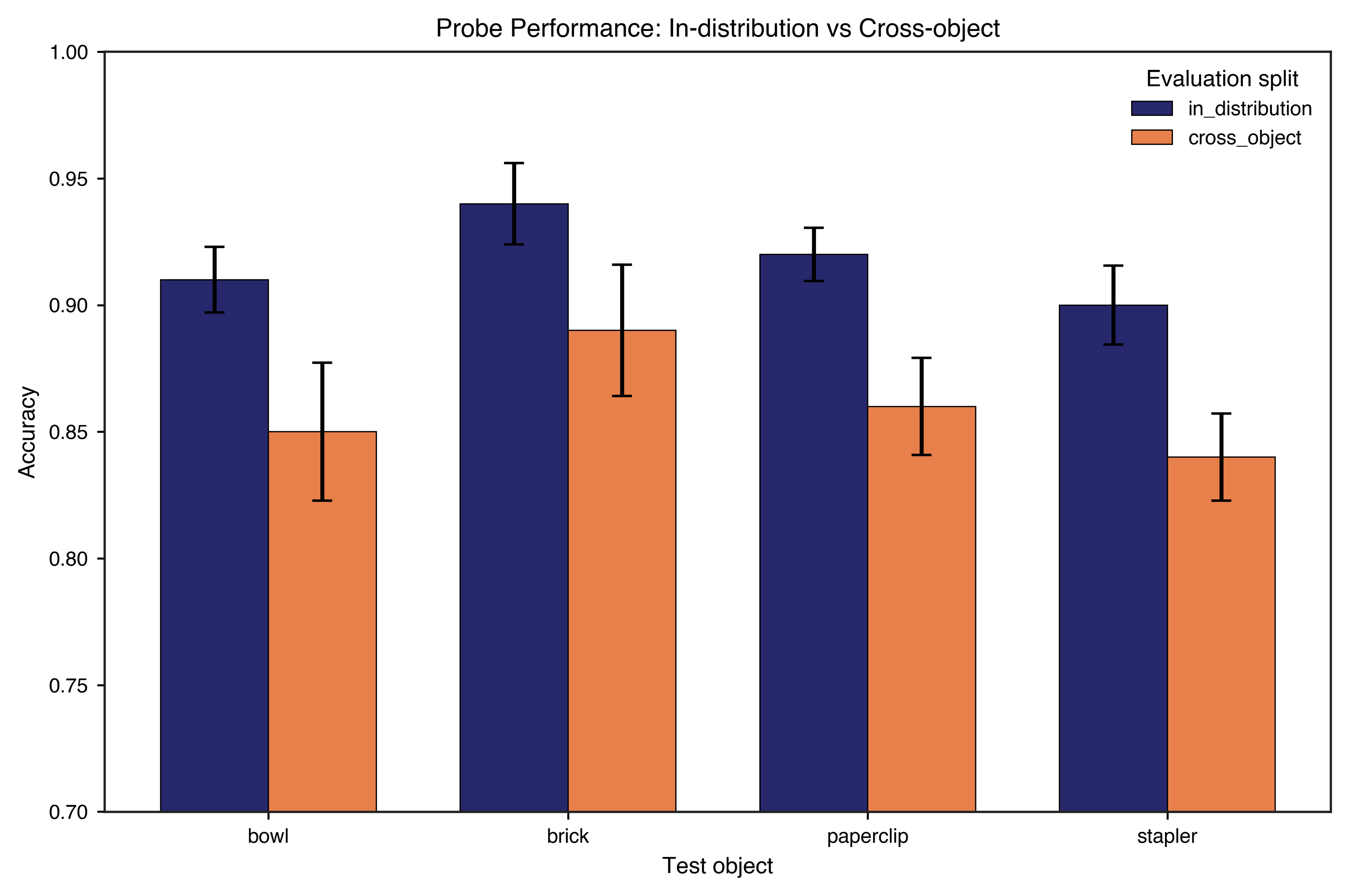}
        \caption{Cross-object generalization performance comparing in-distribution and cross-object probe accuracy across test objects (Llama-3.3-70B-Instruct, layer 48).}
        \label{fig:crossgen_bar}
    \end{subfigure}
    \hfill
    \begin{subfigure}[b]{0.48\textwidth}
        \centering
        \includegraphics[width=\textwidth]{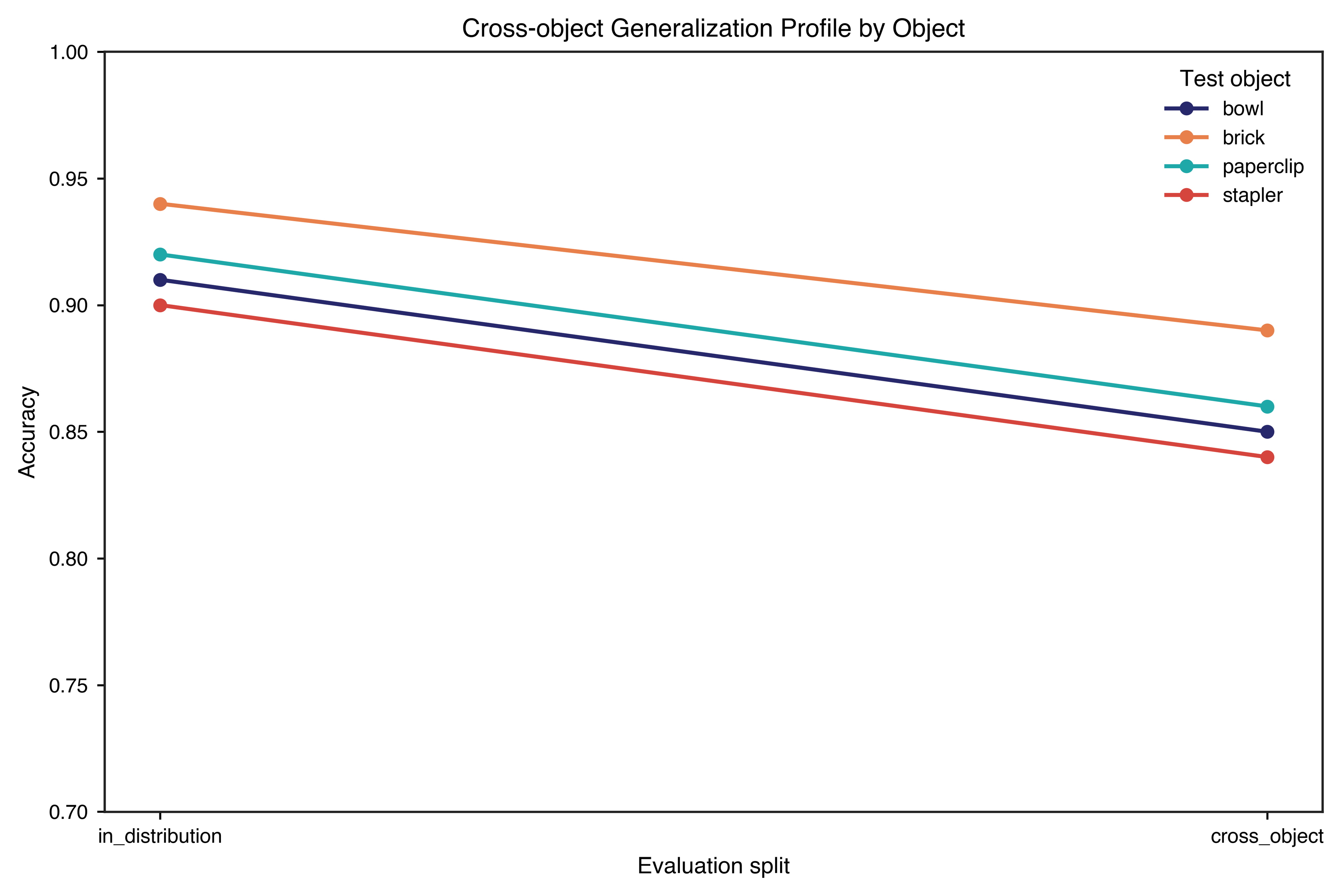}
        \caption{Performance drop from in-distribution to cross-object evaluation for each test object (Llama-3.3-70B-Instruct, layer 48).}
        \label{fig:crossgen_profile}
    \end{subfigure}
    \caption{Cross-object generalization analysis on Llama-3.3-70B-Instruct (layer 48).}
    \label{fig:crossgen}
\end{figure}

\begin{figure}[htbp]
    \centering
    \includegraphics[width=\linewidth]{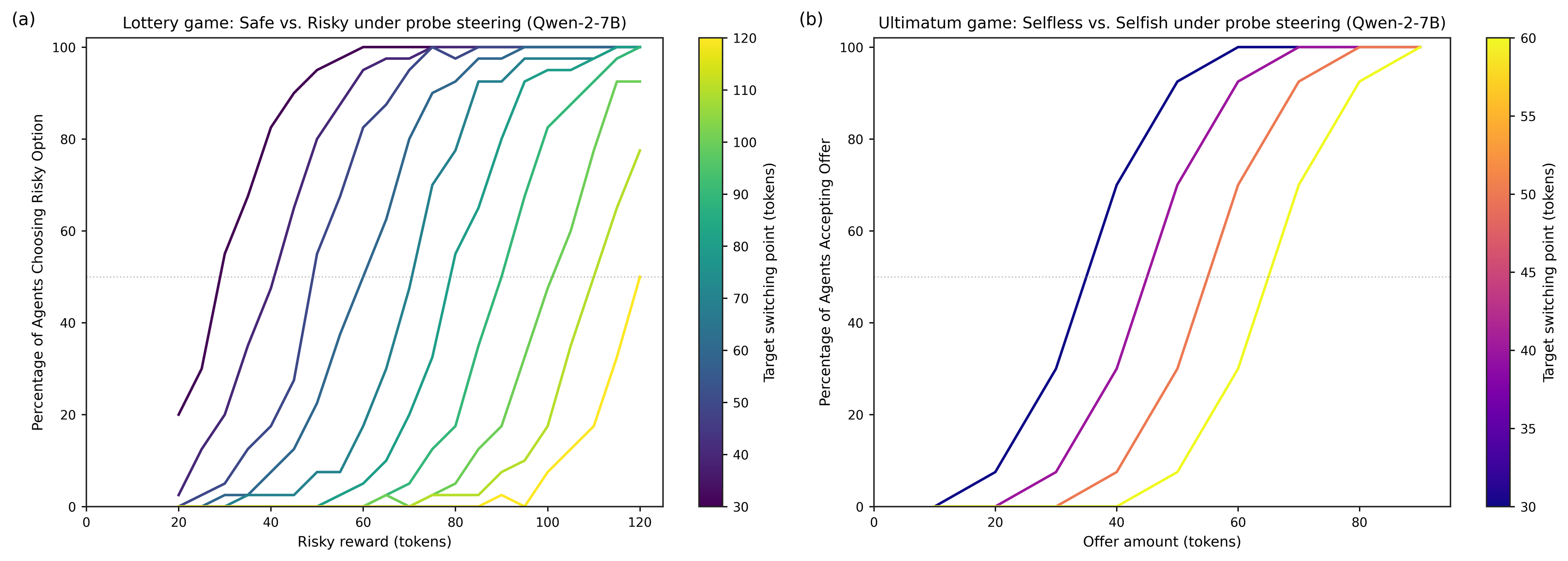}
    \caption{Psychometric curves across calibrated $\lambda$ values for lottery and ultimatum games using probe-based steering on Qwen-2-7B-Instruct (layer 17).}
    \label{fig:qwen_preference_dose}
\end{figure}

\begin{figure}[htbp]
    \centering
    \includegraphics[width=0.9\linewidth]{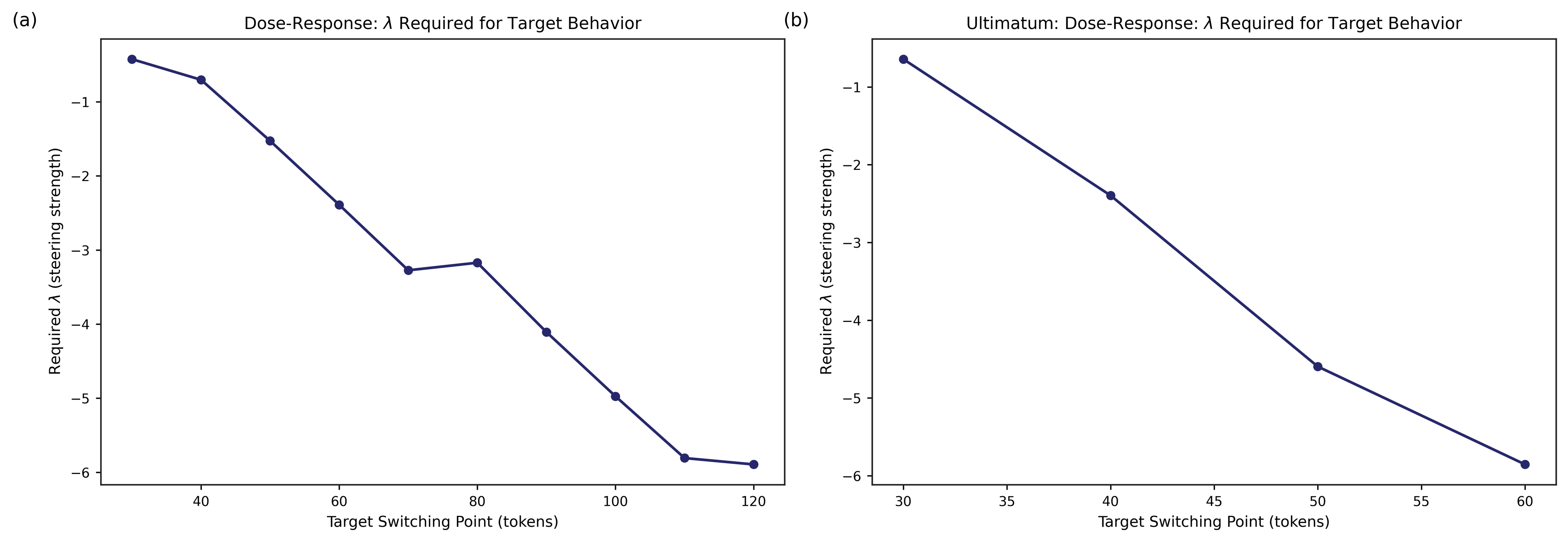}
    \caption{Dose-response showing $\lambda$ required to reach target switching points for lottery and ultimatum games on Qwen-2-7B-Instruct (layer 17).}
    \label{fig:qwen_dose_response}
\end{figure}

\begin{figure}[htbp]
    \centering
    \includegraphics[width=0.9\linewidth]{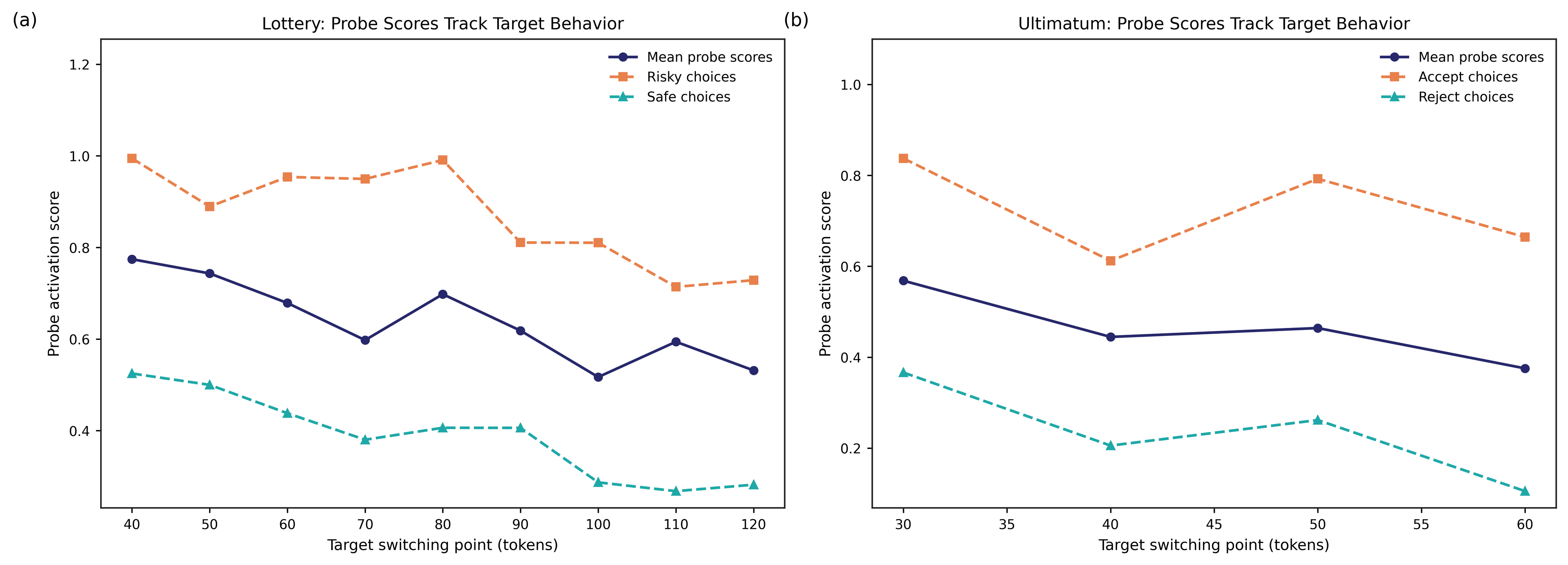}
    \caption{Probe activation score distributions for risky versus safe choices across different steering strengths on Qwen-2-7B-Instruct (layer 17).}
    \label{fig:qwen_probe_behavior}
\end{figure}

\begin{figure}[htbp]
    \centering
    \includegraphics[width=0.6\linewidth]{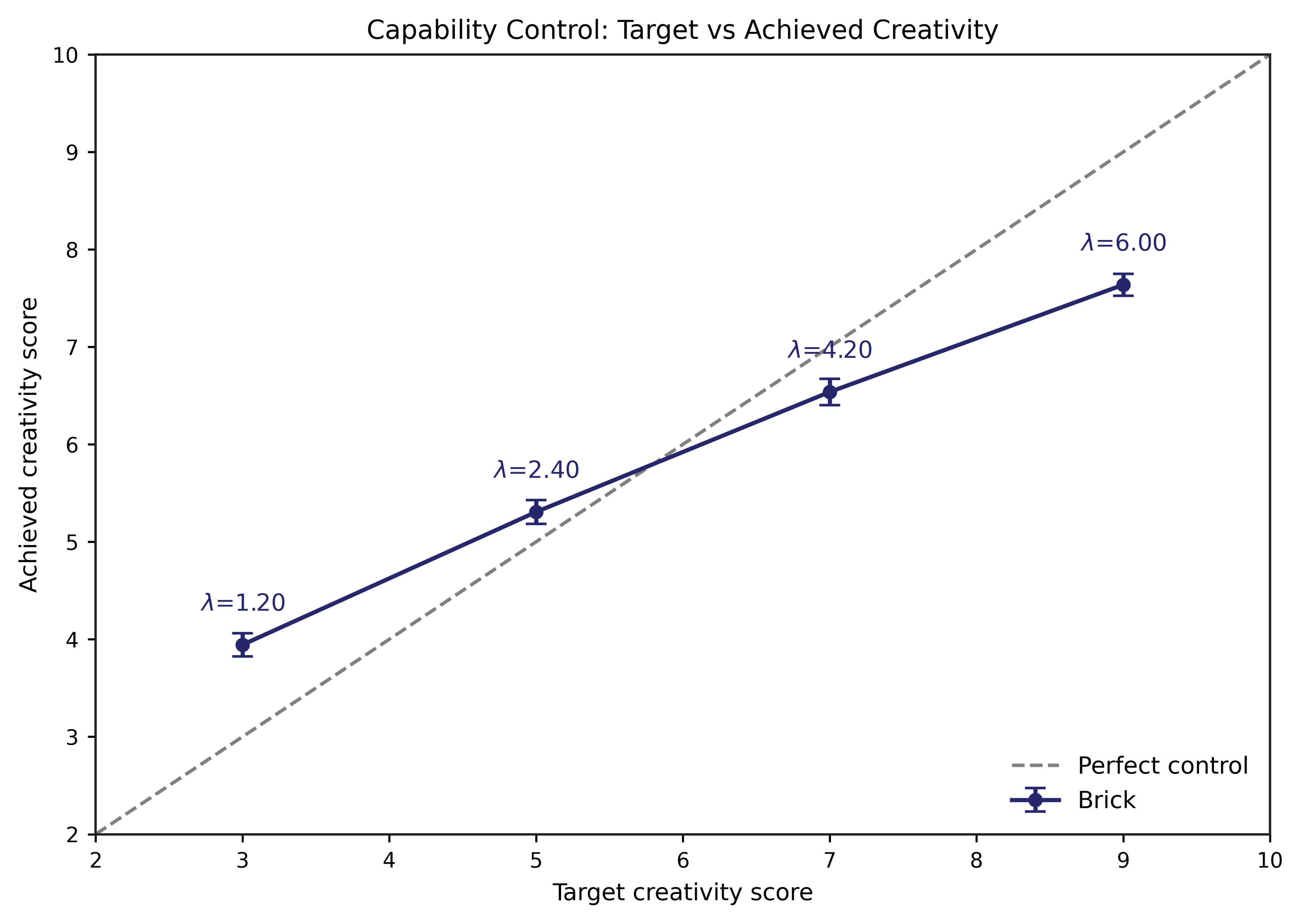}
    \caption{Achieved creativity scores as a function of target scores for the brick prompt using probe-based steering on Qwen-2-7B-Instruct (layer 17).}
    \label{fig:qwen_capability_brick}
\end{figure}

\begin{figure}[htbp]
    \centering
    \includegraphics[width=0.6\linewidth]{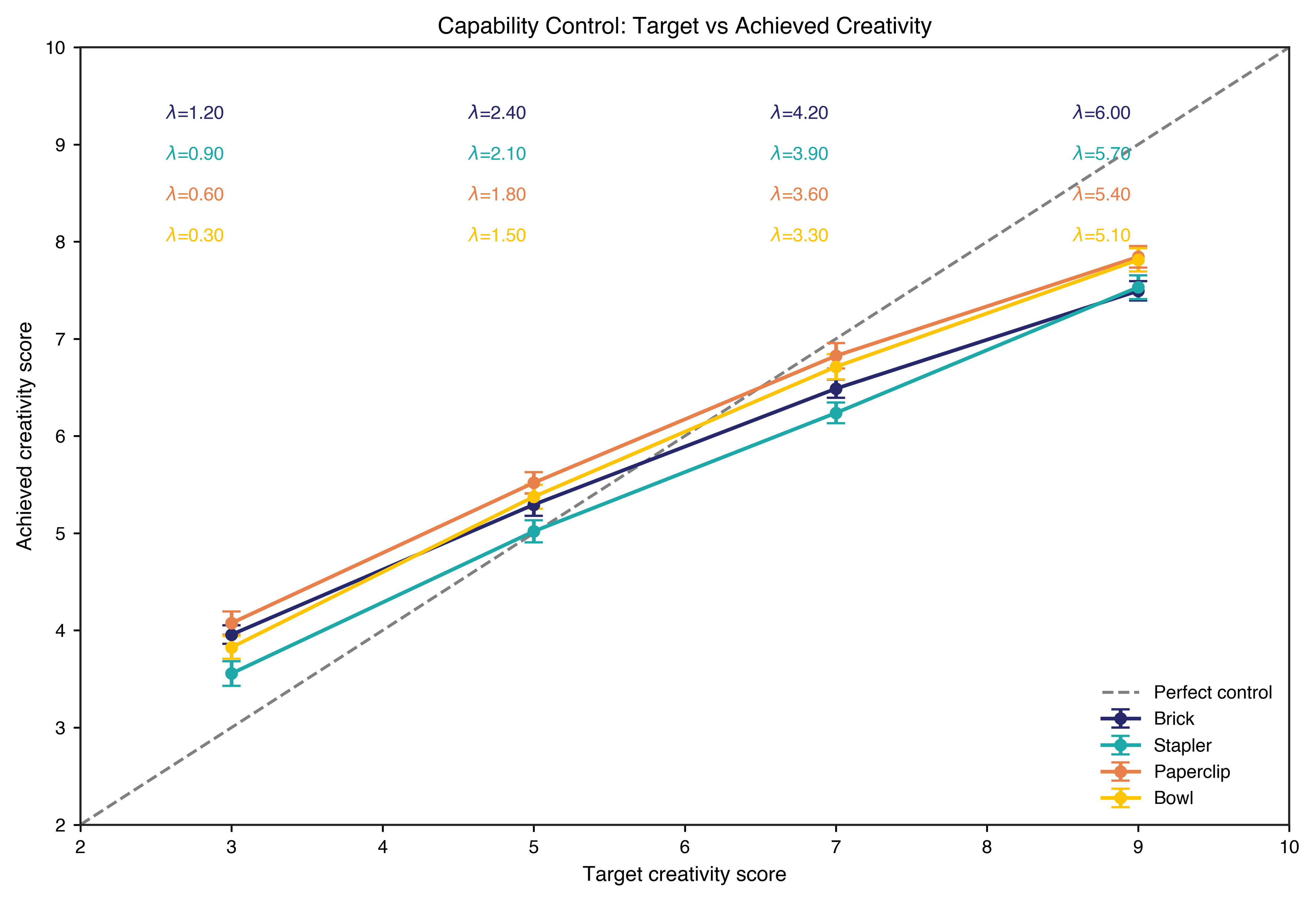}
    \caption{Achieved creativity scores across all four object prompts (brick, stapler, paperclip, bowl) using probe-based steering on Qwen-2-7B-Instruct (layer 17).}
    \label{fig:qwen_capability_full}
\end{figure}

\begin{figure}[htbp]
    \centering
    \begin{subfigure}[b]{0.48\textwidth}
        \centering
        \includegraphics[width=\textwidth]{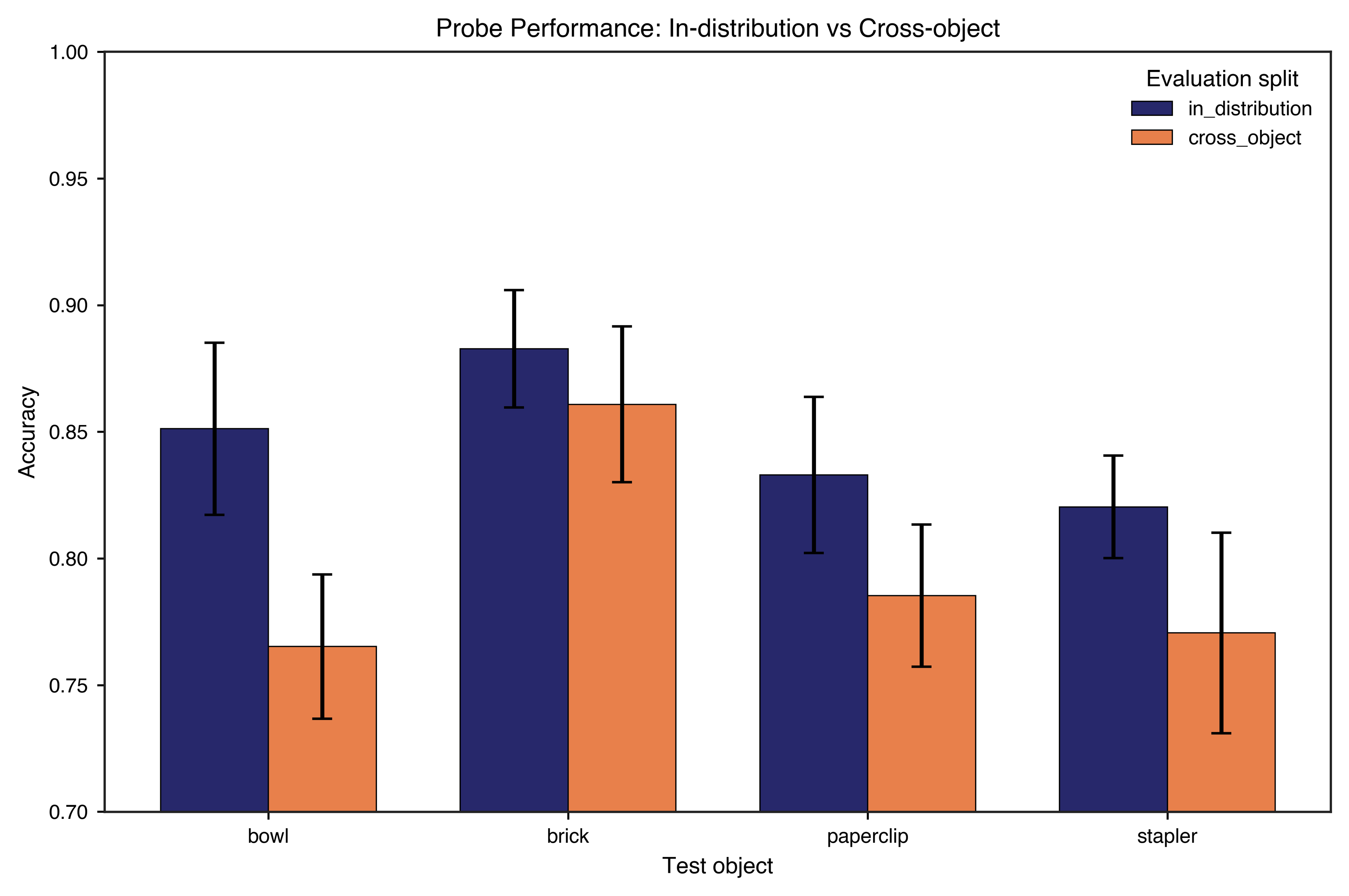}
        \caption{In-distribution vs. cross-object probe accuracy (Qwen-2-7B-Instruct, layer 17).}
        \label{fig:qwen_crossgen_bar}
    \end{subfigure}
    \hfill
    \begin{subfigure}[b]{0.48\textwidth}
        \centering
        \includegraphics[width=\textwidth]{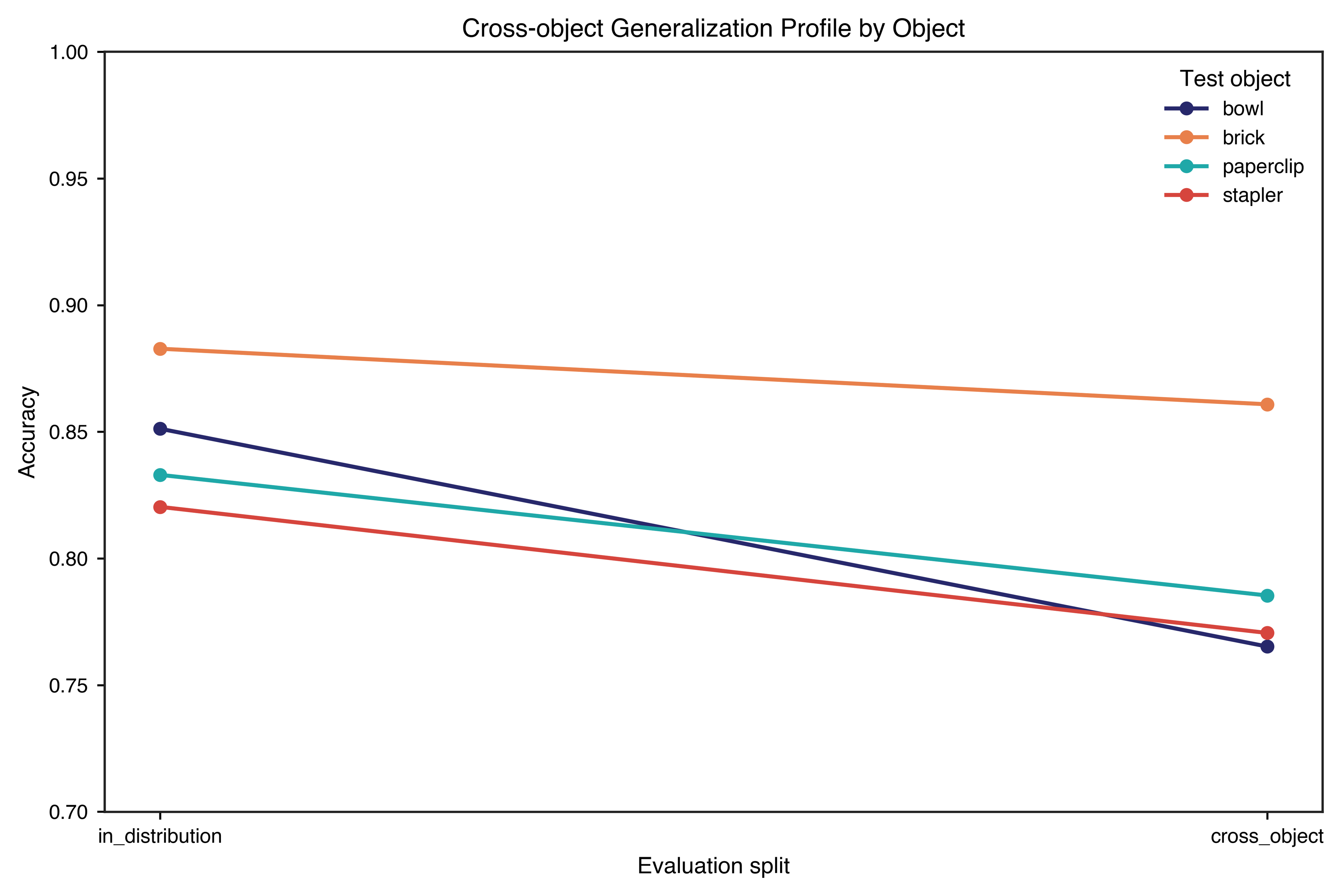}
        \caption{Generalization degradation by test object (Qwen-2-7B-Instruct, layer 17).}
        \label{fig:qwen_crossgen_profile}
    \end{subfigure}
    \caption{Cross-object generalization analysis on Qwen-2-7B-Instruct (layer 17).}
    \label{fig:qwen_crossgen}
\end{figure}